\documentclass[sigconf]{acmart}

\AtBeginDocument{%
	\providecommand\BibTeX{{%
			\normalfont B\kern-0.5em{\scshape i\kern-0.25em b}\kern-0.8em\TeX}}}

\usepackage{multirow}
\usepackage{soul}
\usepackage{mathtools}
\usepackage{amsmath,amsfonts, amsthm}
\usepackage{algorithmic}
\usepackage[ruled,vlined]{algorithm2e}
\usepackage{graphicx}
\usepackage{float}
\usepackage{lipsum}
\usepackage{subfig}
\usepackage{textcomp}
\usepackage{xcolor}
\usepackage{placeins}
\usepackage{hhline}
\usepackage{tabularx}
\usepackage{enumitem}
\usepackage{setspace}
\usepackage{booktabs}
\usepackage{adjustbox}
\usepackage[normalem]{ulem}
\usepackage{indentfirst}
\usepackage{titlesec}
\usepackage[T1]{fontenc}

\usepackage{textcomp, libertine}

\makeatletter
\newcommand{\mathleft}{\@fleqntrue\@mathmargin0pt}
\newcommand{\mathcenter}{\@fleqnfalse}
\makeatother

\theoremstyle{definition}

\let\sigproof\proof\let\proof\relax
\let\sigendproof\endproof\let\endproof\relax
\let\proof\sigproof
\let\endproof\sigendproof
\newcommand\norm[1]{\left\lVert#1\right\rVert}

\newtheorem{remark}{Remark}
\newtheorem{challenge}{Challenge}

\DeclareMathOperator*{\argmax}{arg\,max}

\let\norm\undefined 
\DeclarePairedDelimiter\norm{\lVert}{\rVert}

\setcopyright{acmcopyright}
\copyrightyear{2022}
\acmYear{2022}
\setcopyright{acmlicensed}
\acmConference[KDD '22] {Proceedings of the 28th ACM SIGKDD Conference on Knowledge Discovery and Data Mining}{August 14--18, 2022}{Washington, DC, USA.}
\acmBooktitle{Proceedings of the 28th ACM SIGKDD Conference on Knowledge Discovery and Data Mining (KDD '22), August 14--18, 2022, Washington, DC, USA}
\acmPrice{15.00}
\acmISBN{978-1-4503-9385-0/22/08}
\acmDOI{10.1145/3534678.3539288}

\begin{document}
	\title{Source Localization of Graph Diffusion via Variational Autoencoders for Graph Inverse Problems}
	
    \author{Chen Ling}
    \authornote{Both authors contributed equally to this research.}
    \affiliation{%
    \institution{Emory University}
    \city{Atlanta}
    \country{United States}}
    \email{chen.ling@emory.edu}

    \author{Junji Jiang}
    \authornotemark[1]
    \affiliation{%
    \institution{Tianjin University}
    \city{Tianjin}
    \country{China}}
    \email{anjou_j@tju.edu.cn}

    \author{Junxiang Wang}
    \affiliation{%
    \institution{Emory University}
    \city{Atlanta}
    \country{United States}}
    \email{junxiang.wang@emory.edu}
    
    \author{Liang Zhao}
    \authornote{Corresponding Author}
    \affiliation{%
    \institution{Emory University}
    \city{Atlanta}
    \country{United States}}
    \email{liang.zhao@emory.edu}
	\renewcommand{\shortauthors}{Chen Ling et al.}

	
	\begin{abstract}
	Graph diffusion problems such as the propagation of rumors, computer viruses, or smart grid failures are ubiquitous and societal. Hence it is usually crucial to identify diffusion sources according to the current graph diffusion observations. Despite its tremendous necessity and significance in practice, source localization, as the inverse problem of graph diffusion, is extremely challenging as it is ill-posed: different sources may lead to the same graph diffusion patterns. Different from most traditional source localization methods, this paper focuses on a probabilistic manner to account for the uncertainty of different candidate sources. Such endeavors require to overcome significant challenges along the way including: 1) the uncertainty in graph diffusion source localization is hard to be quantified; 2) the complex patterns of the graph diffusion sources are difficult to be probabilistically characterized; 3) the generalization under any underlying diffusion patterns is hard to be imposed. To solve the above challenges, this paper presents a generic framework: Source Localization Variational AutoEncoder (SL-VAE) for locating the diffusion sources under arbitrary diffusion patterns. Particularly, we propose a probabilistic model that leverages the forward diffusion estimation model along with deep generative models to approximate the diffusion source distribution for quantifying the uncertainty. SL-VAE further utilizes prior knowledge of the source-observation pairs to characterize the complex patterns of diffusion sources by a learned generative prior. Lastly, a unified objective that integrates the forward diffusion estimation model is derived to enforce the model to generalize under arbitrary diffusion patterns. Extensive experiments are conducted on $7$ real-world datasets to demonstrate the superiority of SL-VAE in reconstructing the diffusion sources by excelling the state-of-the-arts on average $20\%$ in AUC score. The code and data are available at: https://github.com/triplej0079/SLVAE.
	\end{abstract}

	
	
	\maketitle
    
    \section{Introduction}
\begin{figure}[!t]
 		\centerline{\includegraphics[width=0.5\textwidth]{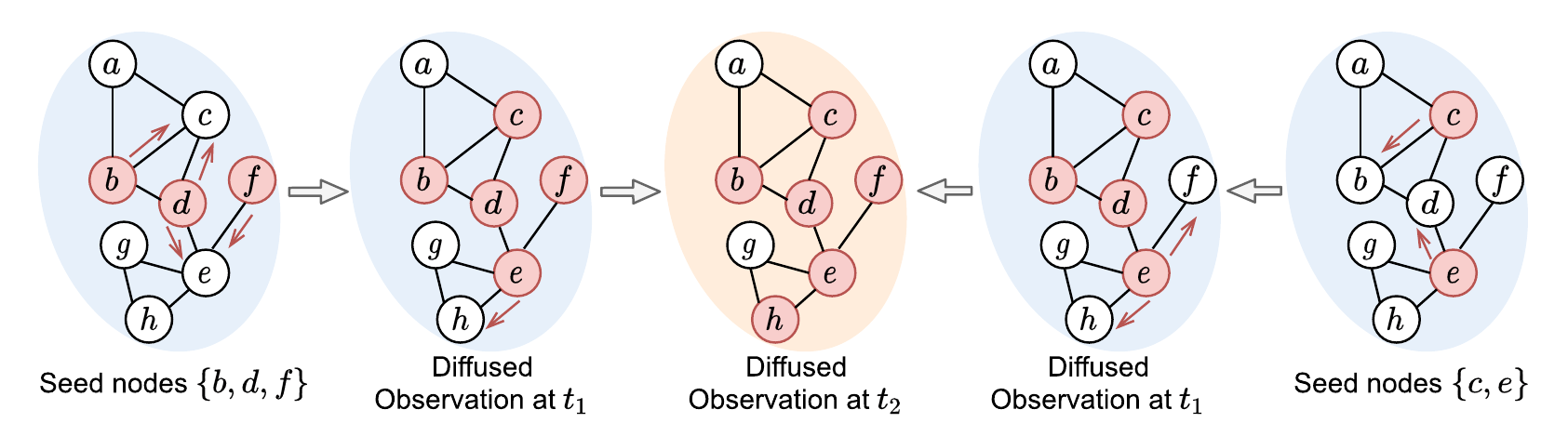}}
 		\vspace{-3mm}
 		\caption{Example of ill-posed source localization problem: two set of diffusion sources with no intersection can produce the same infected graph.}
 		\label{fig: example}
 		\vspace{-6mm}
 	\end{figure}

    Networks formed in the real world are changing the way how people interact and communicate in daily life, and the ubiquity of networks has also made us vulnerable to various network risks. For example, online social media like Twitter and Facebook enable worldwide people to express individual opinions. However, it has also been used to spread rumors and other forms of misinformation, leading to severe consequences: hundreds of death were caused by COVID-related misinformation in the first quarter of 2020 \cite{evanega2020coronavirus}. In addition, computer viruses can rapidly propagate throughout the Internet and infect millions of computers \cite{kephart1993measuring}. In smart grids (i.e., electricity networks), isolated failures could lead to rolling blackouts that create billions of financial losses \cite{amin2007preventing}. From both practical and technical aspects, it is critical to identify the propagation origins accurately, and therefore, diminish the damages through cutting off the critical paths of the propagation. However, as the inverse problem of information propagation estimation, the difficulty of the ill-posed source localization problem is at least two-fold: \emph{1).} different diffusion sources may lead to the exactly same observation. As shown in Fig. \ref{fig: example}, different diffusion sources $\{b, d, f\}$ and $\{c, e\}$ generate the same diffusion pattern, and hence it is highly difficult to distinguish which should be correct merely based on the observation at $t_2$; \emph{2).} locating diffusion sources on the graph may require the exploration of the whole topology space as well as the overall node attributes, which may lead to high computational complexity for finding a near-optimal solution or simplified heuristics to achieve sub-optimal performance~\cite{jiang2016identifying}. 
    
    In past years, researchers have proposed a series of methods \cite{prakash2012spotting, zang2015locating, wang2017multiple, zhu2017catch, dong2019multiple, zhu2016information} to identify diffusion sources. Earlier works \cite{zang2015locating, prakash2012spotting, zhu2016information} leveraged deterministic algorithms to locate a fixed number of sources under a prescribed diffusion pattern (e.g., Susceptible-Infected Model), and their models cannot be generalized to locate a varying number of sources in different diffusion patterns. Moreover, \cite{zhu2017catch, wang2017multiple} extended the previous models to locate multiple diffusion sources under various simulated prescribed diffusion patterns; however, their expressiveness and scalability are still limited due to the incapability of encoding the information of graph topology and the need of enumerating all possible solutions. Furthermore, a recent work~\cite{dong2019multiple} leveraged the Graph Neural Network to encode neighborhood and graph topology information into a latent node representation for the label propagation, which achieves the state-of-the-art performance. To date, all the existing source localization techniques tend to build various deterministic algorithms to compute the diffusion sources directly from a diffused observation. Nevertheless, the ill-posedness of the source localization task says its solution can usually be not unique (as shown in Fig. \ref{fig: example}), while existing approaches still lack the ability to build an adequate model for characterizing such uncertainty of sources that can lead to the same diffusion observations, let alone the quantification of the likelihoods of different candidate inferences of sources. 
    
    The under-exploration of statistic models for accounting for the uncertainty of sources is majorly due to the difficulty in the nature of this problem. Quantifying the uncertainty requires automatically learning the probabilistic models of complex data, which has been a prohibitively challenging domain until the past few years when deep generative models started to excel out. Existing deep generative models have been applied to the relevant inverse problem but typically for image data. Specifically, existing works \cite{bora2017compressed, bora2018ambientgan, kamath2019lower} have been utilizing deep generative models to build a probabilistic model between the input source and the observation so that one can estimate the ``optimal'' source to the observation through the approximated posterior probability. However, approaches that are utilized for solving inverse problems in the computer vision domain cannot be trivially adapted to the graph domain because of the following technical challenges: \emph{1) The difficulty of quantifying the uncertainty in source localization of graph diffusion.} Quantifying the uncertainty in diffusion source localization requires one to build conditional probability of the graph with diffusion sources given the graph with diffused observations. Image-related inverse problem treatments cannot be directly adapted here due to the additional consideration on graph topology as well as how it impacts node patterns. Moreover, the probability distribution of graph data is harder to be modeled and optimized efficiently by algorithms like gradient descent since its non-Euclidean discrete data. 
    \emph{2) The difficulty of characterizing the intrinsic patterns of the diffusion sources.} Characterizing the patterns of diffusion sources is essential since diffusion sources are often conditioned on the intrinsic nature of the nodes and their connections. Such information is apart from the diffused observations but can predominantly help determine the sources. For example, accounts that have conducted abnormal activities in a period tend to be more suspicious of being the misinformation diffusion sources in social networks. However, modeling the distribution of the diffusion sources on the graph requires us to consider combinatorial patterns among discrete statuses of different nodes. Such patterns are exponentially high-dimensional and often intractable, leading to the distribution of diffusion sources unable to be trivially embedded into the source localization problem. \emph{3) The difficulty of imposing the generalization under any underlying diffusion patterns.} Most of the existing source localization methods are tailored for specific diffusion processes such as Linear Threshold, Independent Cascade, and Epidemic models, which usually impose strong assumptions on the diffusion process. It is imperative yet challenging to establish a source localization framework that can adapt to different forward models to locate sources under any information diffusion patterns.

    In this work, to solve the challenges mentioned above, we propose a novel generic framework, namely Source Localization Variational Autoenocder (SL-VAE), for locating the diffusion sources given its diffused observation under arbitrary diffusion patterns. Particularly, to deal with the first challenge, we formulate the probabilistic model by leveraging a forward diffusion estimation model along with the approximated source distribution by deep generative models to quantify the uncertainty. To tackle the second challenge, SL-VAE learns a generative prior from the observed source-observation pairs as the way to encode the prior knowledge of the diffusion sources. Lastly, we derive a unified objective function that integrates the forward information propagation estimation model into the overall learning scheme such that the reconstruction of diffusion sources is fully aware of any diffusion patterns. We summarize our contributions in this work as follows:
    \begin{itemize}[leftmargin=*]
        \item A new generic framework for probabilistic source localization in graph diffusion. 
        To the best of our knowledge, this work is the first attempt to leverage deep graph generative models to characterize the prior and conditional probability for source localization problem of graph diffusion.
        \item A new objective is proposed to mutually learn forward and probabilistic models for reversely locating the diffusion sources. Moreover, the proposed SL-VAE can be coupled with any diffusion estimation models to reconstruct the unique diffusion sources to the diffused observations, which can be generalized under any information diffusion patterns.
        \item A variational inference-based method and associated optimization strategies have been customized for inferring the optimal diffusion source given its diffused observation, which leverages historical observations to characterize the intrinsic patterns of the diffusion sources for better prediction.
        \item Extensive experiments are conducted on $7$ real-world datasets. Compared with existing approaches, SL-VAE achieves superior results in locating diffusion sources under varying diffusion patterns by on average $20\%$ in AUC-ROC score and $10\%$ in F-1 score.
    \end{itemize}
    

    \section{Related Works}\label{sec: related_works}
    \subsection{Information Diffusion Estimation}
    Information diffusion estimation on graphs is a task of approximating the expected number of influenced nodes given a set of diffusion sources \cite{xia2021deepis}. Early works have proposed different approaches: Linear Threshold (LT)~\cite{kempe2003maximizing}, Independent Cascade (IC)~\cite{kempe2003maximizing}, and Epidemic Models~\cite{keeling2005networks} to estimate the influence propagation status. However, these prescribed methods are rather limited in their generalization capability, and computing the influence spread exactly under these models is NP-hard \cite{chen2010scalable}. Following works have been using learning-based models (e.g., recurrent neural networks \cite{chen2019npp, xie2020multimodal}, self-attention mechanism \cite{bielski2018understanding}, and stochastic processes \cite{du2016recurrent,ling2020nestpp}) to predict the diffusion status. However, they also failed to incorporate graph topology in their learning schemes. Recently, graph neural networks (GNNs) have achieved success in many graph mining tasks. Many works \cite{wu2022graph, ko2020monstor, velivckovic2017graph, xia2021deepis} have employed GNNs to predict the influence spread states since GNNs can naturally incorporate graph topology to enhance the estimation accuracy. We refer readers to recent surveys \cite{shelke2019source} for more details.
    
    \subsection{Source Localization}
    As the inverse problem of information diffusion estimation, diffusion source localization of online information diffusion is to retrospectively infer the initial diffusion sources given the current diffused observation, which have a wide range of applications including identifying rumor sources in social networks~\cite{jiang2016identifying} and finding origins of a rolling blackout in smart grids \cite{shelke2019source}. In past years, researchers have proposed various methods \cite{wang2017multiple, prakash2012spotting, zhu2017catch, dong2019multiple, zhu2016information, wang2022} to identify diffusion sources. Early approaches~\cite{prakash2012spotting, prakash2012spotting} focused on identifying the single source of an online disease under the Susceptible-Infected (SI) diffusion pattern by employing the Minimum Description Length principle and approximate multi-source locating algorithm, respectively. Approaches \cite{zhu2016information, zhu2017catch, zang2015locating} have been further proposed to generalize the source localization model to predict rumor sources under the Susceptible-Infected-Recovered (SIR) diffusion pattern with partial observation. Later on, Wang et al. \cite{wang2017multiple} proposed a model named LPSI to automatically detect multiple diffusion sources without knowing any prescribed diffusion patterns, and Dong et al. \cite{dong2019multiple} further leverage GNN to enhance the prediction accuracy of LPSI. However, existing diffusion source localization methods cannot well quantify the uncertainty between different diffusion source candidates, and they usually require searching over the high-dimensional graph topology and node attributes to detect the sources, both drawbacks limit their effectiveness and efficiency. Moreover, existing diffusion source localization methods are tailored for specific diffusion patterns (e.g., LT, IC), which further limits their generalization capability to perform on unseen and arbitrary diffusion patterns.
    
    \subsection{Generative Models for Inverse Problem}
    Due to the ill-posedness of inverse problems, locating diffusion sources is challenging without some prior knowledge about the data (e.g., the distribution of diffusion sources) \cite{hegde2018algorithmic}. Traditional approaches tend to hand-craft priors for solving inverse problems, but these hand-crafted priors are less expressive and usually intractable in many scenarios. Deep generative models provide a different way of modeling the high-dimensional and intractable prior distribution through representing a complex distribution to a simple distribution (e.g., standard normal distribution) with a deterministic transformation \cite{guo2020systematic, ling2021deep, zhang2021tg, du2021graphgt}. Two primary deep generative models: Variational Auto-Encoders (VAEs)~\cite{kingma2013auto} and Generative Adversarial Networks (GANs)~\cite{goodfellow2014generative} have been widely used to learn the generative priors in many image-related inverse problems (e.g., image inpainting~\cite{yu2018generative} and image super-resolution~\cite{ledig2017photo}). To the best of our knowledge, the proposed SL-VAE is the first work that utilizes deep generative models to solve graph inverse problems in practical application (i.e., source localization). 
    
    
    \section{Source Localization VAE (SL-VAE)}\label{sec: model}
    This section first provides the problem formulation before moving on to derive the overall objective from divergence-based variational inference. We then provide the overall structure of SL-VAE in the form of the probabilistic graphical model. Finally, we provide a novel optimization algorithm to optimize the diffusion source given a corresponding diffused observation.

    \subsection{Problem Formulation}\label{sec: formulation}
    Given a graph $G = (V, E)$, where $V$ is the set of nodes and $E$ is the set of edges. Suppose there is a set of seed indicators $x \in \{0, 1\}^{|V|}$ (i.e., diffusion source), where $1$ means infected and $0$ means uninfected, a probability of each node being infected: $y = [0, 1]^{|V|}$ (i.e., diffused observation) can be estimated through a forward diffusion estimation model. The problem of information diffusion source localization (i.e., the inverse problem of the information diffusion estimation) is then defined as reconstructing a unique solution $\Tilde x \in \{0, 1\}^{|V|}$ from the diffused observation $y$ such that the empirical loss $\norm{\Tilde x - x}_2^2$ is minimized.   However, reconstructing $\Tilde x$ from $y$ is exceptionally difficult due to the following challenges.

    
    \begin{challenge}[\textbf{Quantifying the uncertainty}]\label{challenge: 1}
    To quantify the uncertainty in the diffusion source localization, we need to build a probabilistic model to characterize the conditional probability $p(x|y)$. However, both $x$ and $y$ are affiliated to the graph $G$, leading to characterize the conditional probability more challenging since the topology of $G$ determines the overall diffusion process, and existing works cannot be directly adapted due to the incapability of considering the complex and irregular graph topology.
    \end{challenge}
    
    \begin{challenge}[\textbf{Intrinsic patterns of the diffusion source}]\label{challenge: 2}
    The intrinsic patterns of the diffusion source $x$ are hard to be modeled since they are unknown to us in most cases. Even if we could model the intrinsic patterns of $x$ as the prior distribution $p(x)$, such distribution $p(x)$ is often high-dimensional and intractable, which makes maximizing the joint likelihood $p(x, y)$ to be hard and computationally inefficient.
    \end{challenge}
    
    \begin{challenge}[\textbf{Generalization under various diffusion patterns}]\label{challenge: 3}
    The underlying diffusion process from $x$ to $y$ is affected by numerous factors, including the type of network disease, immunity power, transmission rate, and network parameters. However, existing works tend to pre-define a diffusion pattern (e.g., IC, LT) and analyze the diffusion source localization on such patterns. Therefore, it is necessary to generalize the diffusion source localization method for handling any underlying diffusion patterns in order to make it practical in real life. 
    \end{challenge}
    
    \subsection{Quantification of Diffusion Source}
    The first challenge posed in Section \ref{sec: formulation} requires us not only to build a probabilistic model for quantifying the uncertainty in the diffusion source localization, but also efficiently leverage the graph topology to characterize the conditional probability $p(x|y)$. Given the fact that the diffused observation $y$ is conditioned on the graph $G$ and the diffusion source $x$, a conditional probability $p(y|x, G)\cdot p(x)$ can be obtained, where the $p(x)$ is the distribution of the infection sources. We can naturally utilize the Maximum A Posteriori (MAP) approximation to estimate the optimal diffusion source $\Tilde x$ by maximizing the following probability:
    \begin{equation*}
        \Tilde x = \underset{x}{\argmax} \:p(y|x, G)\cdot p(x) = \underset{x}{\argmax} \:p(x, y| G).
    \end{equation*}
    However, since the distribution of the diffusion source $p(x)$ is often intractable, as suggested in the second challenge proposed in Section \ref{sec: formulation}, we alternatively leverage deep generative models to characterize such implicit distribution. 
    
    The strategy is to map the intractable and potentially high-dimensional $p(x)$ to latent embeddings in low-dimensional semantic space in order to approximate $p(x)$ and reduce the computational cost. Specifically, the latent random variable $z \in \mathbb{R}^k$ ($k \ll |V|$) is obtained by an approximate posterior $p(z|x, y, G)$, and $p(z)$ is the graph prior distribution in latent semantic lower dimensions. We could define a joint probability distribution: $p(x, y, G, z)=p(x,y,G|z)\cdot p(z)$. Therefore, the posterior: $p(z|x, y, G)$ can be leveraged to infer the lower-dimension latent variable $z$ given the joint likelihood $p(x, y, G)$.  However, because of the intractable $p(x)$, we instead model the approximate posterior $q_{\phi}(z|x, y, G)$ parametrized by $\phi$, and  Kullback-Leibler (KL) divergence is utilized to measure the approximation error between $q_{\phi}(z|x, y, G)$ and $p(z|x, y, G)$. Therefore, the approximate posterior $q_{\phi}(z|x,y, G)$ can be obtained as:
    \begin{align}\label{eq:kl}
        q_{\phi}(z|x,y, G)&=\underset{\phi}{\min}\:KL\big[q_{\phi}(z|x, y, G)\vert \vert p(z|x,y, G)\big]\\
        &\hspace{-5.5em}=\underset{\phi}{\min}\big[\mathbb{E}_{q_{\phi}}[\log q_{\phi}(z|x, y, G)]-\mathbb{E}_{q_{\phi}}[\log p(x, y, G, z)]+\log p(x, y, G)\big].\nonumber
    \end{align}
    \begin{figure}[!t]
    \centering
    \includegraphics[width=0.48\textwidth]{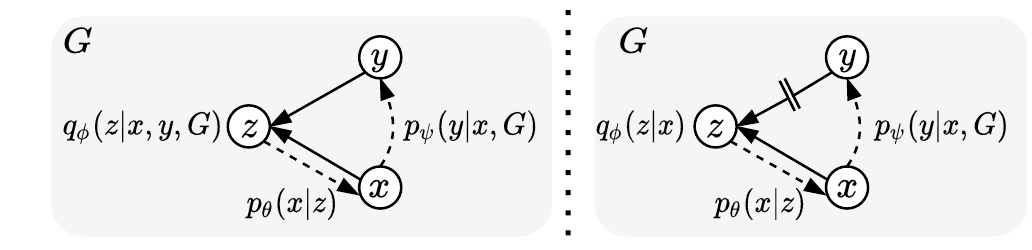}
    \vspace{-8mm}
    \caption{Graphical model of SL-VAE. For the left figure, solid arrows represent the variational approximation $q_{\phi}(z|x, y, G)$ to the intractable posterior $p(z|x, y, G)$, and dashed arrows denote the generative process $p_{\psi}(y|x, G)\cdot p_{\theta}(x|z)$. Since $y$ and $z$ are conditionally independent given $x$ and $G$, we could simplify SL-VAE to the right of Fig. \ref{fig: architecture}.}
    \label{fig: architecture}
    \vspace{-4mm}
    \end{figure}
    For the sake of simplicity, we omit the subscript of $\mathbb{E}_{q_{\phi}(z|x, y, G)}$ as $\mathbb{E}_{q_{\phi}}$ when the context is clear. Note that Eq. \eqref{eq:kl} cannot be computed directly because of the intractable joint distribution $p(x, y, G)$. However, given an approximate posterior $q_{\phi}(z|x,y, G)$, the Evidence Lower BOund (ELBO) allows us approximating the posterior as:
    \begin{equation*}
        ELBO=\mathbb{E}_{q_{\phi}}[\log \:p(x,y,G,z)] - \mathbb{E}_{q_{\phi}}[\log\:q_{\phi}(z|x,y, G)].
    \end{equation*}
    By Jensen’s inequality, we can instead approximate the posterior through maximizing the ELBO, which is more computationally efficient than directly calculating the KL divergence. Therefore, we can minimize the KL divergence between the approximate posterior $q_{\phi}(z|x, y, G)$ and $p(z|x, y, G)$ by optimizing the negative ELBO:
    \begin{equation}\label{eq: objective_1}
         -ELBO={-\mathbb{E}_{q_{\phi}}[\log\: p_{\theta}(x,y,G|z)]}+{KL[q_{\phi}(z|x, y, G)||p(z)]},
    \end{equation}
    where the likelihood $p_{\theta}(x,y,G|z)$ is parametrized by $\theta$ as a generative network (i.e., decoder), and the $q_{\phi}(z|x,y,G)$ parametrized by $\phi$ performs as an inference network (i.e., encoder). We provide the graphical model of SL-VAE in the left part of Fig. \ref{fig: architecture}.
    

    \textbf{Independence between the latent variable $z$ and the diffused observation $y$.}
    In most of information diffusion estimation models~\cite{velivckovic2017graph, ko2020monstor, xia2021deepis}, the diffused observation $y$ is only determined by the diffusion source $x$ under the graph $G$. Therefore, we can further decouple the likelihood $p_{\theta}(x,y, G|z)$ in Eq. \eqref{eq: objective_1} as:
    \begin{equation}\label{eq:decouple}
        \log p_{\theta}(x,y, G|z) = \log [p_{\psi}(y|x, G)] + \log [p_{\theta}(x|z)],
    \end{equation}
    where $p_{\psi}(y|x, G)$ parametrized by $\psi$ is the likelihood function of a forward information diffusion model that measures $p(y)$ from $p(x)$ and $G$. In addition, the second term $p_{\theta}(x|z)$ in Eq. \eqref{eq:decouple} reveals that the latent variable $z$ only encodes information from $x$ (i.e., $y \bot z|x, G$). According to the assumption, we could also simplify the encoder $q_{\phi}(z|x, y, G)$ in Eq. \eqref{eq:kl} to $q_{\phi}(z|x)$. And finally, the negative ELBO can be extended from Eq. \eqref{eq: objective_1} as:
    \begin{align*}
    &\resizebox{\hsize}{!}{$
        -ELBO = -\mathbb{E}_{q_{\phi}}\big[\underbrace{\log\: p_{\psi}(y|x,G)}_{\text{Forward Loss}}+\underbrace{\log\: p_{\theta}(x|z)}_{\text{Reconstruction Loss}}\big] + \underbrace{KL[q_{\phi}(z|x)||p(z)]}_{\text{Regularization}},
        $}
    \end{align*}
    where the KL divergence of above decomposition enforces the approximation of the posterior $q_{\phi}(z|x)$ such that the likelihood of both $p_{\theta}(x|z)$ and $p_{\psi}(y|x, G)$ can be maximized. The proposed Bayesian variational inference framework allows us to jointly maximize both likelihoods $p_{\psi}(y|x, G)$ and $p_{\theta}(x|z)$ in order to construct a direct mapping between the observation $y$ to a unique estimated diffusion source $\Tilde x$. That means, Challenge \ref{challenge: 2} can be solved by generatively modeling the intrinsic pattern of diffusion sources as $p_{\theta}(x|z)$. The updated probabilistic graphical model of SL-VAE by considering the conditional independence between $z$ and $y$ is illustrated in the right part of Fig. \ref{fig: architecture}.

    \textbf{Monotonicity constraint on information diffusion.} In addition to optimize the derived variational inference framework, the information diffusion on network is regularized by the monotone increasing property~\cite{dhamal2016information}, namely $y^{(i)} \succeq y^{(j)} ,\: \forall \: x^{(i)} \supseteq x^{(j)}$. Intuitively, if one diffusion source set $x^{(i)}$ is the superset of another one $x^{(j)}$, then the probability of each node being infected in $y^{(i)}$ (estimated from $x^{(i)}$) should be greater or equal to $y^{(j)}$ (estimated from $x^{(j)}$), such that $y^{(i)} \succeq y^{(j)}$. The assumption is applied in many network information diffusion tasks (e.g., influence maximization \cite{dhamal2016information} and information forward estimation \cite{shelke2019source}), where each node should contribute non-negatively to the information diffusion. Hence, owing to the monotone increasing property of the information diffusion, we formulate the following constrained learning objective:
    \begin{align}\label{eq: objective_3}
        -ELBO= &-\mathbb{E}_{q_{\phi}}\big[{\log\: p_{\psi}(y|x,G)}+{\log\: p_{\theta}(x|z)}\big]+ KL[q_{\phi}(z|x)||p(z)]\nonumber\\ &\:\text{s.t.}\:  y^{(i)} \succeq y^{(j)} , \forall \: x^{(i)} \supseteq x^{(j)}
    \end{align}
    However, such a learning objective with inequality constraints would bring a huge number of constraints when the number of nodes is huge. To address this issue, we transform the constrained Eq. \eqref{eq: objective_3} into its augmented Lagrangian form as follows: 
    \begin{align}\label{eq: objective_4}
        -ELBO= &-\mathbb{E}_{q_{\phi}}\big[{\log\: p_{\psi}(y|x,G)}+{\log\: p_{\theta}(x|z)}\big]\\
        &\hspace{-2em}+ KL[q_{\phi}(z|x)||p(z)]\nonumber+\lambda \big|\big|\max(0, y^{(j)}-y^{(i)})\big|\big|_2^2, \forall \: x^{(i)} \supseteq x^{(j)},
    \end{align}
    where we transform the inequality constraint in Eq. \eqref{eq: objective_3} to minimize $\rVert\max(0, y^{(j)}-y^{(i)})\big|\big|_2^2$, and $\lambda > 0$ denotes regularization hyperparameter. 
    
    \textbf{Objective Function.} Therefore, the training procedure of the proposed SL-VAE model has been transformed into a new problem with much simpler constraints as shown in Eq. \eqref{eq: objective_4}.
    \begin{align}\label{eq: objective_5}
        \mathcal{L}_{train} = \min_{\psi, \phi, \theta} \Big[&-\mathbb{E}_{q_{\phi}}\big[\log\: p_{\psi}(y|x,G)+\log\: p_{\theta}(x|z)\big]\\ 
        &\hspace{-4.5em}+KL\big[q_{\phi}(z|x)\big|\big| p(z)\big] + \lambda\cdot \big|\big|\max(0, y^{(j)}-y^{(i)})\big|\big|_2^2 \Big], \forall \: x^{(i)} \supseteq x^{(j)}\nonumber
    \end{align}
    where we sample a $x^{(i)}$ and many $x^{(j)}$'s (such that $x^{(i)} \supseteq x^{(j)}$) as training samples for each mini-batch, and the corresponding $y^{(i)}$ and $y^{(j)}$'s can be estimated by arbitrary diffusion patterns. 
    
    To sum up, SL-VAE provides an end-to-end Bayesian inference framework such that we could integrate the propagation estimation scheme into the VAE training. This unified learning framework allows us to reconstruct the diffusion sources under any diffusion patterns as long as the forward diffusion estimation model $p_{\psi}(y|x, G)$ can successfully characterize the diffusion pattern. Therefore, Challenge $3$ is addressed. In this work, we adopt Multi-layer Perceptron (MLP) structures for both encoder $q_{\phi}(z|x)$ and decoder $p_{\theta}(x|z)$ in our training. The choice of the forward diffusion model $p_{\psi}(y|x, G)$ can be arbitrary as long as it is differentiable, which we will discuss the different choices of $p_{\psi}(y|x, G)$ in the experiment section.

    \subsection{Prediction of Diffusion Source}
    After the training phase, we aim to predict the optimal diffusion source $\Tilde x$ given the diffused observation $y$. Since the distribution $p(x)$ is directly determined by $p(z)$ (often modeled as standard Gaussian distribution) such that $p(x)=\sum_{z}p_{\theta}(x|z)\cdot p(z)$, one can sample $\Tilde x \sim p(x)$ by marginalizing $p(z)$ to perform MAP. Therefore, we derive the following optimization problem w.r.t. $x$ for finding the optimal diffusion source given the diffused observation: 
    \begin{align}\label{eq: infer_1}
    \max_x\: \big[&p_{\psi}(y|x, G)\cdot \sum_{z}p_{\theta}(x|z)\cdot p(z)\big].
    \end{align}
    However, marginalizing $p(z)$ requires to sample as many samples as possible in order to make the distribution of the sample matches the desired distribution, which burdens the computational complexity. On the other hand, Eq. \eqref{eq: infer_1} does not contain information of the diffused observation $y$, yet the diffused observation is the ingredient we would like to grasp when inferring the diffusion source.
    
    \begin{remark}[\textbf{Replacing the prior by posterior}]
    When posterior $q_{\phi}(x)$ is not collapsed, the latent random variable $z$ is sampled from $q_{\phi}(z|x) = \mathcal{N}(\mu, \sigma^2)$, where both $\mu$ and $\sigma$ are obtained through stable functions of $x$ in the training set. In other words, encoder can distill useful information from $x$ into $\mu$ and $\sigma$. If VAE can well approximate the posterior $q_{\phi}(z|x)$ to match the prior $p(z)$, we could sample $z$ from the posterior distribution $q_{\phi}(z|x)$ instead of $p(z)$.
    \end{remark}
    

    Hence, we could derive the objective function $\mathcal{L}_{pred}(x)$ w.r.t. $x$ by extending Eq. \eqref{eq: infer_1} as follows:
    \begin{align}\label{eq: infer_2}
        \mathcal{L}_{pred} &= \max_x \big[
        p_{\psi}(y|x, G) \cdot \sum_{z} \sum_{\hat x} p_{\theta}(x|z)\cdot q_{\phi}(z|\hat x)\big]\\
        &\hspace{-1em}= \min_x \Big[-\log p_{\psi}(y|x, G) - \log \big[\sum_{z} \sum_{\hat x} p_{\theta}(x|z)\cdot q_{\phi}(z|\hat x)\big]\Big],\nonumber
    \end{align}
    where $\hat x$ denotes diffusion sources from the training set. Therefore, the optimal $\Tilde x$ can be obtained by maximizing Eq. \eqref{eq: infer_2}.
    
    \begin{algorithm}[t]
    \caption{SL-VAE Inference Framework}\label{algo: inference_framework}
    \begin{algorithmic}[1]
    \REQUIRE $p_{\psi}(y|x, G)$; $p_{\theta}(x|z)$; $q_{\phi}(z|x)$; $\mathcal{L}_{pred}$; $\mathcal{L}_{init}$; Numbers of iterations: $n_{init}<n_{opt}$; Success probability $\tau$; Threshold $\delta$; Learning rate $\alpha$
    \STATE $x \sim Bin(|V|, \tau)$ \tcp{Sampling an initial $\Tilde x_{init}$.}
    \tcp{Optimize $x$ with $\mathcal{L}_{init}$ for initialization.}
    \FOR {$n=0, ..., n_{init}$}
        \STATE $x \leftarrow x - \alpha\cdot \nabla \mathcal{L}_{init}$
        \STATE $x \leftarrow trim(x, 0, 1)$
        \STATE $x \leftarrow threshold(x, \delta)$
    \ENDFOR\\
    \tcp{Optimize $x$ with $\mathcal{L}_{pred}(x)$ for final prediction.}
    \FOR {$n=0, ..., n_{opt}$}
        \STATE $x \leftarrow x - \alpha\cdot \nabla \mathcal{L}_{pred}(x)$
        \STATE $x \leftarrow trim(x, 0, 1)$
        \STATE $x \leftarrow threshold(x, \delta)$
    \ENDFOR
    \STATE $\Tilde x \leftarrow x$   \tcp{Output the predicted diffusion source.}
    \end{algorithmic}
    \end{algorithm}

    \begin{remark}[\textbf{Initialization of $x$}]
    In order to reduce the search space when optimizing Eq. \eqref{eq: infer_2}, rather than randomly initializing the input (e.g., $x \sim Bin(|V|, \tau)$, where $Bin$ is the binomial distribution and $\tau$ is a random probability of $x_i$ being seed indicator), we instead solve the following MAP problem that leverages prior knowledge of the observed diffusion sources $\hat x$ to infer a starting $x$ for Eq. \eqref{eq: infer_2}.
    \begin{equation}\label{eq: init}
            \mathcal{L}_{init} = \min_x \big[-\log p_{\psi}(y|x,G) - \log p_{\theta}(x|\bar z)\big],
    \end{equation}
    where $\bar z = \frac{1}{N} q_{\phi}(z|\hat x)$ is taken as the mean of all latent variables obtained from $N$ diffusion sources in the training set.
    \end{remark}

    The overall inference framework is summarized in Algorithm \ref{algo: inference_framework}. Specifically, we firstly sample an initial diffusion source $x$ from a Binomial distribution ($\tau = 0.5$) on Line $1$. From Line $2$ - $6$, we iteratively solve the optimization problem proposed in Eq. \eqref{eq: init} via gradient descent optimizer (e.g., Adam). In order to ensure the validity of the predicted $x$ during optimization, we regulate the value of $x$ in the range of $[0, 1]$ by using $trim(x, 0, 1)$ on Line $4$, and transform the continuous value of $x$ back to discrete (i.e., from $[0, 1]^{|V|}$ to $\{0, 1\}^{|V|}$) through the threshold $\delta$ on Line $5$. The updated $x$ is then fed as input to solve Eq. \eqref{eq: infer_2} in order to obtain the optimal $\Tilde x$ to the observation $y$ as described from Line $7$ - $11$.
    
    \textbf{Simplifying the Objective for Prediction.} Since the diffused observation $y$ fits the Gaussian distribution (i.e., continuous in $[0, 1]$) to indicate the probability of a node being infected, and the diffusion source fits the Bernoulli distribution to identify which node is a seed indicator. Therefore, we can derive the following function for optimizing $x$ with a diffused observation $y$ as:
    \begin{equation*}
        \mathcal{L}_{pred}=\min_{x} \Big[\norm{y-p_{\psi}(y|x,G)}_2^2 - \log \big[\sum_j^N \prod_i^{|V|} f_{\theta}(z^{j}_{i})^{ x_{i}}(1-f_{\theta}(z^{j}_{i}))^{(1- x_{i})} \big]\Big],
    \end{equation*}
    where $f_{\theta}$ denotes the decoder in the VAE, N and |V| are the number of training samples and nodes in the graph, respectively. For each $x_i \in x$, there is a corresponding latent variable $z_i$. $f_{\theta}(z_i) \in [0, 1]$ quantifies the probability of the single node $x_i$ being the diffusion source. The derivation of above equation is provided in Appendix.

    \begin{figure*}[!t]
		\subfloat[Jazz]{\label{fig: q1_jazz}
			\hspace{-3mm}\includegraphics[width=0.2\textwidth]{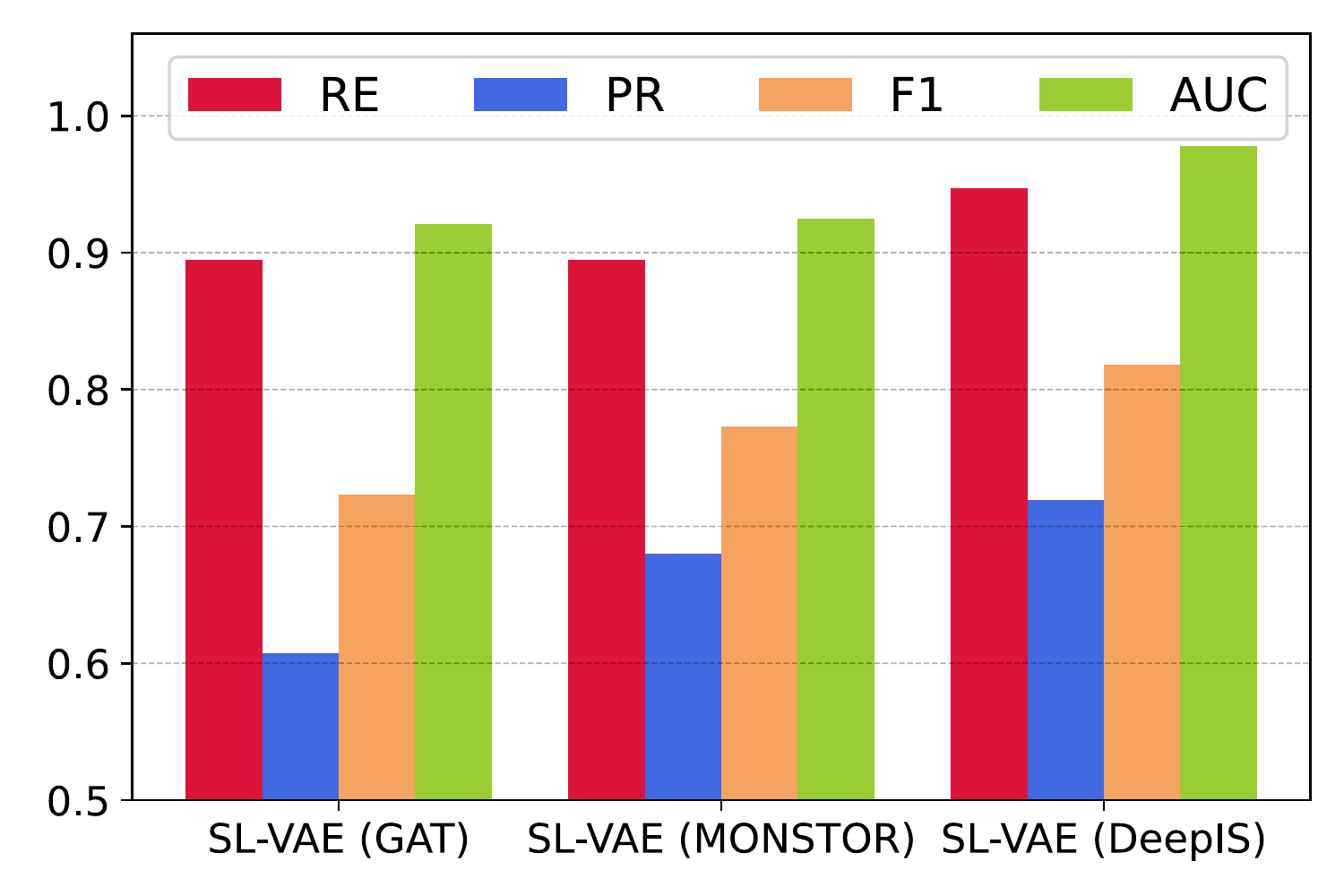}}
		\subfloat[Cora\_ML]{\label{fig: q1_cora}
			\includegraphics[width=0.2\textwidth]{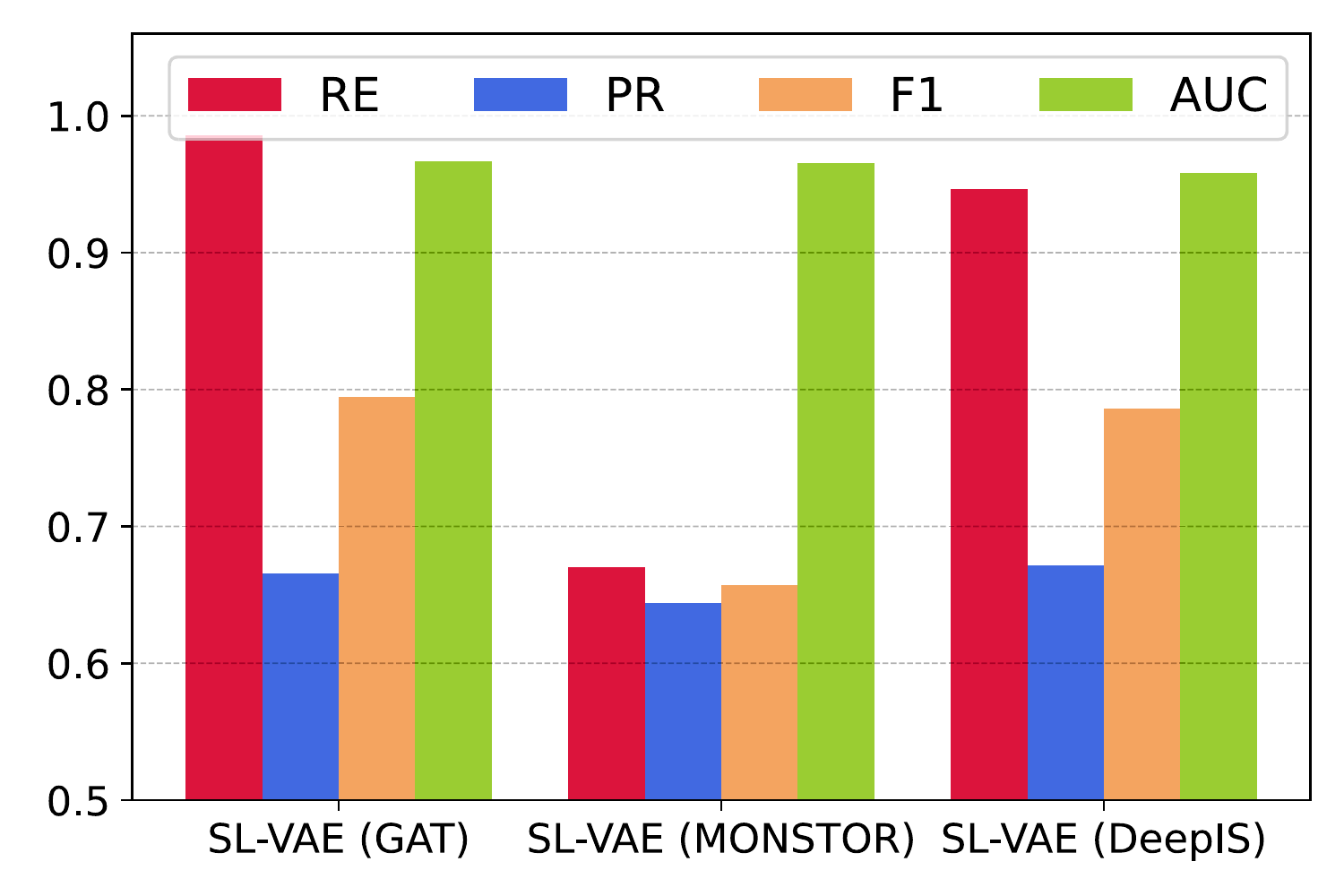}}
		\subfloat[Power Grid]{\label{fig: q1_power}
			\includegraphics[width=0.2\textwidth]{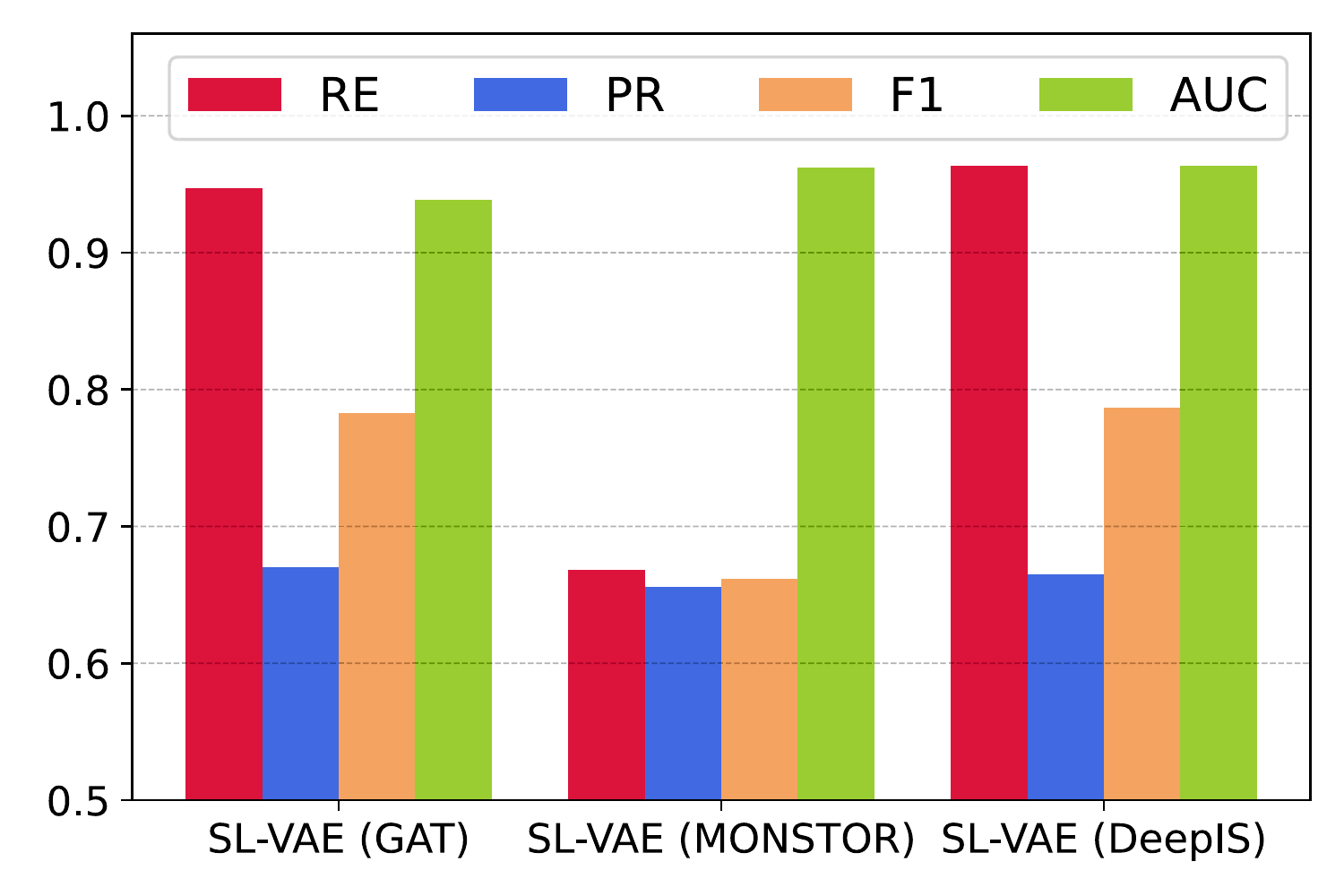}}
		\subfloat[Karate]{\label{fig: q1_karate}
			\includegraphics[width=0.2\textwidth]{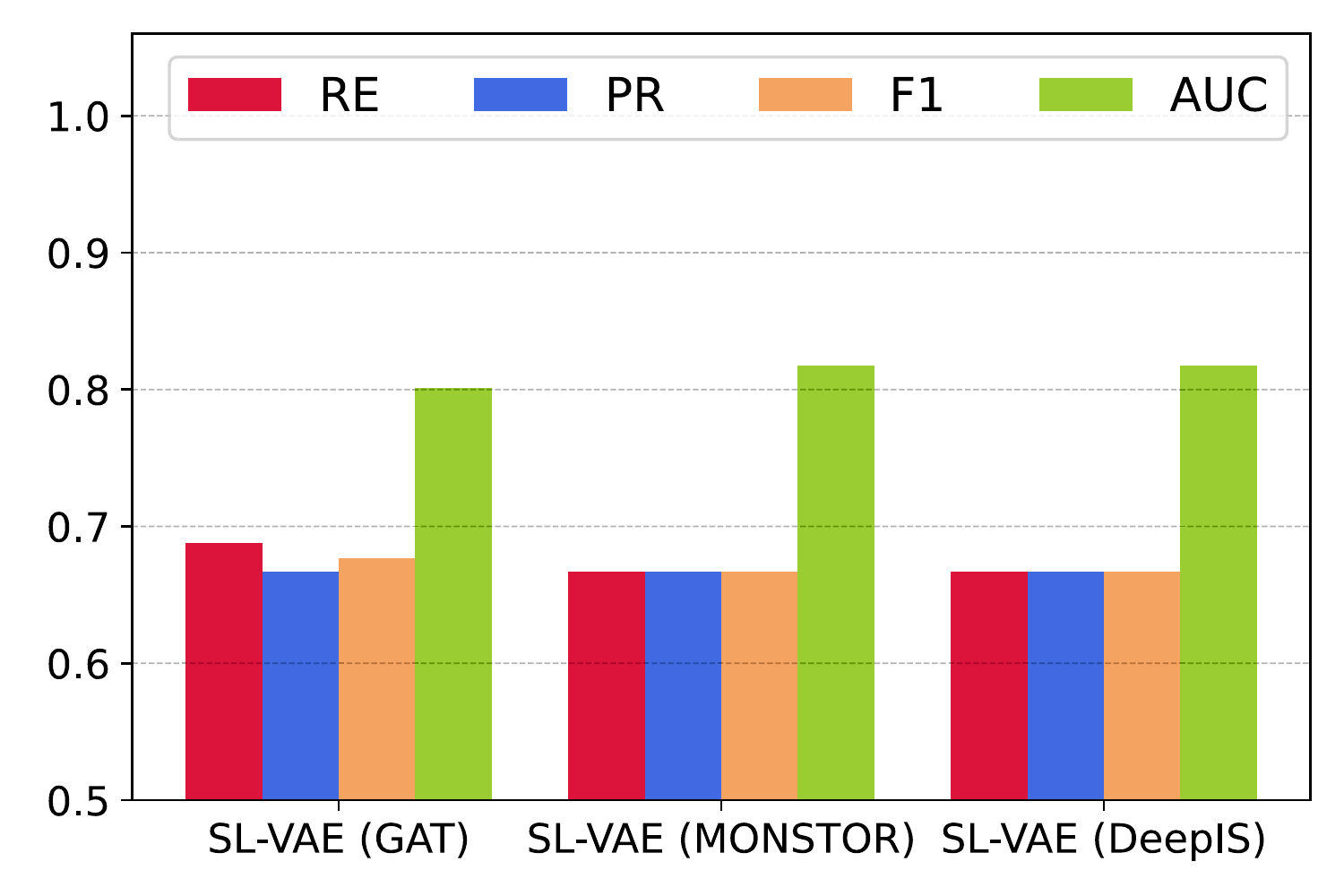}}
		\subfloat[Network Science]{\label{fig: q1_net}
			\includegraphics[width=0.2\textwidth]{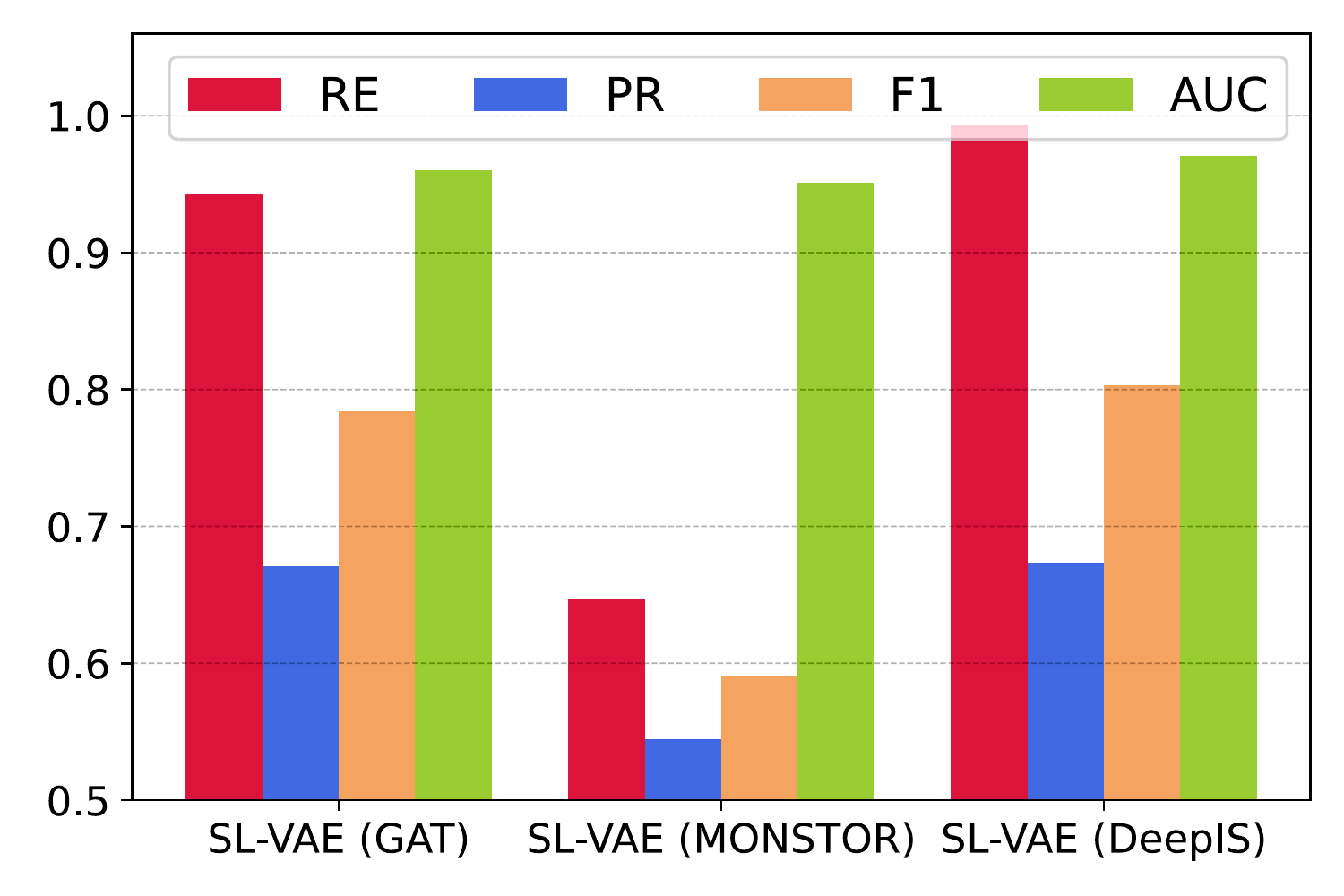}}
		\vspace{-4mm}
		\caption{The performance of SL-VAE with different forward models under the SI diffusion pattern.}
		\vspace{-4mm}
		\label{fig: q1}
	\end{figure*}
    
    \section{Experiment}\label{sec: exp}
    For this study, we utilize real-world datasets to evaluate our proposed model for answering the following questions:
    \begin{itemize}[leftmargin=*]
        \item \textbf{Q1. Flexibility}: How accurate does SL-VAE perform in source localization task when equipped with various forward diffusion estimation models?
        \item \textbf{Q2. Accuracy and Adaptation}: How does SL-VAE perform comparing with other source localization methods under different diffusion patterns (e.g., SI, SIR, Real-world Scenarios)? 
        \item \textbf{Q3. Ablation Study}: How does each component of SL-VAE contribute to the overall system?
        \item \textbf{Q4. Scalability}: How rapid does the training/inference time of SL-VAE grow as the size of the input graph increase comparing with other methods?
    \end{itemize}

    \subsection{Experiment Setting}
    \textbf{Data.} We compare our proposed SL-VAE with other baseline methods over $7$ real-world datasets. A more detailed data description and statistics can be found in Appendix. For datasets \textit{Karate}, \textit{Jazz}, \textit{Cora-ML}, \textit{Power Grid}, and \textit{Network Science} that do not have information diffusion propagation but only graph topology, we randomly choose $10\%$ of nodes as the diffusion sources and simulate the information propagation based on the SI and SIR epidemic model with $200$ iterations until convergence. Both susceptible (S) and recovered (R) nodes are regarded as uninfected nodes ($=0$), and the rest of nodes are infected ($=1$). Both \textit{Digg} and \textit{Memetracker} are composed of information propagation cascades. For each information cascade, we choose top $5\%$ of nodes and bottom $30\%$ of nodes as diffusion sources and infected nodes based on their infected time, respectively. We randomly sample two graphs for both real-world datasets so that we could demonstrate the performance comparison in terms of the growth of network size. 
    
\textbf{Comparison Methods.} 
We illustrate the performance of SL-VAE on various experiments against two sets of methods. 
\begin{itemize}[leftmargin=*]
    \item \emph{Diffusion Estimation Models.} Since SL-VAE can be utilized as a plug-in to any diffusion estimation methods, we select three state-of-the-art diffusion estimation methods as the forward function $p_{\psi}(y|x, G)$ in the SL-VAE framework. \emph{1).} GAT~\cite{velivckovic2017graph} uses the attention mechanism to calculate aggregation weights between nodes, which demonstrates its superiority in estimating the information propagation among other GNNs.  \emph{2).}  MONSTOR~\cite{ko2020monstor} estimates the total influence spread of a diffusion source by stacking multiple GCN models. \emph{3).}  DeepIS~\cite{xia2021deepis} combines GNN structures and characteristics of the influence diffusion model to estimate the susceptibility of each node.
    \item \emph{Source Localization Methods.} In terms of the performance accuracy for the network source localization task, we compare SL-VAE with three baselines. \textit{1).} NetSleuth~\cite{prakash2012spotting} aims at identifying multiple diffusion sources in a network; however, it only works when the underlying information diffusion pattern follows the SI model. \textit{2).} LPSI~\cite{wang2017multiple} propagates infected node information in the network and predicts the rumor sources based on the convergent node labels without the requirement of knowing the underlying information propagation model. \textit{3).} OJC~\cite{zhu2017catch} aims at locating sources in networks with partial observations, which has strength in detecting network sources under the SIR diffusion pattern.   \textit{4).} GCNSI \cite{dong2019multiple} learns latent node embedding with GCN to identify multiple rumor sources close to the actual source.
    \end{itemize}
    For non-learning-based source localization methods (e.g., LPSI, OJC, and NetSleuth), we set their hyper-parameters accordingly based on their original papers. For deep learning based GCNSI, we tune the hyper-parameter for their best performance in each dataset. All experiments are repeated for $10$ times for performance convergence, and we report their average score of each metric.
    
    \textbf{Implementation Details.} We leverage $3$-layers MLP with non-linear transformation for both decoder $p_{\theta}(x|z)$ and encoder $q_{\phi}(z|x)$ in the proposed SL-VAE. In addition, the choice of the forward model $p_{\psi}(y|x, G)$ can vary. We particularly choose multiple state-of-the-art information propagation models, such as Graph Attention Network (GAT)~\cite{velivckovic2017graph}, MONSTOR~\cite{ko2020monstor}, and DeepIS~\cite{xia2021deepis}. For GAT, we set the number of attention heads and the dimension of each attention channel to $8$. For MONSTOR, We set the number of GNN stacks to $3$, and for each stack, we adopt a $2$-layer GCN network. For DeepIS, we follow their original setting: a $2$-layer MLP network as the aggregation network and number of hidden units is set to be $64$. The learning rate is set to be $0.002$, and the number of epochs is set to $1,000$ for all datasets. The iteration numbers are set to $n_{init}=20, n_{opt}=50$ in Algorithm \ref{algo: inference_framework} uniformly for all dataset.

\begin{table*}[t]
\centering
\resizebox{0.96\textwidth}{!}{%
\begin{tabular}{@{}c|cccc|cccc|cccc|cccc|cccc@{}}
\toprule
        & \multicolumn{4}{c|}{Jazz}         & \multicolumn{4}{c|}{Cora-ML}      & \multicolumn{4}{c|}{Power Grid}   & \multicolumn{4}{c|}{Karate} &  \multicolumn{4}{c}{Network Science} \\ \midrule
Methods & RE     & PR     & F1     & AUC    & RE     & PR     & F1     & AUC    & RE     & PR     & F1     & AUC    & RE     & PR      & F1     & AUC    & RE     & PR      & F1     & AUC\\ \midrule
LPSI    & 0.4789 & 0.1054 & 0.1716 & 0.4841 & 0.5954 & 0.1556 & 0.2466 & 0.6675 & 0.4953 & 0.4546 & 0.4737 & 0.9337 & 0.4667 & 0.1861  & 0.2855 & 0.6344 & 0.6044  & 0.4231  & 0.4978 & 0.8378 \\
GCNSI   & 0.4368 & 0.1589 & 0.2329 & 0.6428 & 0.3619 & 0.1182 & 0.1780 & 0.5383 & 0.3477 & 0.1413 & 0.2099 & 0.5044 & 0.4333 & 0.1999 & 0.2613 & 0.6022 & 0.2247 & 0.1375 & 0.1706 & 0.4759 \\
OJC     & 0.1798 & 0.1005 & 0.1289 & 0.5045 & 0.2239 & 0.2036 & 0.2133 & 0.5633 & 0.2871 & 0.1044 & 0.1531  & 0.5011 & 0.3611 & 0.2708 & 0.3095 & 0.6335 & 0.1233 & 0.3708 & 0.1851 & 0.5331 \\
Netsleuth & 0.1315 & 0.1087 & 0.1191 & 0.5432 & 0.2647 & 0.2647 & 0.2647 & 0.4688 & 0.5972 & 0.4975 & 0.5428 & 0.7651 & 0.3333 & 0.3333 & 0.3333 & 0.4355 & 0.3948 & 0.3283 & 0.3585 & 0.6528 \\ 
SL-VAE   & $\mathbf{0.9474}$ & $\mathbf{0.7193}$ & $\mathbf{0.8182}$ & $\mathbf{0.9777}$ & $\mathbf{0.9466}$ & $\mathbf{0.6717}$ & $\mathbf{0.7858}$ & $\mathbf{0.9582}$ & $\mathbf{0.9636}$ & $\mathbf{0.6648} $ & $\mathbf{0.7868}$ & $\mathbf{0.9636}$ & $\mathbf{0.6667}$  & $\mathbf{0.6667}$ & $\mathbf{0.6667}$ & $\mathbf{0.8172}$ & $\mathbf{0.9937}$ & $\mathbf{0.6738}$ & $\mathbf{0.8031}$ & $\mathbf{0.9705}$ \\\bottomrule
\end{tabular}
}
\caption{Performance over comparison methods under SI diffusion pattern. (Best is highlighted with bold.)}
\vspace{-6mm}
\label{tab: evaluation_1}
\end{table*}
    
    \begin{table*}[t]
    \centering
    \resizebox{0.96\textwidth}{!}{%
    \begin{tabular}{@{}c|cccc|cccc|cccc|cccc|cccc@{}}
    \toprule
        & \multicolumn{4}{c|}{Jazz}         & \multicolumn{4}{c|}{Cora-ML}      & \multicolumn{4}{c|}{Power Grid}   & \multicolumn{4}{c|}{Karate} &  \multicolumn{4}{c}{Network Science} \\ \midrule
    Methods & PR     & RE     & F1     & AUC    & PR     & RE     & F1     & AUC    & PR     & RE     & F1     & AUC    & PR      & RE      & F1     & AUC    & PR      & RE      & F1     & AUC\\ \midrule
    LPSI    & 0.1153 & 0.3632 & 0.1698 & 0.5005 & 0.1072 & 0.4779 & 0.1752 & 0.4986 & 0.4865 & 0.4721 & 0.4784 & 0.5821 & 0.1284  & 0.4167  & 0.1936 & 0.5144 & 0.1362  & 0.4326  & 0.2072 & 0.5614 \\
    GCNSI   & 0.1419 & 0.3737 & 0.2055 & 0.6411 & 0.1158 & 0.3381 & 0.1725 & 0.5321 & 0.1133 & 0.2371 & 0.1533 & 0.5038 & 0.0721 & 0.1167 & 0.0879 & 0.5036 & 0.1047 & 0.3511 & 0.1613 & 0.5433 \\
    OJC     & 0.1543 & 0.2201 & 0.1814 & 0.5012 & 0.1414 & 0.1679 & 0.1535 & 0.5110 & 0.1414 & 0.1679 & 0.1535 & 0.5009 & 0.3750 & 0.1944 & 0.2500 & 0.5771 & 0.3977 & 0.1231 & 0.1800 & 0.5097 \\ 
    SL-VAE   & $\mathbf{0.6667}$ & $\mathbf{0.8421}$ & $\mathbf{0.7442}$ & $\mathbf{0.9749}$ & $\mathbf{0.6695}$ & $\mathbf{0.5623}$ & $\mathbf{0.6112}$ & $\mathbf{0.9686}$ & $\mathbf{0.6846}$ & 
    $\mathbf{0.6458}$ & $\mathbf{0.6646}$ & $\mathbf{0.9689}$ & $\mathbf{0.6250}$  & $\mathbf{0.8333}$ & $\mathbf{0.7143}$ & $\mathbf{0.8289}$ & $\mathbf{0.6541}$ & $\mathbf{0.5713}$ & $\mathbf{0.6099}$ & $\mathbf{0.9711}$ \\
    \bottomrule
    \end{tabular}
    }
    \caption{Performance over comparison methods under SIR diffusion pattern. (Best is highlighted with bold.)}
    \vspace{-6mm}
    \label{tab: evaluation_2}
    \end{table*}

\begin{table*}[t]
\centering
\resizebox{0.9\textwidth}{!}{%
\begin{tabular}{@{}c|cccc|cccc|cccc|cccc@{}}
\toprule
        & \multicolumn{4}{c|}{Digg-7556}         & \multicolumn{4}{c|}{Digg}      & \multicolumn{4}{c|}{Memetracker-7884}   & \multicolumn{4}{c}{Memetracker} \\ \midrule
Methods & PR     & RE     & F1     & AUC    & PR     & RE     & F1     & AUC    & PR     & RE     & F1     & AUC    & PR      & RE      & F1     & AUC    \\ \midrule
LPSI    & 0.0026 & 0.0123 & 0.0043 & 0.4432 & 0.0079 & 0.2727 & 0.0155 & 0.5618 & 0.0132 & 0.3184 & 0.0253 & 0.5112 & 0.0087 & 0.2913 & 0.0169 & 0.5377 \\
GCNSI   & 0.0114 & 0.3700 & 0.0221 & 0.4450 &  0.0123 & 0.2100 & 0.0232 & 0.4129 & 0.0211 & 0.3219 & 0.0396 & 0.4357 & 0.0197 & 0.2342 & 0.0363 & 0.4103 \\
OJC     & 0.0118 & 0.0107 & 0.0112 & 0.5023 & 0.0635 & 0.0696 & 0.0664 & 0.5142 & 0.0542 & 0.0433 & 0.0481  & 0.4812  & 0.0331 & 0.0207 & 0.0255 & 0.5077 \\
SL-VAE   & $\mathbf{0.4131}$ & $\mathbf{0.6217}$ & $\mathbf{0.4655}$ & $\mathbf{0.5541}$ & $\mathbf{0.4297}$ & $\mathbf{0.5421}$ & $\mathbf{0.4792}$ & $\mathbf{0.6213}$ & $\mathbf{0.5113}$ & $\mathbf{0.6214}$ & $\mathbf{0.5610}$ & $\mathbf{0.5954}$ & $\mathbf{0.4612}$ & $\mathbf{0.5181}$ & $\mathbf{0.4880}$ & $\mathbf{0.6245}$ \\ \bottomrule
\end{tabular}
}
\caption{Performance evaluation over comparison methods under the real-world diffusion. (Best is highlighted with bold.)}
\vspace{-8mm}
\label{tab: evaluation_3}
\end{table*}

    \textbf{Evaluation Metrics.} 
    We use two metrics to evaluate the performance of our proposed model: \emph{1).} F1-Score (F1): since network source localization is \emph{de facto} a classification task that we need to accurately distinguish diffusion sources to all nodes, we leverage the most commonly used evaluation metrics  - F1-score for this task. F1-score is taken as the harmonic mean of classification precision (PR) and recall (RE), so we report the PR and RE score along with F1 as well.
    \emph{2).} ROC-AUC Curve (AUC): since most real-world scenarios tend to have an imbalance between the number of diffusion sources and non-source nodes (i.e., positive and negative samples), we additionally leverage the ROC-AUC curve as another evaluation metrics, which is less sensitive to the data imbalance. 
    
    \subsection{Q1: Flexibility of SL-VAE}
    To assess the flexibility of SL-VAE, we evaluate the model performance when providing various forward diffusion estimation models: GAT, MONSTOR, and DeepIS under the SI diffusion pattern. The results are visualized in Fig. \ref{fig: q1}, which are quite revealing in two aspects. Firstly, the performance of SL-VAE in recovering diffusion sources while equipped with various forward models is satisfying, the AUCs for each dataset are above $90\%$ with only one exception, and the F1 scores can also achieve $80\%$ on average. Second, the stability of SL-VAE in classifying diffusion sources is also well demonstrated. As shown in Fig. \ref{fig: q1}, we cannot observe any noticeable difference in performance between each variant of SL-VAE. In other words, as long as the information diffusion model can successfully capture the diffusion pattern, SL-VAE can be utilized as the inverse model with high flexibility to recover the diffusion sources. Hence, considering the overall performance for all datasets, we choose DeepIS as the forward model of SL-VAE and solely use SL-VAE to refer (SL-VAE + DeepIS) in the following experiments.

        \begin{figure*}[!t]
		\subfloat[GCNSI]{\label{fig: karate_gcnsi}
			\includegraphics[width=0.145\textwidth]{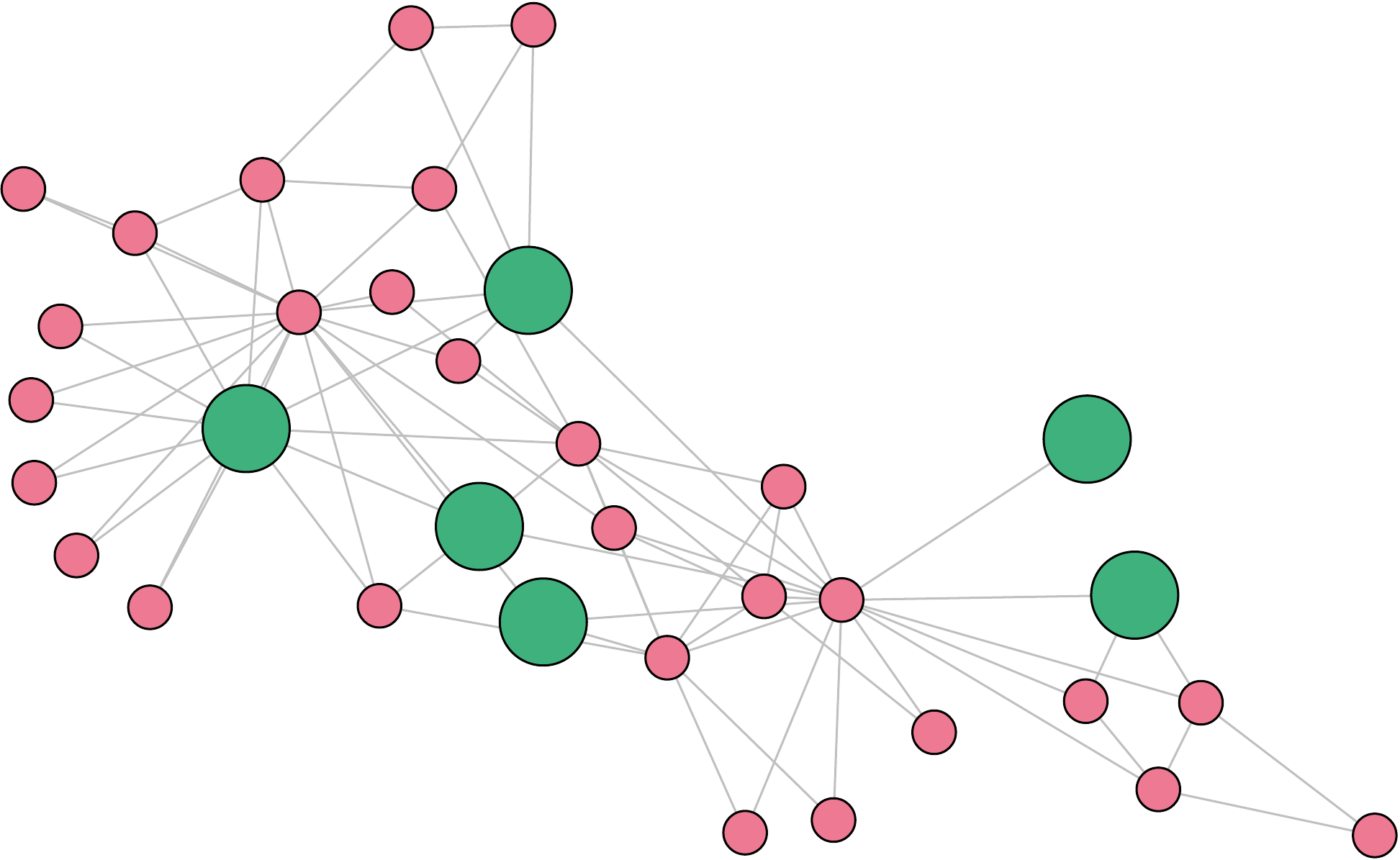}}
		\subfloat[LPSI]{\label{fig: karate_lpsi}
			\includegraphics[width=0.145\textwidth]{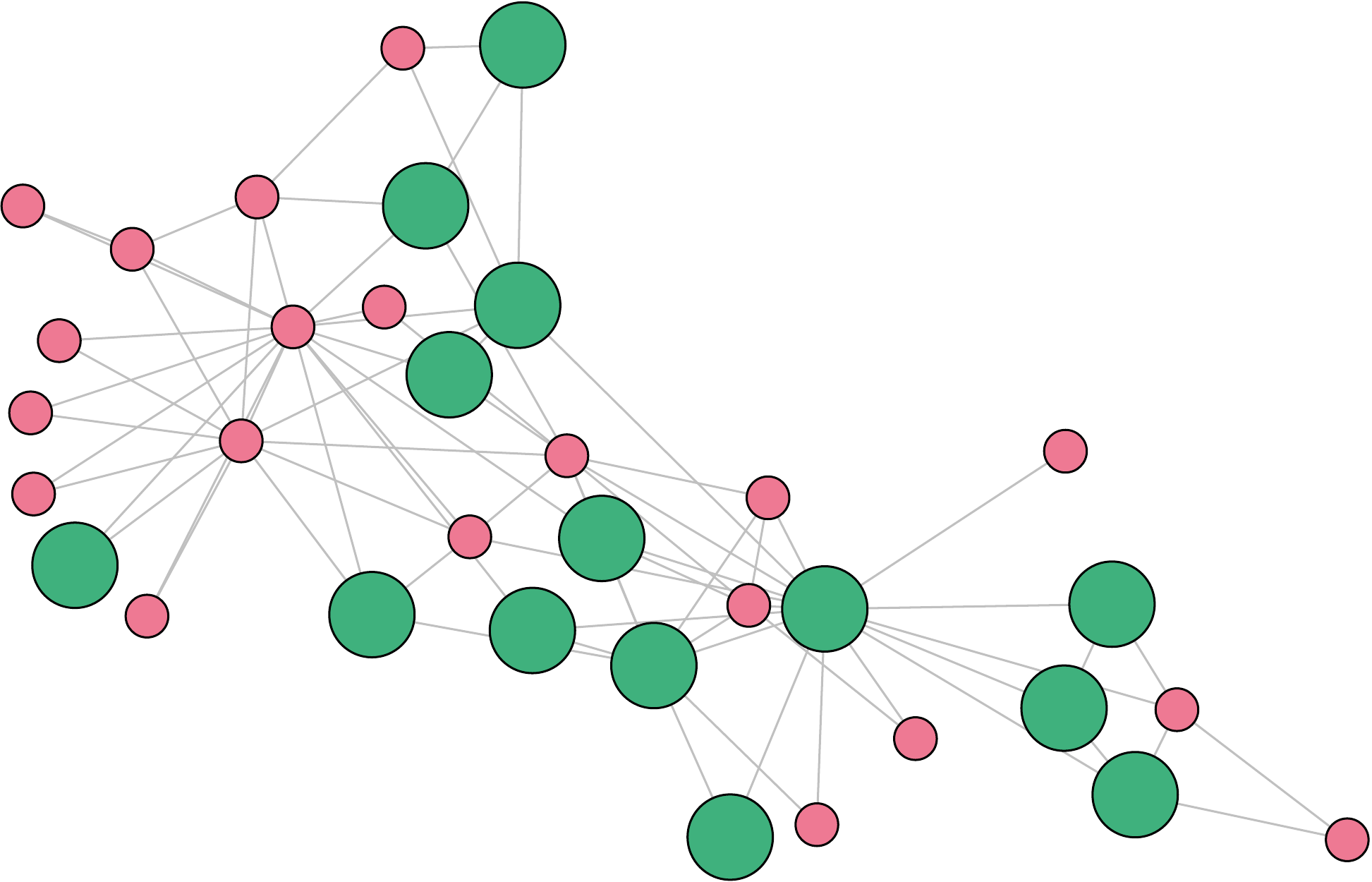}}
		\subfloat[Netsleuth]{\label{fig: karate_netslu}
			\includegraphics[width=0.145\textwidth]{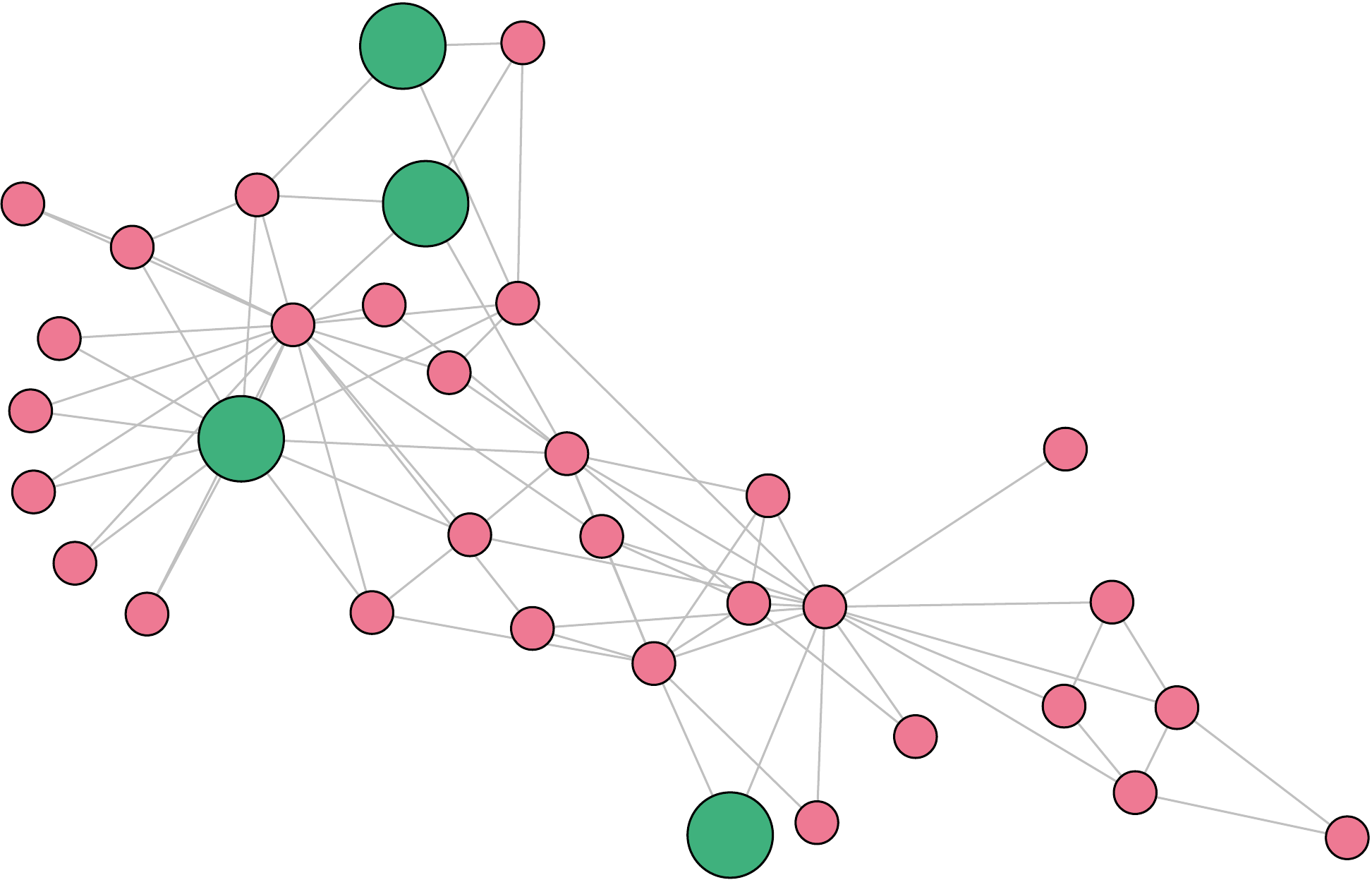}}
		\subfloat[OJC]{\label{fig: karate_ojc}
			\includegraphics[width=0.145\textwidth]{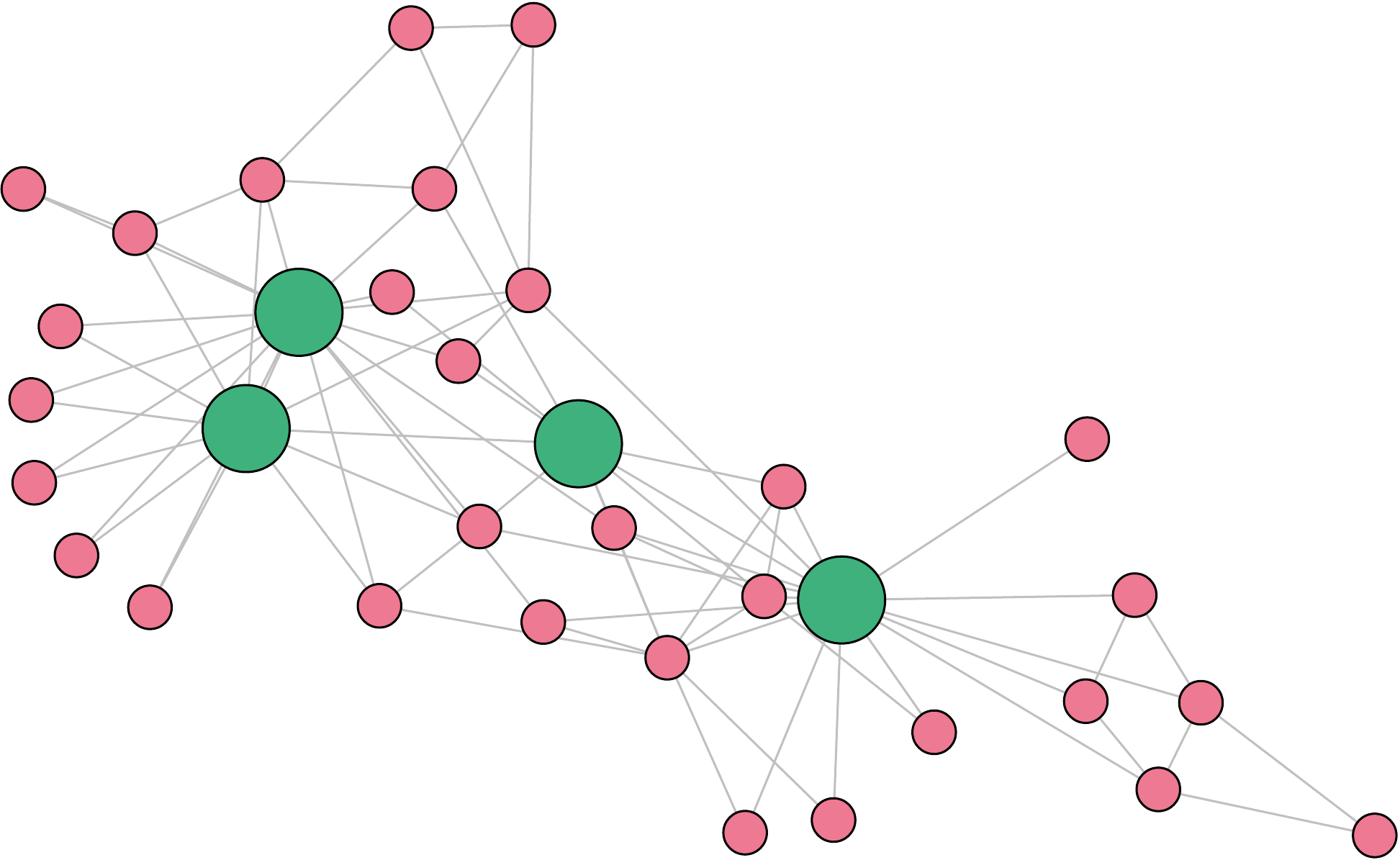}}
		\subfloat[\textbf{SL-VAE}]{\label{fig: karate_slvae}
			\includegraphics[width=0.145\textwidth]{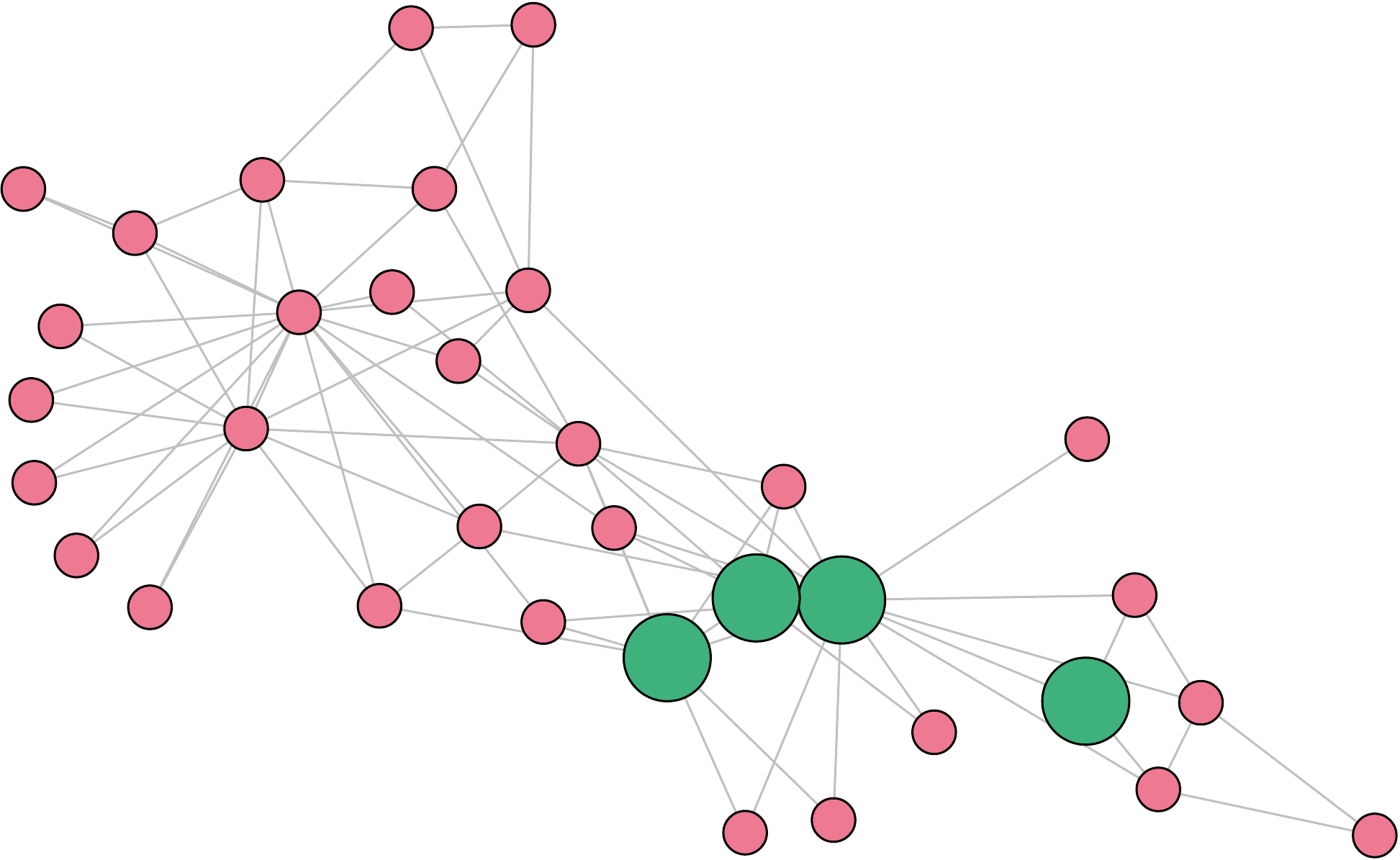}}
		\subfloat[Real]{\label{fig: karate_real}
			\includegraphics[width=0.145\textwidth]{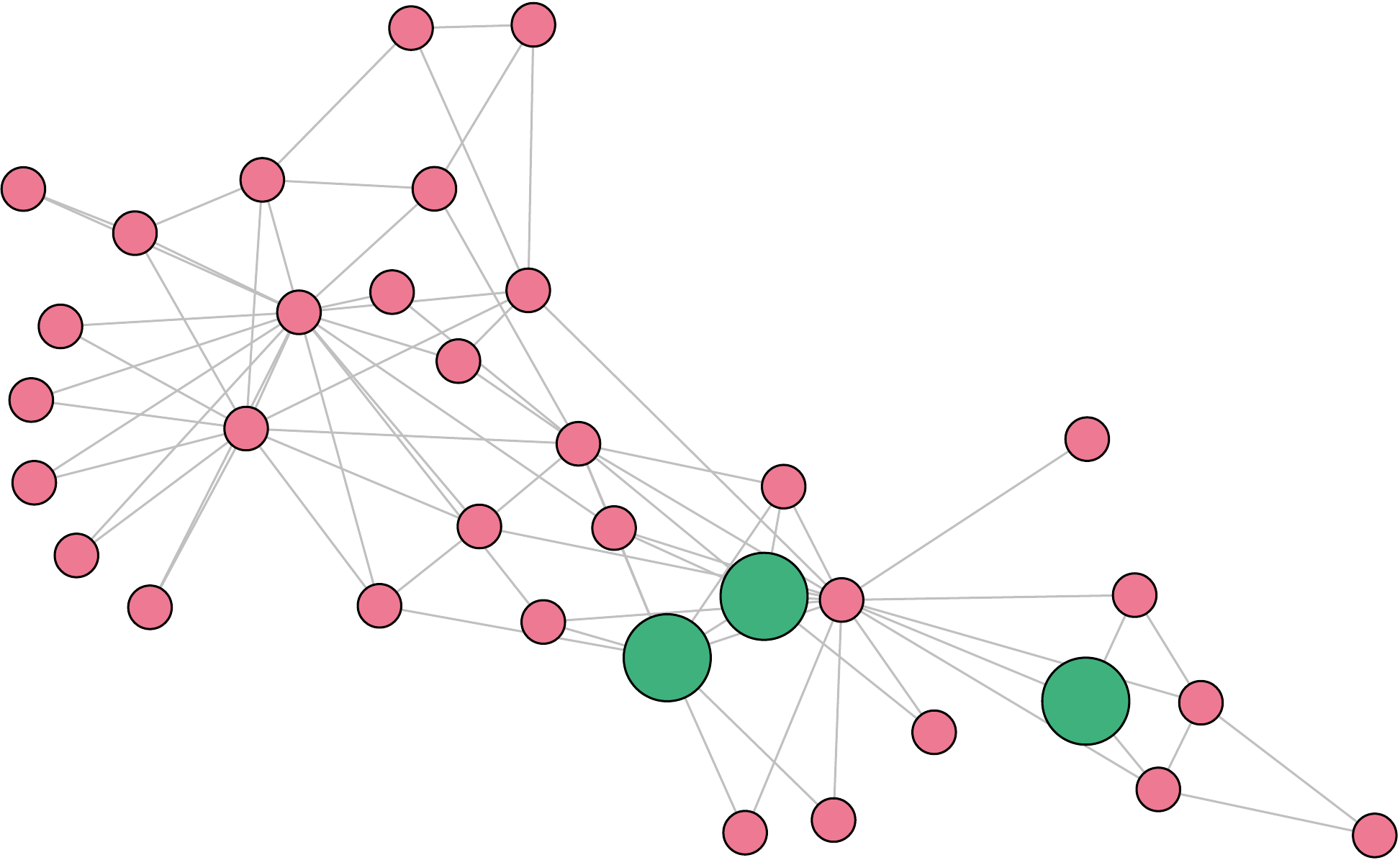}}\\\vspace{-3mm}
		\subfloat[GCNSI]{\label{fig: jazz_gcnsi}
			\includegraphics[width=0.16\textwidth]{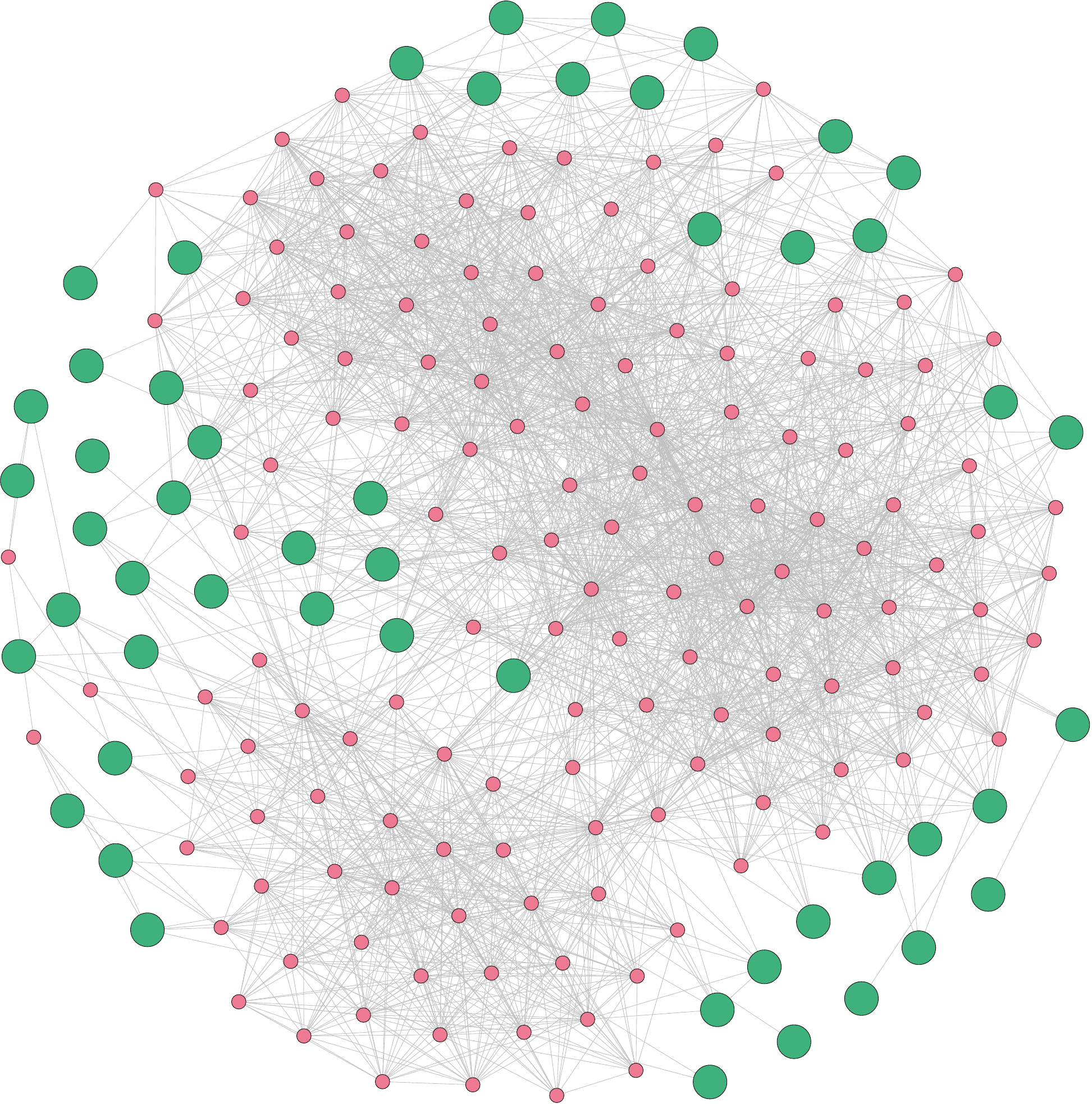}}
		\subfloat[LPSI]{\label{fig: jazz_lpsi}
			\includegraphics[width=0.16\textwidth]{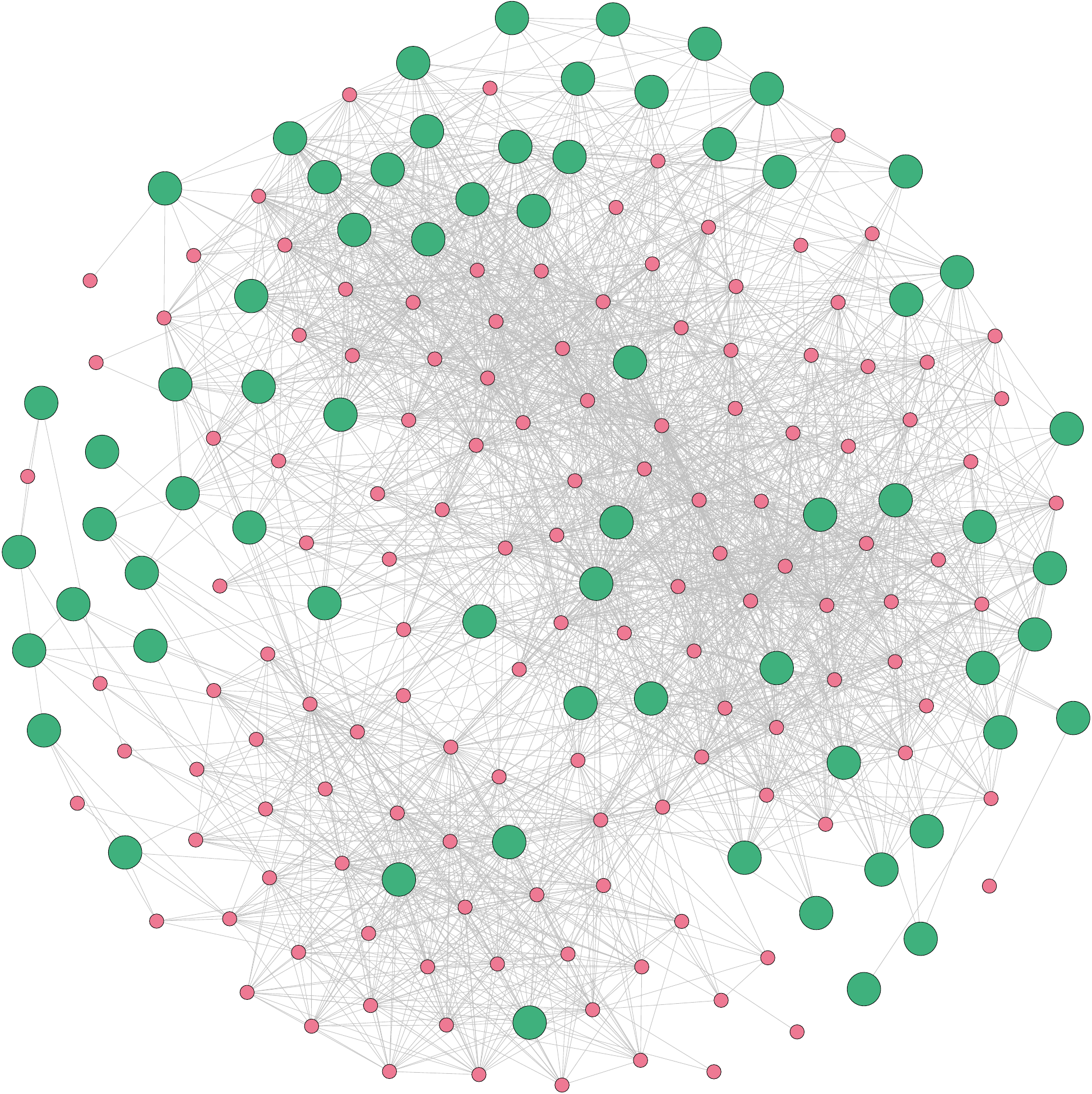}}
		\subfloat[Netsleuth]{\label{fig: jazz_netslu}
			\includegraphics[width=0.16\textwidth]{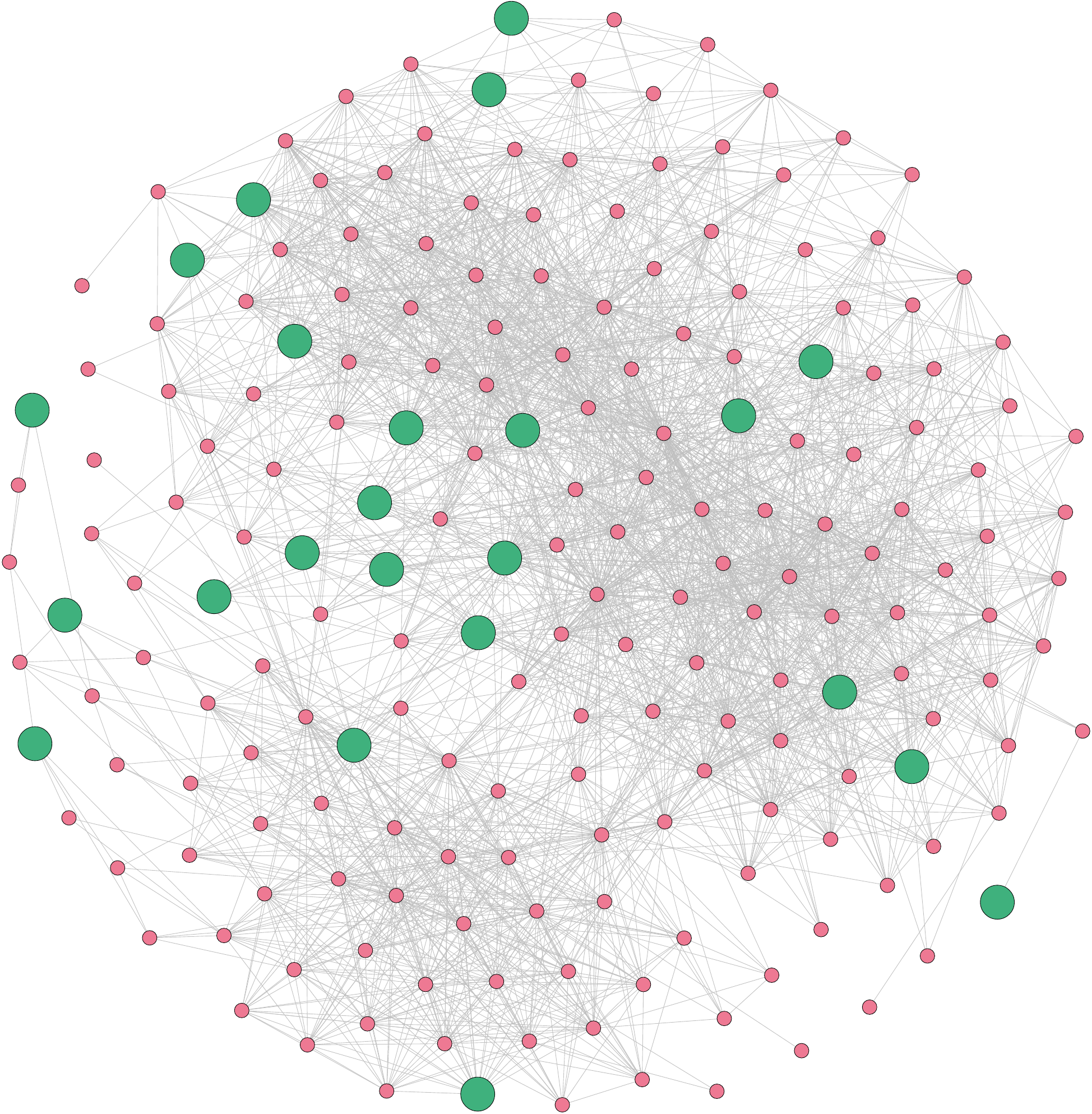}}
		\subfloat[OJC]{\label{fig: jazz_ojc}
			\includegraphics[width=0.16\textwidth]{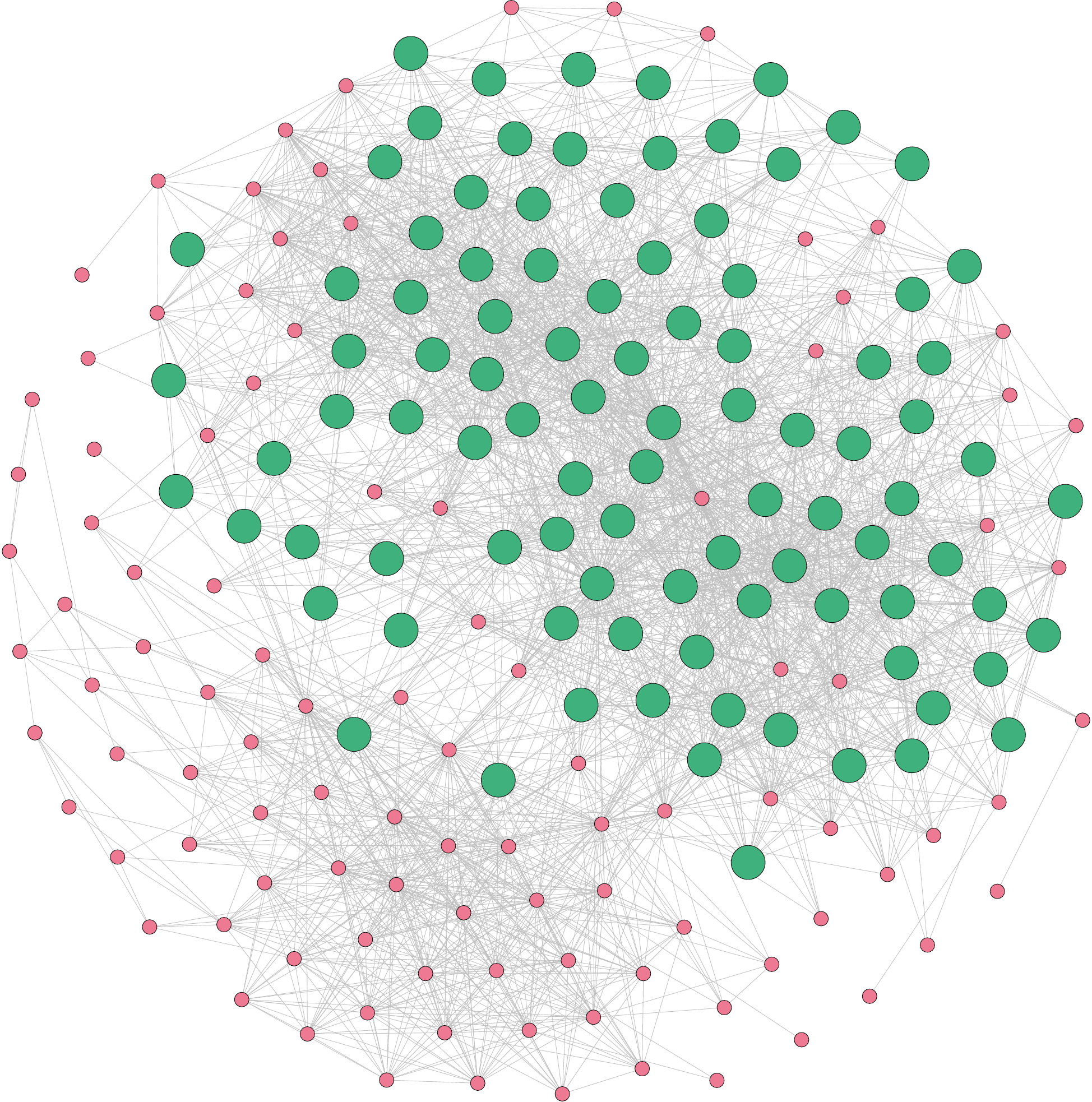}}
		\subfloat[\textbf{SL-VAE}]{\label{fig: jazz_slvae}
			\includegraphics[width=0.16\textwidth]{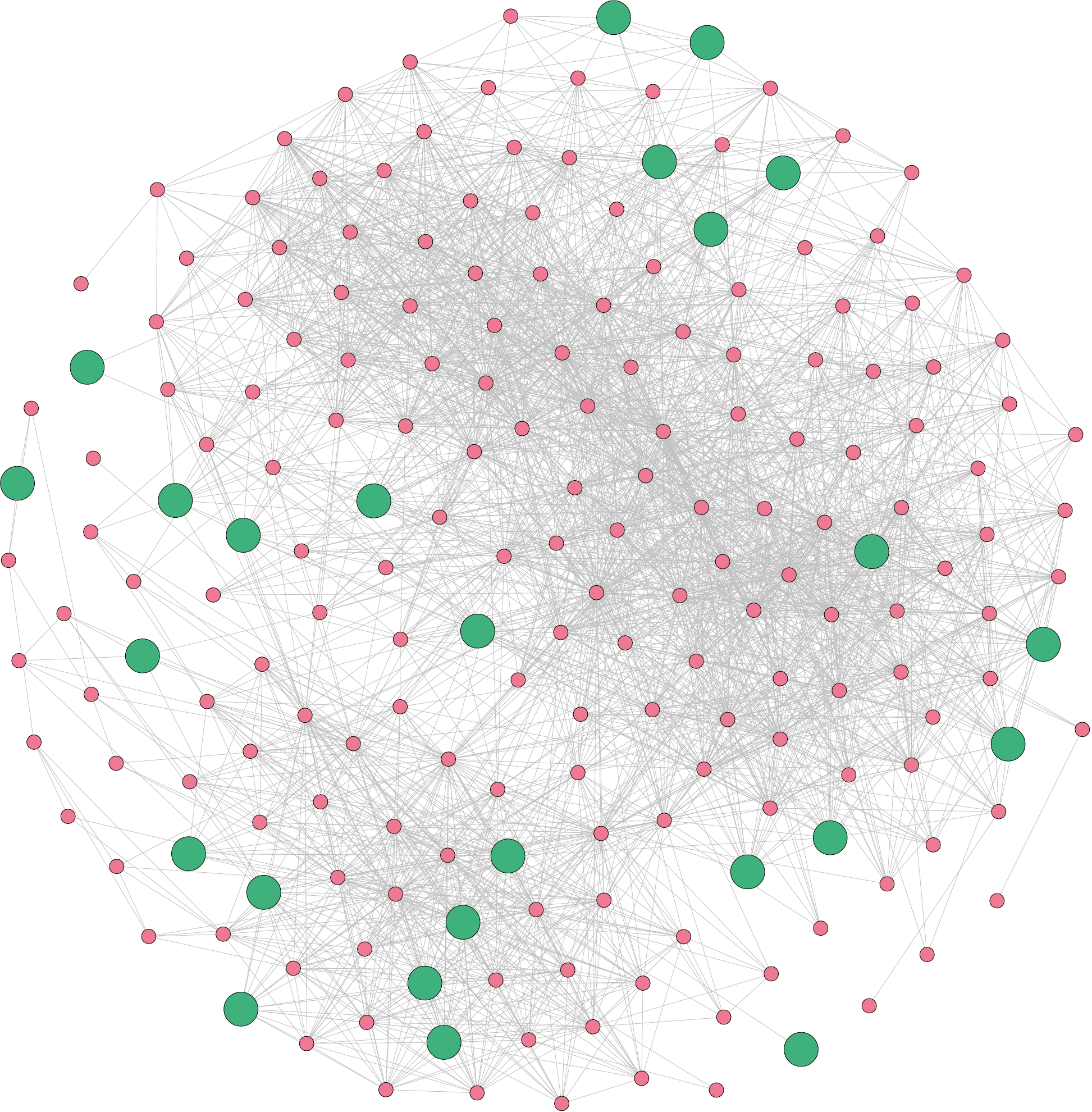}}
		\subfloat[Real]{\label{fig: jazz_real}
			\includegraphics[width=0.16\textwidth]{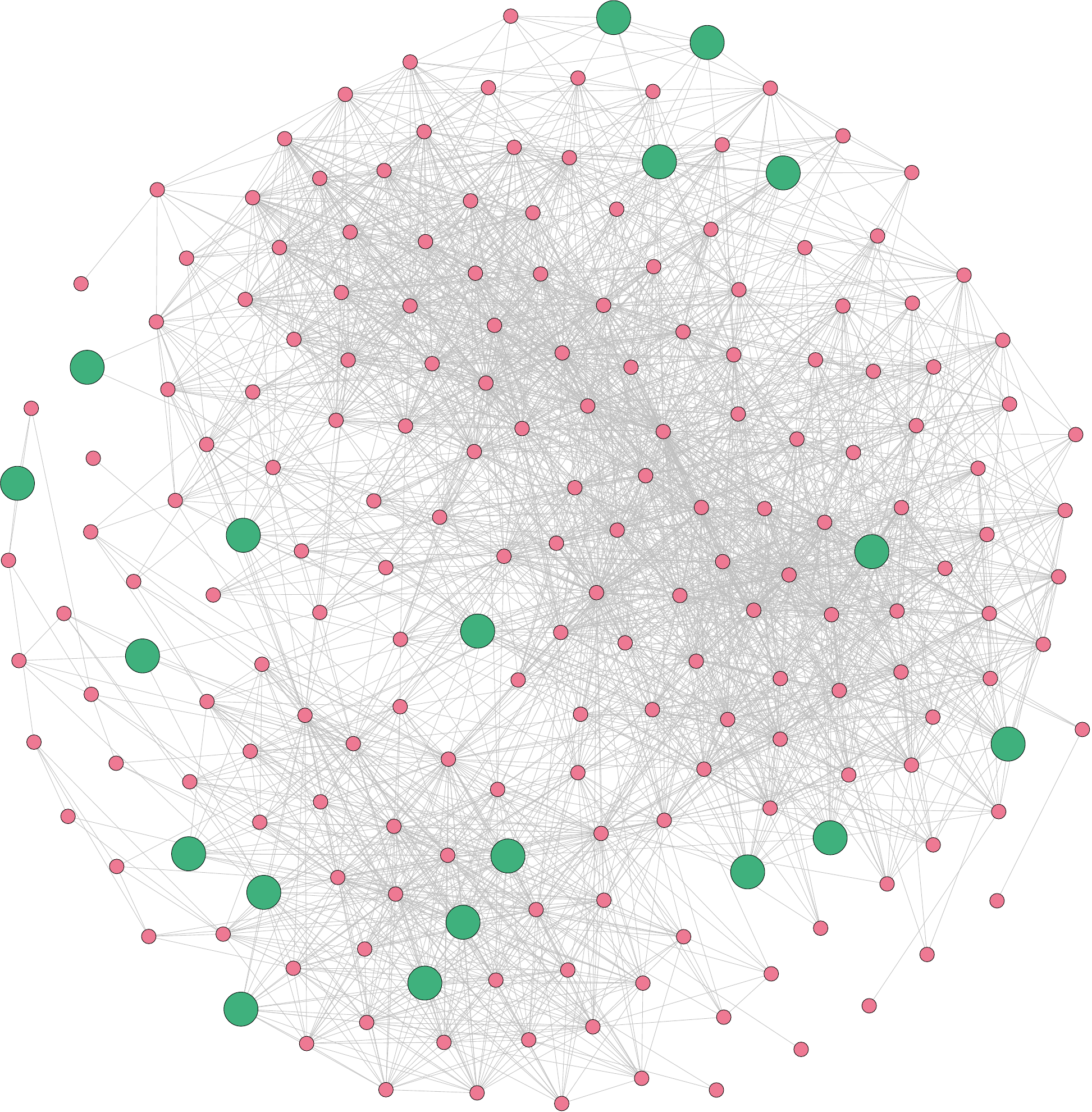}}
		\vspace{-5mm}
		\caption{The visualized comparison between the generated diffusion sources and the ground truth. Fig. \ref{fig: karate_gcnsi} - \ref{fig: karate_real} are visualizations of Karate dataset, and Fig. \ref{fig: jazz_gcnsi} - \ref{fig: jazz_real} are visualizations of Jazz dataset. The predicted and ground truth diffusion sources are marked with green color while the rest of nodes are marked with red color.}
		\label{fig: karate}
		\vspace{-4mm}
	\end{figure*}
    
    \subsection{Q2: Accuracy of SL-VAE}
    We then evaluate the performance of SL-VAE against other source localization approaches. Note that a sturdy source localization algorithm should be capable of locating diffusion sources under various diffusion patterns. Therefore, we choose two epidemic models (i.e., SI and SIR) as the underlying propagation model to test the effectiveness of SL-VAE. Since the SIS model is similar to SI and SIR model, we only adopt SI and SIR models for comparison. For comparison methods, LPSI and GCNSI are tested under both diffusion patterns. NetSleuth is only tested under the SI diffusion pattern due to its design. Similarly, OJC is particularly designed for locating sources under the SIR model, and it can be adapted to the SI model with minor modifications. We present the performance comparison of all methods in Table \ref{tab: evaluation_1} and \ref{tab: evaluation_2}, respectively.
    
    \textbf{Performance under SI Diffusion Pattern.} It is apparent from Table \ref{tab: evaluation_1} that SL-VAE excels other approaches in terms of both F1 and AUC scores. For instance, SL-VAE achieves the best result in Cora-ML and Network Science datasets by excelling other methods on average $20\%$ in each evaluation metric. Specifically, SL-VAE achieves over $0.95$ AUC score while the average AUC score of other methods is mere $0.62$. The performance of SL-VAE in the Karate dataset is not better than other datasets primarily due to the smaller node size in the Karate dataset; however, SL-VAE still achieves the best performance than others with larger scales. Although other comparison methods can achieve better performance in a single metric (e.g., LPSI delivers higher RE in several datasets and OJC scores higher PR in Power Grid dataset), their overall prediction performances are still not comparable to SL-VAE. Since the ratio of diffusion sources to other nodes is highly imbalanced and other comparison approaches cannot sufficiently capture the data distribution of the diffusion sources, other baseline methods tend to predict an inaccurate amount of source nodes. The above observation further substantiates that our proposed SL-VAE can accurately predict the diffusion sources despite the scarcity of diffusion sources by leveraging the deep generative model to learn such a prior.

    \textbf{Performance under SIR Diffusion Pattern.} Furthermore, we demonstrate the performance comparison between each baseline under the SIR diffusion pattern, and the results are shown in Table \ref{tab: evaluation_2}. Note that locating diffusion sources under the SIR diffusion pattern is much harder since the immunity of individual nodes causes the incompleted observation and brings more randomness to the source localization task. As can be seen from Table  \ref{tab: evaluation_2}, SL-VAE still achieves the best performance among all methods by at least $15\%$ better than the second-best model in terms of AUC score. Note that OJC is specially designed for detecting diffusion sources under the SIR diffusion pattern by the proposed candidate selection algorithm, yet SL-VAE still outperforms it in most evaluation metrics with a few exceptions. In addition, what is striking about SL-VAE in Table \ref{tab: evaluation_1} and \ref{tab: evaluation_2} is SL-VAE consistently predicts high-quality diffusion sources no matter what underlying diffusion pattern is.

    
    \begin{table}[t]
    \centering
    \resizebox{0.49\textwidth}{!}{%
    \begin{tabular}{@{}c|cc|cc|cc|cc|cc@{}}
    \toprule
    Datasets & \multicolumn{2}{c|}{Jazz} & \multicolumn{2}{c|}{Cora-ML} & \multicolumn{2}{c|}{Power Grid} & \multicolumn{2}{c|}{Karate} & \multicolumn{2}{c}{Network Science} \\ \midrule
      & F1          & AUC         & F1             & AUC            & F1               & AUC  & F1               & AUC & F1               & AUC              \\\midrule
    SL-VAE (a)  &       0.6254      &       0.8763      &       0.5180         &        0.8868        &         0.6692         &     0.9021    &    0.4936      &   0.7123     &    0.5459      &  0.9112   \\
        SL-VAE (b)  &      0.8072       &     0.9542        &      0.7589          &        0.9374        &     0.7124             &    0.9102     &    0.5653      &   0.7329     &    0.7264      &  0.9512   \\
    SL-VAE   & 0.8182   & 0.9777   & 0.7854      & 0.9582         & 0.7868         & 0.9636           & 0.6667    &     0.8172     &    0.8031 &        0.9705             \\ \bottomrule
    \end{tabular}%
    }
    \caption{Ablation Study}
    \vspace{-12mm}
    \label{tab: ablation}
    \end{table}
    
    \textbf{Performance under Real-world Diffusion Pattern.} All results are consistent with the notion that SL-VAE obtains the best performance among all baselines under the diffusion patterns including SI and SIR model. To demonstrate the effectiveness of SL-VAE under the real-world information diffusion pattern, we further conduct experiments on two large-scale real-world social network datasets: Digg and Memetracker, and we report the prediction performance in Table \ref{tab: evaluation_3}. We randomly subsample two sub-networks: Digg-7556 (with $|V| = 7,556$) and Memetracker-7884 (with $|V| = 7,884$) to demonstrate the performance change in terms of the growth of graph size. As can be clearly seen from the table, almost all baselines experience a performance decline, which is primarily due to two reasons: \emph{1).} node size grows in both real-world graphs significantly comparing with other datasets used in Table \ref{tab: evaluation_1} and \ref{tab: evaluation_2}, which makes the data imbalance problem more severe. \emph{2).} the underlying diffusion pattern in real-world network is intrinsically more complex than the prescribed diffusion pattern, which makes locating sources on real-world data is more challenging. However, with the aid of the learned generative prior, SL-VAE still outperforms other comparison methods by a large margin (e.g., at least $30\%$ better in F1 score and $10\%$ better in AUC score than other models).

    \subsection{Q3: Ablation Study}
    We further conduct the ablation study to investigate the importance of each component of SL-VAE. For the first ablated model, instead of optimizing the joint likelihood as shown in Eq. \eqref{eq: infer_2}, we directly maximize the likelihood of $p_{\psi}(y|x)$ while $x$ is obtained through Eq. \eqref{eq: init}. For the second ablated model, we proceed to directly optimize Eq. \eqref{eq: infer_2} without the proposed initialization step (i.e., Eq. \eqref{eq: init}). For simplicity, we denote both ablated models as SL-VAE (a) and SL-VAE (b), respectively. The performance comparison on $5$ datasets between the ablated models and the original SL-VAE under the SI diffusion pattern is shown in Table \ref{tab: ablation}. Overall, the performance will degrade if any components of our proposed SL-VAE are removed. Particularly, compared with other approaches in Table \ref{tab: evaluation_1}, SL-VAE (a) has already provided a comparable estimation of diffusion sources with the aid of the proposed initialization (Eq. \eqref{eq: init}). For example, SL-VAE (a) achieves better results in F1 score than any other methods in the Jazz, Cora-ML, and Karate dataset, and SL-VAE (a) still achieves the second-best in the rest of datasets. After the proposed joint likelihood (i.e., comparing SL-VAE (a) and (b)) is added back to the inference model, both F1 and AUC have been improved by $5-20\%$ and $2-7\%$, respectively. Notably, the performance of SL-VAE (b) achieves the best among others, which suggests that incorporating the prior distribution of the observed diffusion sources can effectively improve the prediction accuracy. 
    
\begin{figure}[!t]
    \vspace{-5mm}
		\subfloat[Different forward models]{\label{fig: running_1}
		\hspace{-3mm}\includegraphics[width=0.25\textwidth]{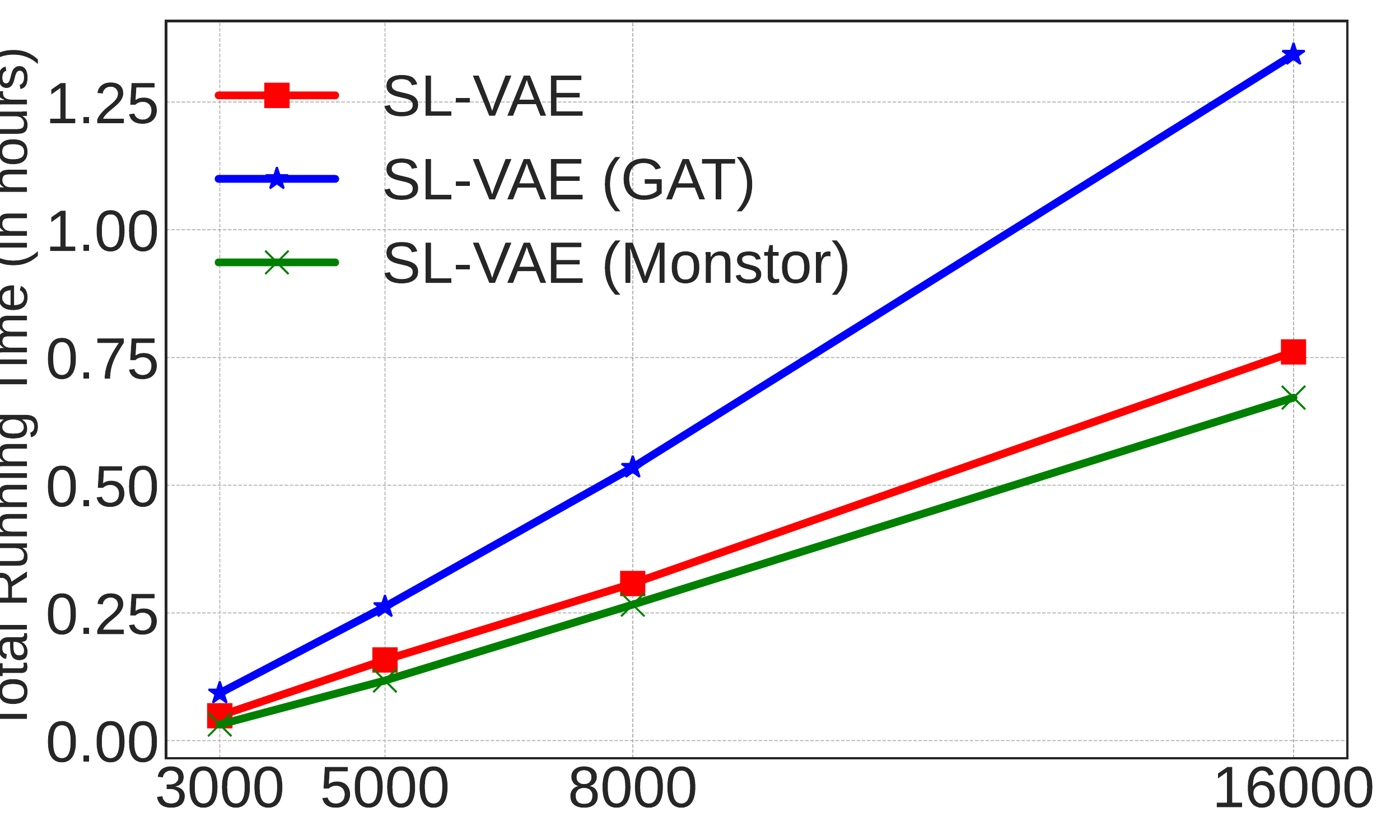}}
		\subfloat[Different comparison methods]{\label{fig: running_2}
	\includegraphics[width=0.25\textwidth]{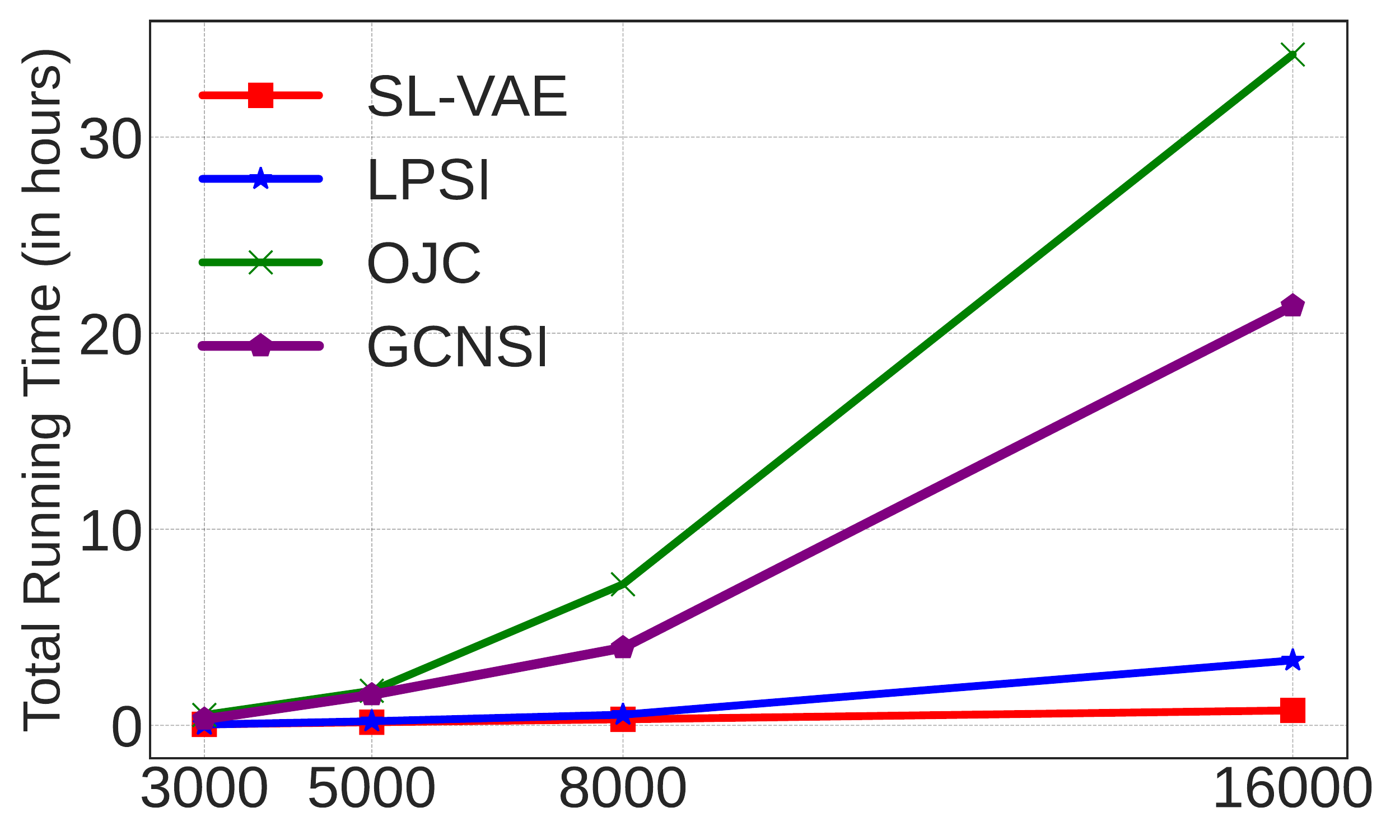}}
	\vspace{-5mm}
		\caption{Runtime comparison in terms of the graph size.}
	\vspace{-7mm}
		\label{fig: running_time}
	\end{figure}
    \subsection{Q4: Model Scalability}
    To analyze the scalability of the proposed model, we record the average runtime  for $10$ rounds for the training of all comparison methods until convergence. The runtime is shown with respect to the number of nodes (i.e., $3,000$, $5,000$, $8,000$, and $16,000$ nodes) in Digg dataset, and the results are presented in Fig. \ref{fig: running_time}: \emph{1)} the runtime of SL-VAE with different forward models (Fig. \ref{fig: running_1}); \emph{2)} the runtime of SL-VAE against other source localization algorithms (Fig. \ref{fig: running_2}). 
    
    As shown in Fig. \ref{fig: running_1}, all variants of SL-VAE demonstrate the linear runtime with the growth of graph size, and differences only depend on the complexity of the forward model. Moreover, as depicted in Fig. \ref{fig: running_2}, only LPSI has comparable runtime with SL-VAE in node size ($\le 5,000$), other models are slower than SL-VAE in operating on large graphs ($\ge 8,000$) since they need to rely on enumerating the graph structure for locating diffusion sources. Since SL-VAE leverages deep generative model to avoid solving the ill-posed source localization problem, it exhibits efficient linear runtime against other models while achieving a better performance.

    \subsection{Visualization}
    Finally, we visualize the overall reconstruction performance of the diffusion sources under the SI diffusion pattern among all comparison methods in Fig. \ref{fig: karate}, where we select two datasets (i.e., Karate and Jazz) with relatively smaller graph size for better demonstration, and the rest of visualizations are exhibited in Appendix. Visually, the diffusion sources recovered by SL-VAE exhibit the most similar distribution to the real ones. For other baselines, LPSI and OJC tend to predict excess amount of sources in both datasets, respectively. Even though both GCNSI and Netsleuth can predict the similar amount of diffusion sources as in the ground truth, their prediction performance is not satisfied since they both fail to correctly quantify the uncertainty. To sum up, these observations align with the quantitative results demonstrated in Table \ref{tab: evaluation_1} and \ref{tab: evaluation_2}.

	\section{Conclusion}\label{sec: con}
Diffusion source localization is an essential yet challenging task with numerous network science applications. This paper proposes a novel yet generic framework, namely SL-VAE, that incorporates the deep generative model to approximate the distribution of diffusion sources. SL-VAE leverages the learned generative priors to approximate the intrinsic patterns of diffusion sources directly, and uncertainty in network information diffusion can thus be quantified. Finally, an integrated objective is derived to jointly learn both the generative model and an arbitrary forward diffusion estimation model to impose the model to generalize under any complex diffusion patterns. Extensive experiments and case studies conducted on seven real-world datasets demonstrate the consistent and superior performance of SL-VAE in various underlying information diffusion patterns. Specifically, the AUC score of SL-VAE outperforms other approaches by, on average, $20\%$ under each diffusion pattern. 
	
	\section*{Acknowledgement}
    This work was supported by the National Science Foundation (NSF) Grant No. 1755850, No. 1841520, No. 2007716, No. 2007976, No. 1942594, No. 1907805, a Jeffress Memorial Trust Award, Amazon Research Award, NVIDIA GPU Grant, and Design Knowledge Company (subcontract number: 10827.002.120.04).
    

	\bibliographystyle{ACM-Reference-Format}
	\bibliography{reference}

\clearpage
\appendix

\section{Appendix}
\subsection{Inference Derivation}
For the final inference function derived in Eq. \eqref{eq: infer_2}, we have:
    \begin{align}
        \mathcal{L}_{infer}(x)
        &= p_{\psi}(y|x, G) \cdot \sum_{z} \sum_{\hat x} p_{\theta}(x|z)\cdot q_{\phi}(z|\hat x)\\\nonumber
        &= -\log p_{\psi}(y|x, G) - \log \big[\sum_{z} \sum_{\hat x} p_{\theta}(x|z)\cdot q_{\phi}(z|\hat x)\big],\nonumber
    \end{align}

    Most existing forward diffusion estimation models $p_{\psi}(y|x, G)$ computes the diffused observation $y \in [0, 1]^{|V|}$ to indicate the probability of how likely this node would be infected, which indicates the value of $y$ is continuous in the range $[0, 1]$ that fit Gaussian distribution. According to this assumption, the left term in Eq. \eqref{eq: infer_2} turns to measure the empirical risk between the ground truth $y$ and $\Tilde y = p_{\psi}(y|x, G)$ from the forward diffusion estimation model. Therefore, the first term in  Eq. \eqref{eq: infer_2} is written as the Mean Squared Loss (MSE): $\norm{y-p_{\psi}(y|\Tilde x, G)}_2^2$. For the second term in Eq. \eqref{eq: infer_2}, we minimize the log likelihood of $\log \big[\sum_{z} \sum_{\hat x} p_{\theta}(x|z)\cdot q_{\phi}(z|\hat x)\big]$. Since the estimated diffusion source $\Tilde x \in \{0, 1\}^{|V|}$ that fit the Bernoulli distribution, minimizing such log likelihood is equivalent to minimizing the probability mass function. Therefore, we can marginalize the probability of both $\hat x$ (from training set) and $z$ (sampled from $q_{\phi}(z|\hat x)$) so that the probability mass function can be denoted as:
    \begin{align}
        \mathcal{L}_{pmf} &= \log \big[\sum_{z} \sum_{\hat x} p_{\theta}(x|z)\cdot q_{\phi}(z|\hat x)\big]\\\nonumber
        &= \log \big[\sum_j^N \prod_i^{|V|} f_{\theta}(z^{j}_{i})^{\Tilde x_{i}}(1-f_{\theta}(z^{j}_{i}))^{(1-\Tilde x_{i})} \big]\Big]\nonumber
    \end{align}

    Therefore, the inference function for optimizing $\Tilde x$ with a diffused observation $y \in [0, 1]^{|V|}$ in Eq. \eqref{eq: infer_2} can be derived as:
    \begin{equation*}
        \Tilde x=\min_{\Tilde x} \Big[\norm{y-p_{\psi}(y|\Tilde x, G)}_2^2 - \log \big[\sum_j^N \prod_i^{|V|} f_{\theta}(z^{j}_{i})^{\Tilde x_{i}}(1-f_{\theta}(z^{j}_{i}))^{(1-\Tilde x_{i})} \big]\Big],
    \end{equation*}

    However, the derived $\mathcal{L}_{pmf}$ may experience numerical underflow when calculating the product term $\prod_i^{|V|} f_{\theta}(z^{j}_{i})^{\Tilde x_{i}}(1-f_{\theta}(z^{j}_{i}))^{(1-\Tilde x_{i})}$ if the graph size $|V|$ is relatively large. Moreover, we can leverage the log-sum-exp trick to simplify the product term. Specifically, suppose $p_j(x, z)$ be the product of the probabilities over $i$ given $j, x$ and $z$, we could drive the $\log [p_j(\Tilde x, z)]$ as:
    \begin{equation*}
        \log[p_j(\Tilde x, z)]=\sum_i \log[f_{\theta}(z^{j}_{i})^{\Tilde x_{i}}(1-f_{\theta}(z^{j}_{i}))^{(1-\Tilde x_{i})}].
    \end{equation*}
    Unlike computing the product of many small values, computing the sum of their logs has little danger of underflow. Now, let $v^*(\Tilde x, z)$ be the maximal value of $\log[p_j(\Tilde x, z)]$ over all $j$ such that:
    \begin{equation*}
        v^*(\Tilde x, z) = \max_j \log[p_j(\Tilde x, z)]
    \end{equation*}
    Using the log-sum-exp trick, the $\mathcal{L}_{pmf}$ can be computed as:
    \begin{equation*}
        \mathcal{L}_{pmf} = v^*(\Tilde x, z)+\log \sum_j^m\big[\exp(\log p_j(\Tilde x, z) - v^*(\Tilde x, z))\big]
    \end{equation*}

\subsection{Additional Experiments}
\subsubsection{Dataset Description.}
\begin{table}[!t]
    \centering
    \resizebox{0.38\textwidth}{!}{%
    \begin{tabular}{@{}cccccc@{}}
    \toprule
             & Nodes & Edges & Average Degree  & Clustering Coefficient  \\ \midrule \midrule
    Karate & 34    & 78    & 2.294            &  0.255                   \\ \midrule 
    Jazz &  198   & 2,742    & 27.69            &  0.52                      \\ \midrule 
    Cora-ML & 2,810    & 7,981  & 5.68           &  0.246                    \\ \midrule 
    Network Science   & 1,565  & 13,532 & 17.29          &       0.741           \\ \midrule
    Power Grid & 4,941    & 6,594  & 2.669          &  0.081                       \\ \midrule
    MemeTracker     & 12,529  & 70,466 & 11.248 &       0.092         \\ \midrule
    Digg     & 15,912 & 78,649 & 9.885 &      0.083          \\ \bottomrule
    \end{tabular}%
    }
    \caption{Dataset Overview}
    \label{tab: dataset}
    \vspace{-9mm}
    \end{table}
We conduct experiments on $7$ real-world datasets, the basic statistics are shown in Table \ref{tab: dataset}:
\begin{itemize}[leftmargin=*]
        \item \textit{Karate} \cite{rossi2015nr}. Each node in the network is a member and each edge represents a tie between two members of the club.
        \item \textit{Jazz} \cite{rossi2015nr}. This dataset is a Jazz musicians collaboration network, where each node represents a musician and each edge represents two musicians have played together in a band.
        \item \textit{Cora-ML} \cite{rossi2015nr}. This network contains computer science research papers, where each node represents a paper and each edge represents one paper cites the other one.
        \item \textit{Power Grid} \cite{watts1998collective}. This is a topology network of the Western States Power Grid of the US. An edge represents a power supply line. A node is either a generator, a transformator or a substation.
        \item \textit{Network Science} \cite{rossi2015nr}. This is a coauthorship network between scientists working on network theory, where nodes represent scientists and edges represent two scientists have a collaborated. 
        \item \textit{Digg}~\cite{rossi2015nr}. This is a real-world social network that contains stories promoted to Digg's front page over a period of a month in 2009. The propagation of each story is denoted as one diffusion cascade. 
        \item \textit{Memetracker} \cite{leskovec2009meme}. Memetracker tracks the posts that appear most frequently over time across this entire online news spectrum. The propagation of each story is denoted as one diffusion cascade. 
    \end{itemize}

\subsubsection{Baseline Description.}
We evaluate the performance of SL-VAE against other source localization approaches. It is worth noting that a sturdy source localization algorithm should be capable of locating diffusion sources under different diffusion patterns. Therefore, we choose two epidemic models (i.e., SI and SIR) as the underlying propagation model to test the effectiveness of SL-VAE and approaches. Considering that the SIS model is similar to SI and SIR model, we only use SI and SIR model for comparison. As shown in Table \ref{tab: under}, for comparison methods, LPSI and GCNSI are tested under both diffusion patterns. NetSleuth is only tested under the SI diffusion pattern because of its design. Similarly, OJC is particularly designed for locating sources under the SIR diffusion pattern, it can be adapted to SI model with minor modifications.
\begin{table}[!h]
    \centering
    \resizebox{0.32\textwidth}{!}{%
    \begin{tabular}{@{}cccccc@{}}
    \toprule
        & LPSI  & OJC   & NetSleuth & GCNSI  & SL-VAE \\ \midrule
    SI  & $\checkmark$   & $\checkmark/\times$   & $\checkmark$     & $\checkmark$     & $\checkmark$     \\
    SIR & $\checkmark$ & $\checkmark$ & $\times$     & $\checkmark$ & $\checkmark$ \\ \bottomrule
    \end{tabular}%
    }
    \caption{Testing  under different propagation patterns}
    \label{tab: under}
    \vspace{-8mm}
    \end{table}
    
\subsubsection{Additional Visualization.}
We provide additional visualizations of the diffusion source prediction under the SI diffusion pattern among all baselines as follows. The predicted/ground truth diffusion sources are all marked with green while the rest of nodes are marked with red. 
\begin{figure*}[!t]
		\subfloat[GCNSI]{\label{fig: coraml_gcnsi}
			\includegraphics[width=0.165\textwidth]{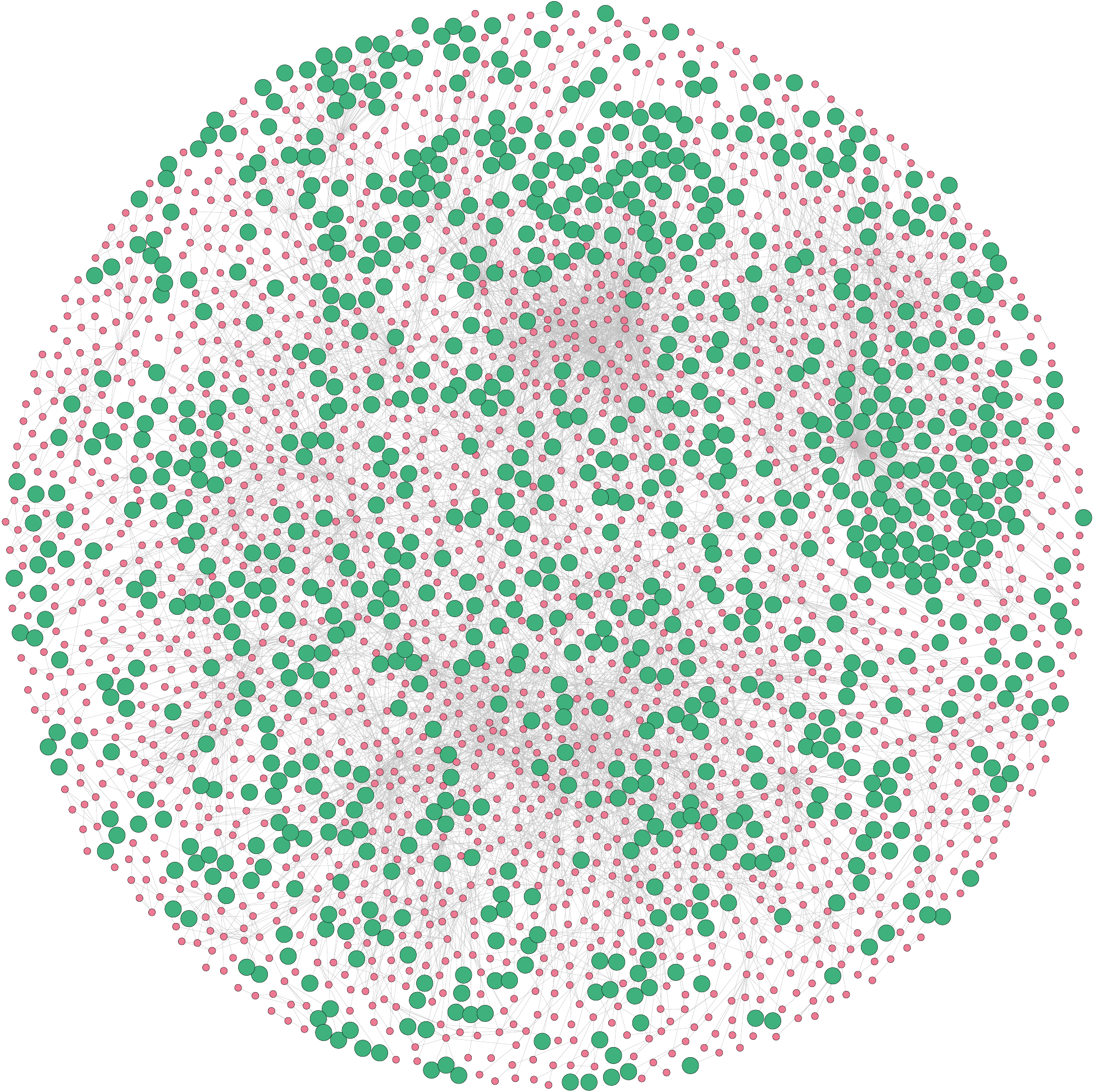}}
		\subfloat[LPSI]{\label{fig: coraml_lpsi}
			\includegraphics[width=0.165\textwidth]{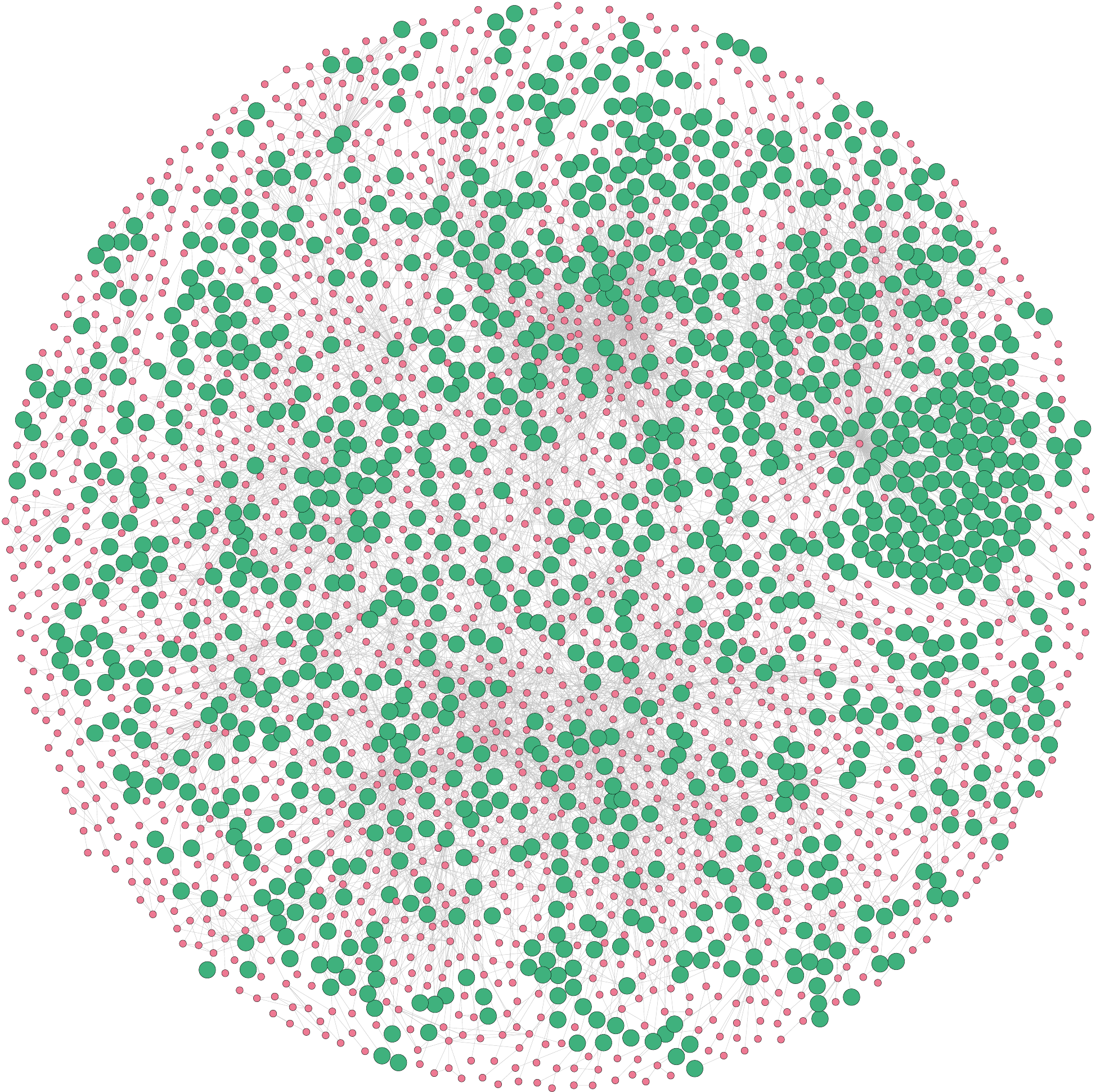}}
		\subfloat[Netsleuth]{\label{fig: coraml_netslu}
			\includegraphics[width=0.165\textwidth]{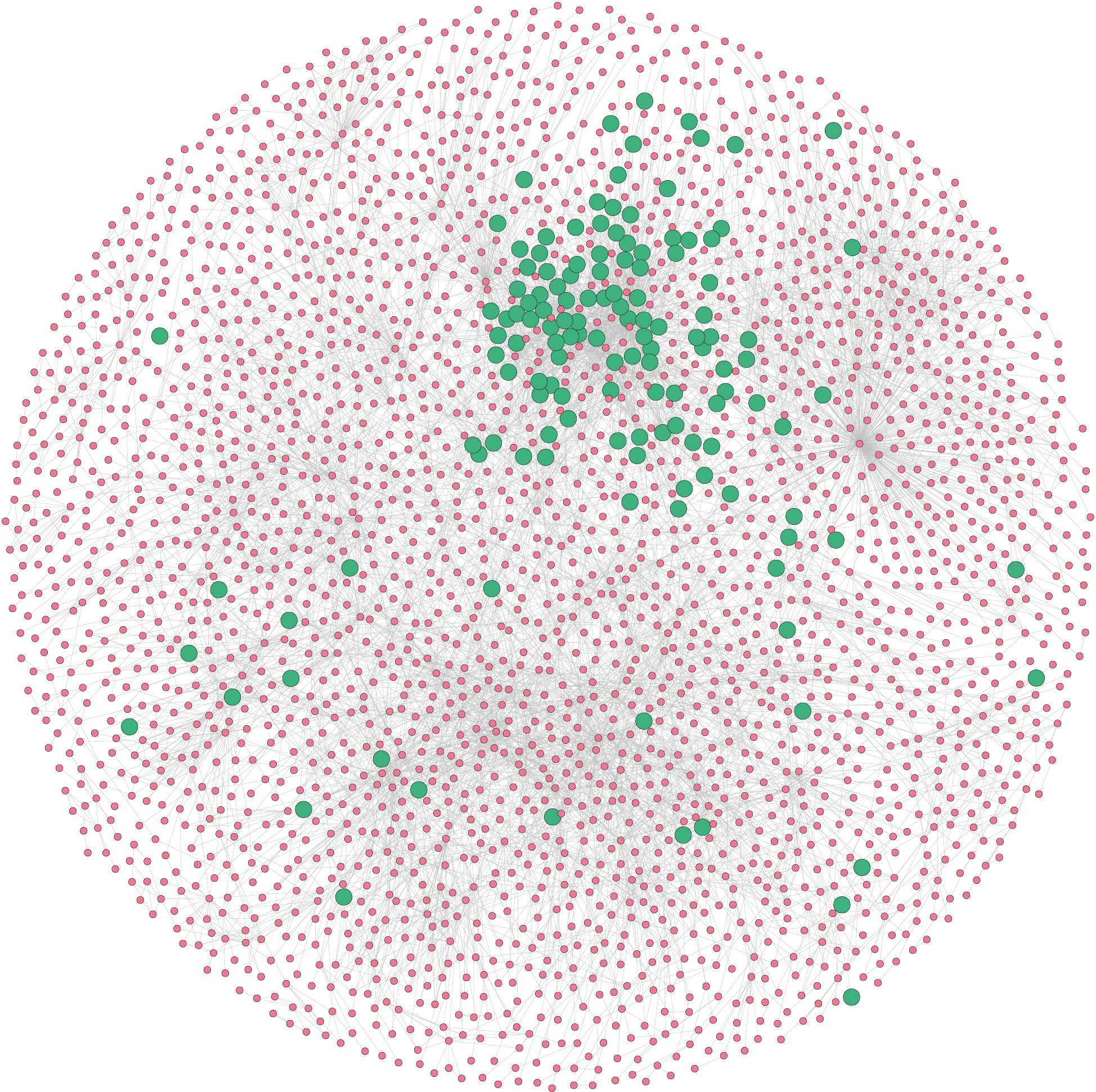}}
		\subfloat[OJC]{\label{fig: coraml_ojc}
			\includegraphics[width=0.165\textwidth]{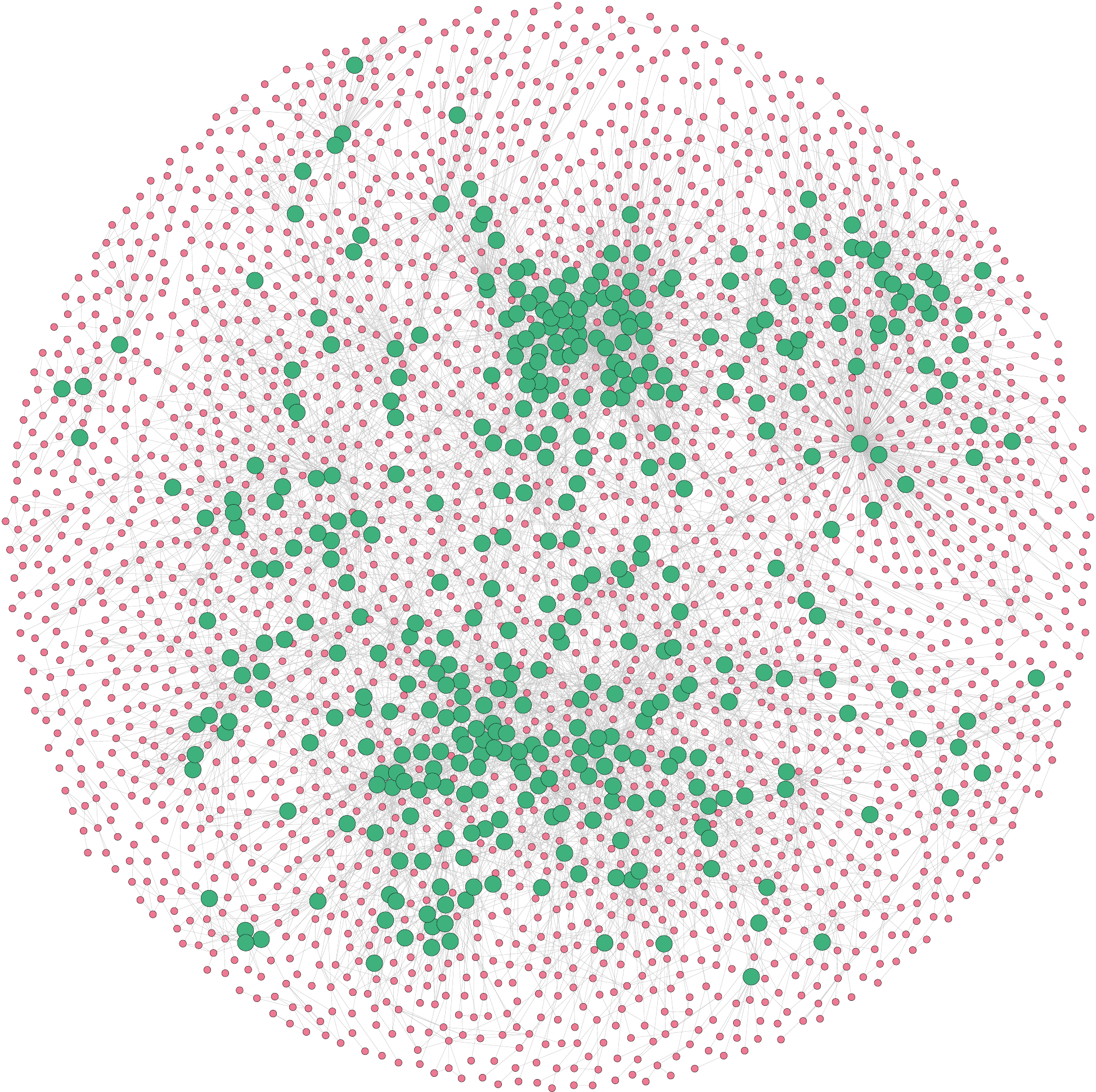}}
		\subfloat[SL-VAE]{\label{fig: coraml_slvae}
			\includegraphics[width=0.165\textwidth]{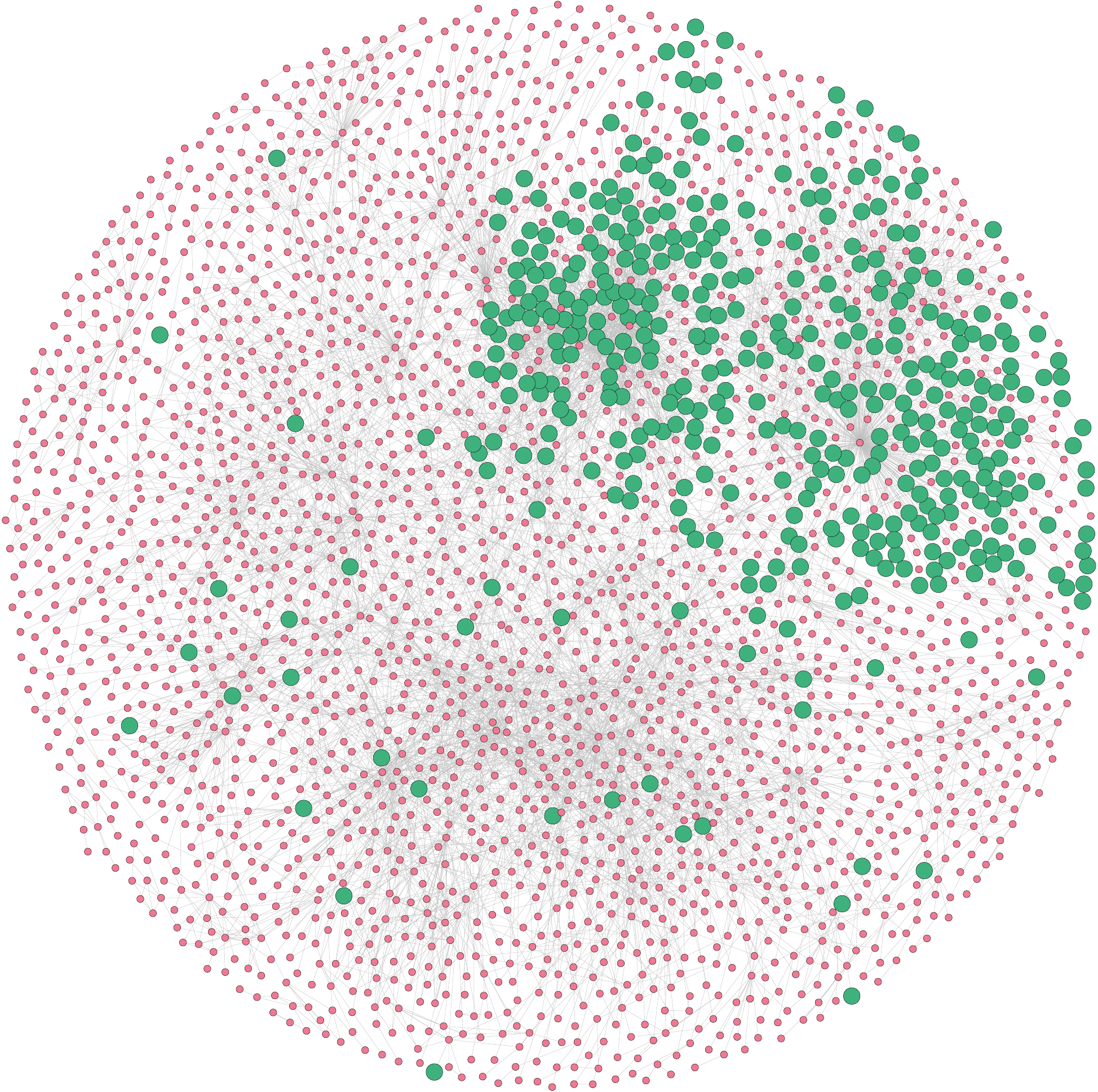}}
		\subfloat[Real]{\label{fig: coraml_real}
			\includegraphics[width=0.165\textwidth]{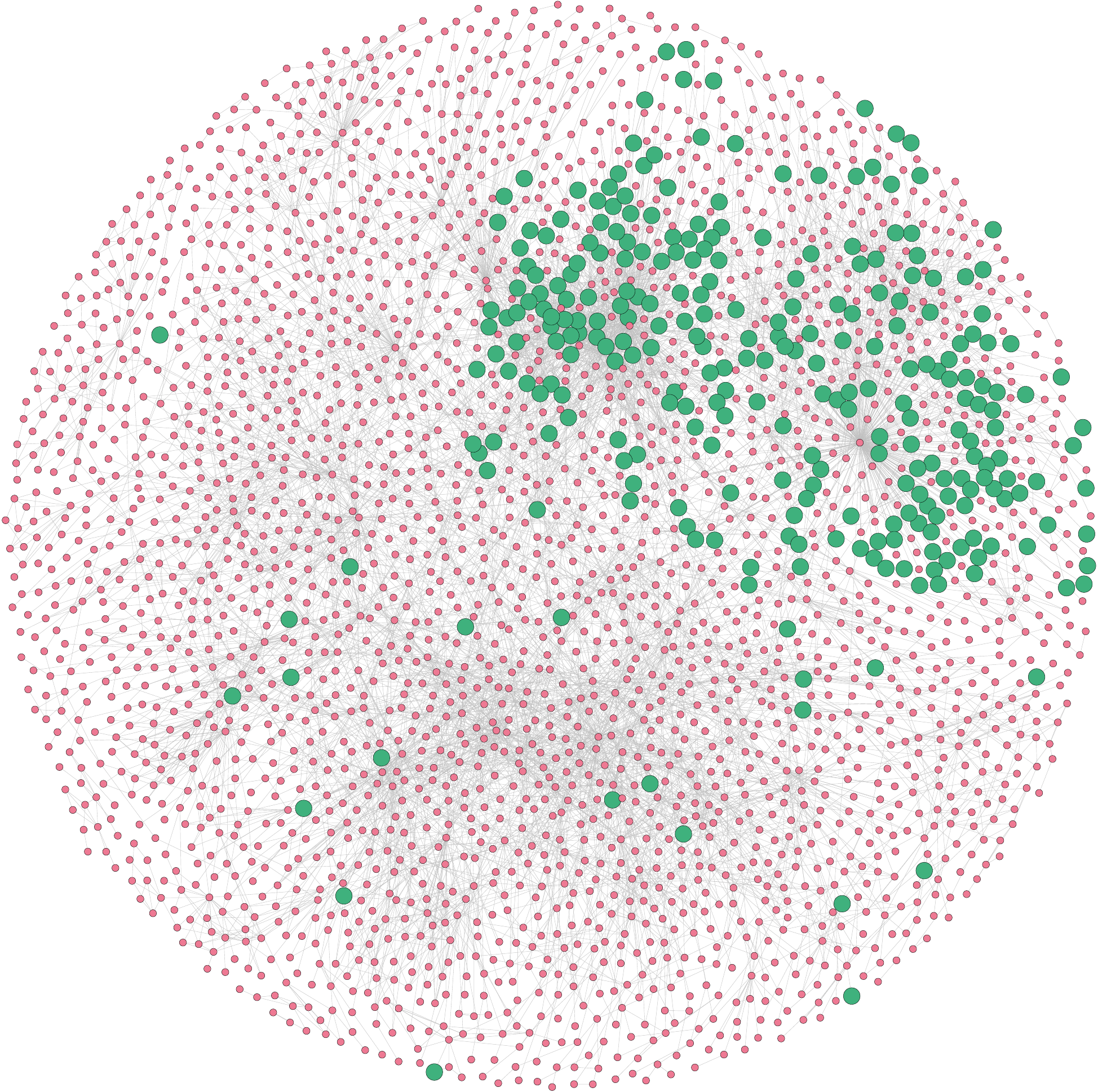}}
		\vspace{-3mm}
		\caption{The visualized comparison between the generated diffusion sources and the ground truth (Jazz).}
		\label{fig: cora_ml}
	\end{figure*}

\begin{figure*}[!t]
		\subfloat[GCNSI]{\label{fig: net_gcnsi}
			\includegraphics[width=0.165\textwidth]{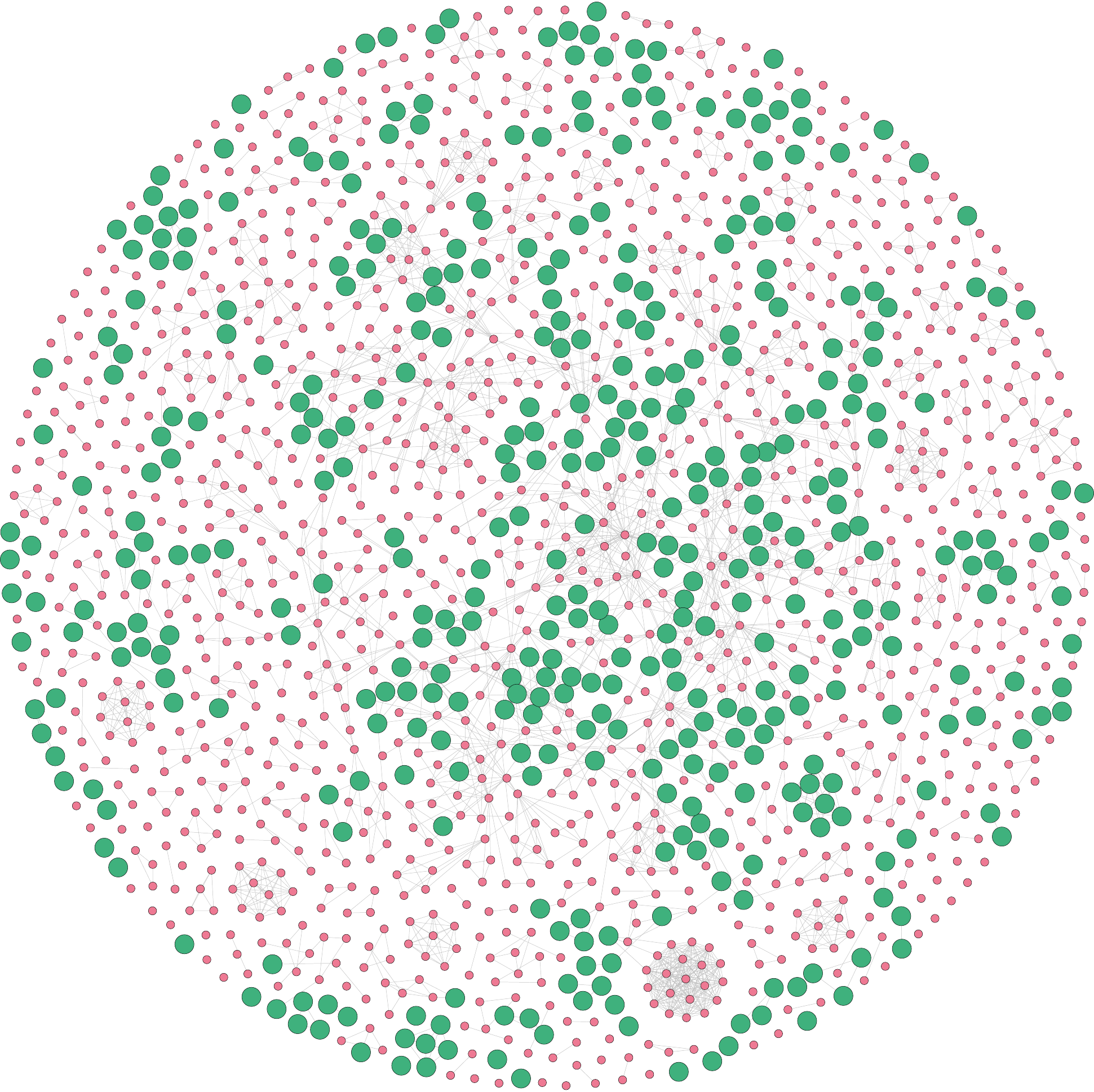}}
		\subfloat[LPSI]{\label{fig: net_lpsi}
			\includegraphics[width=0.165\textwidth]{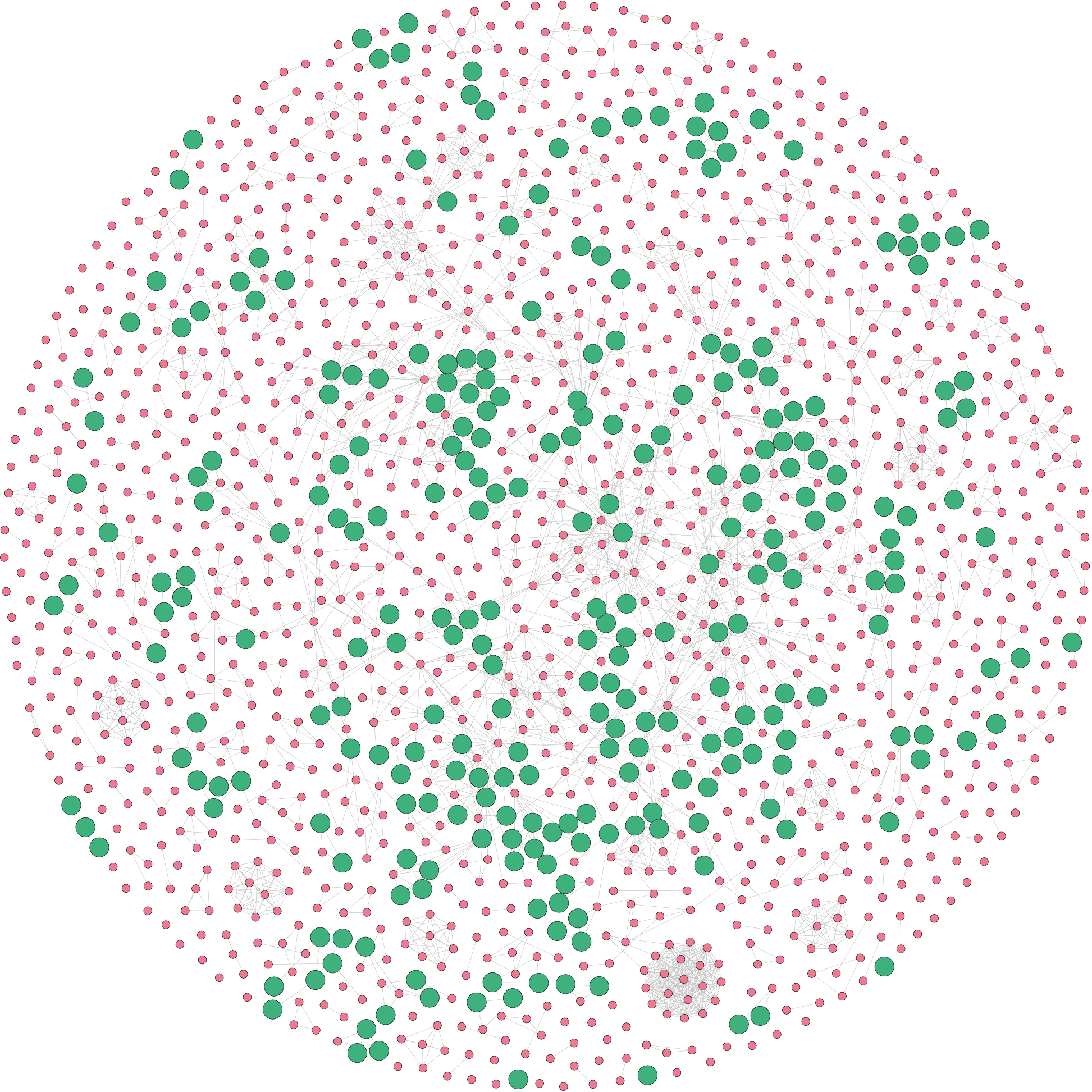}}
		\subfloat[Netsleuth]{\label{fig: net_netslu}
			\includegraphics[width=0.165\textwidth]{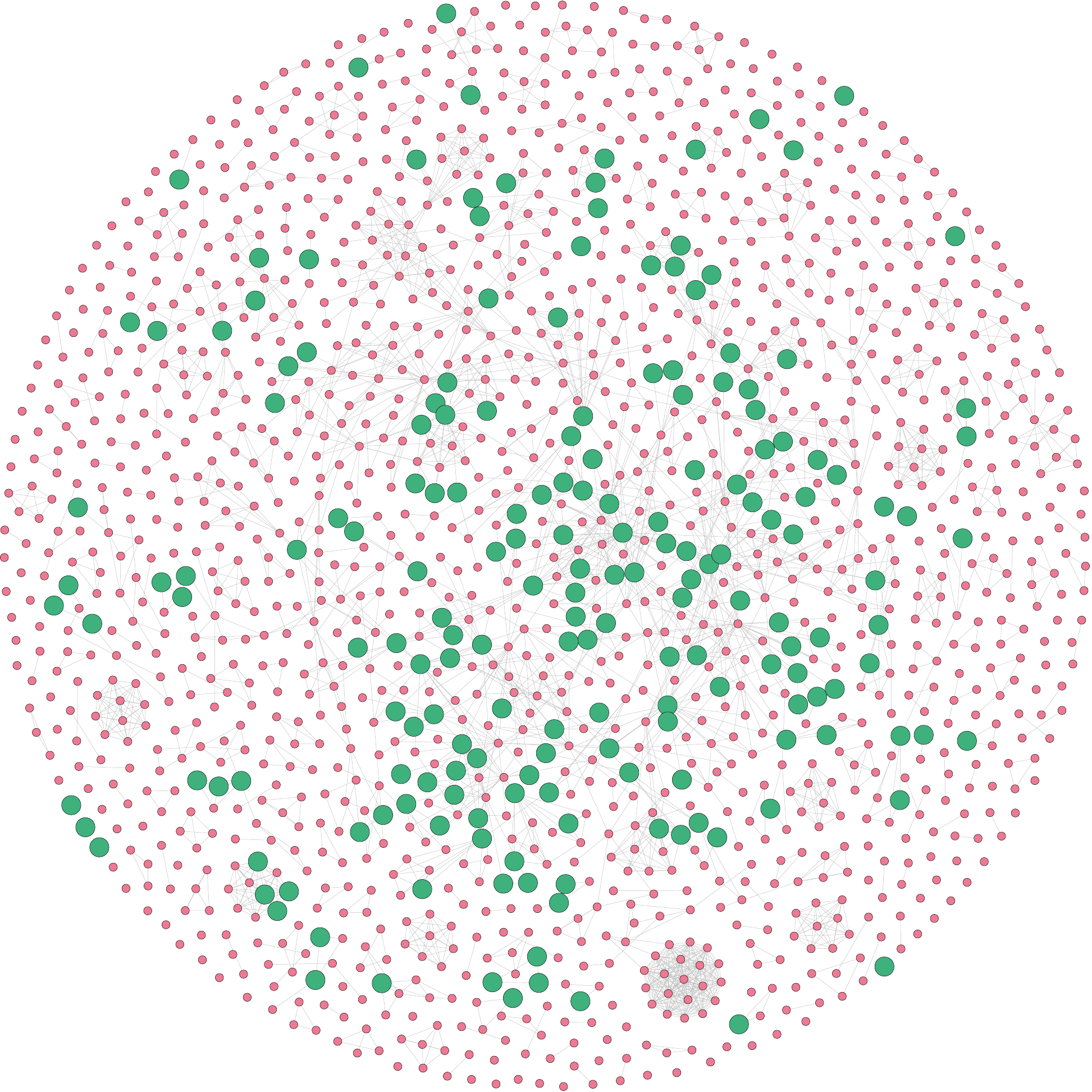}}
		\subfloat[OJC]{\label{fig: net_ojc}
			\includegraphics[width=0.165\textwidth]{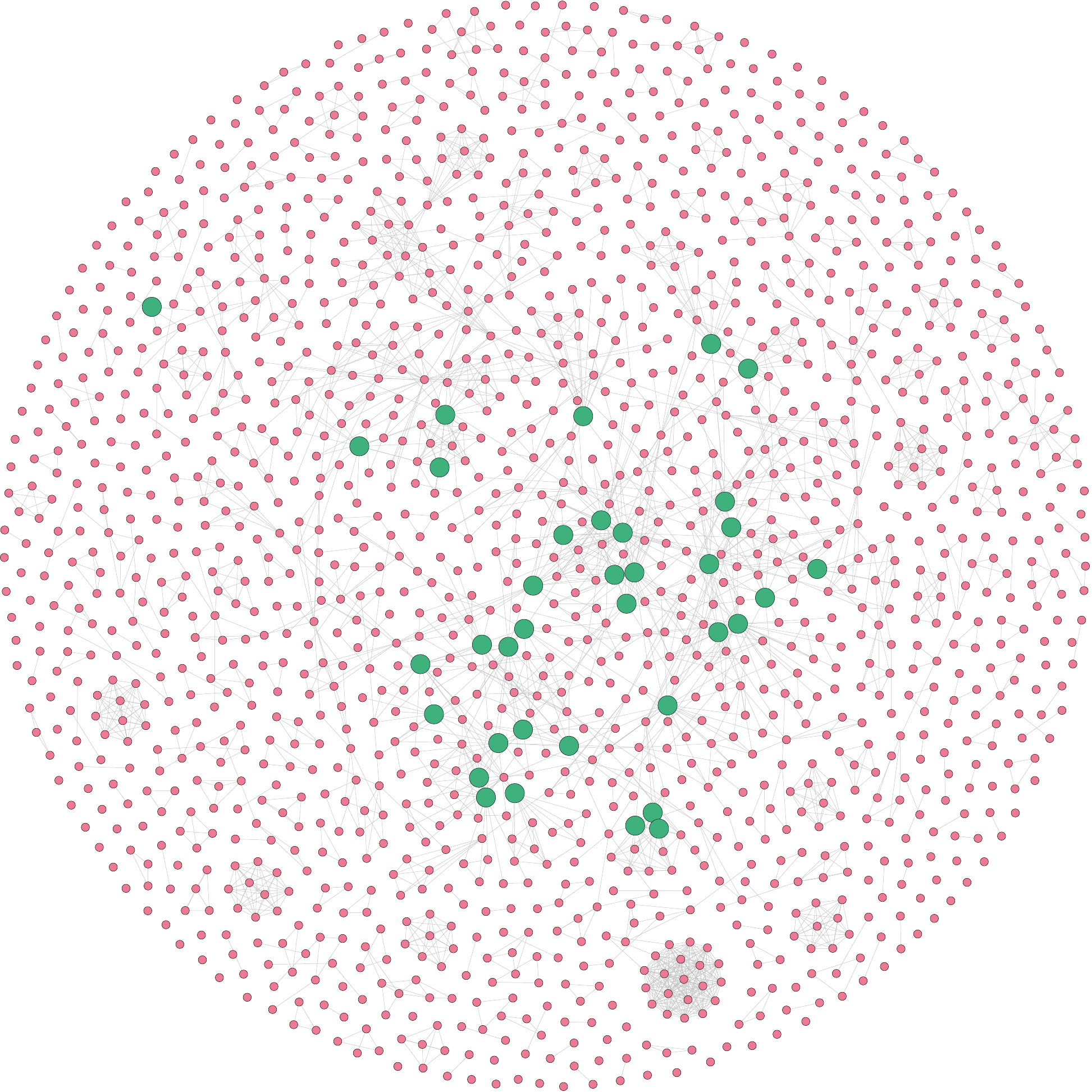}}
		\subfloat[SL-VAE]{\label{fig: net_slvae}
			\includegraphics[width=0.165\textwidth]{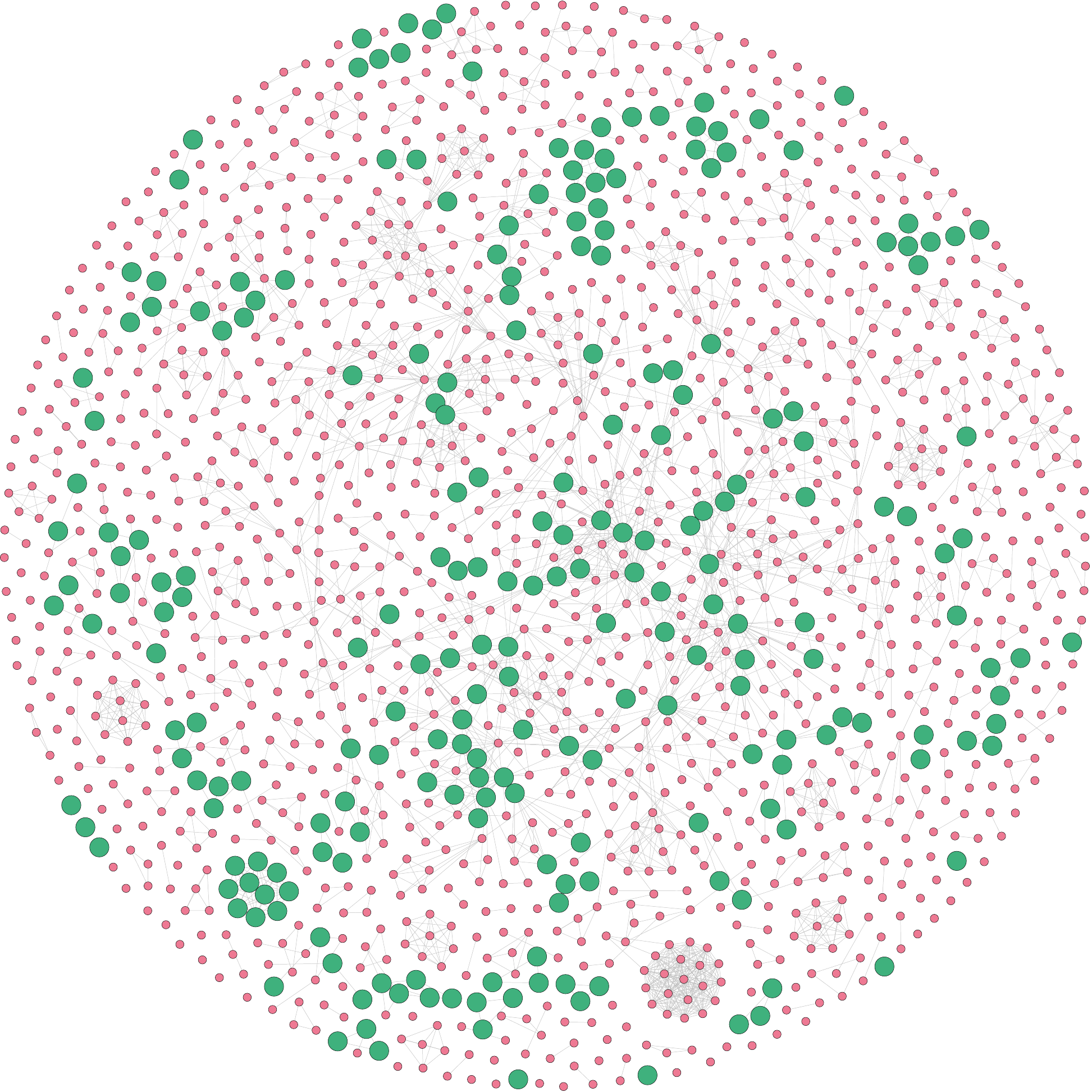}}
		\subfloat[Real]{\label{fig: net_real}
			\includegraphics[width=0.165\textwidth]{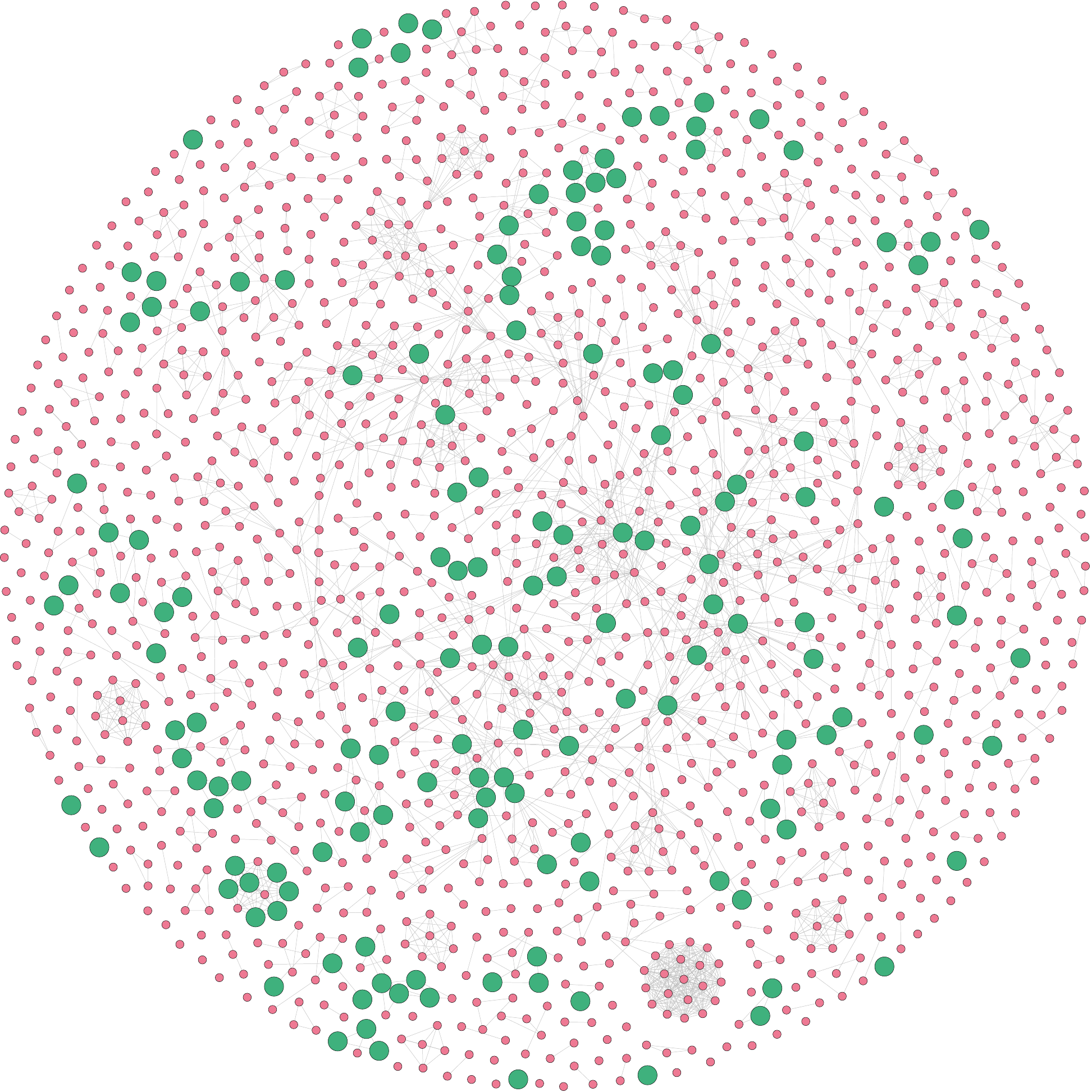}}
		\vspace{-3mm}
		\caption{The visualized comparison between the generated diffusion sources and the ground truth (Network Science).}
		\label{fig: net}
	\end{figure*}
	
\begin{figure*}[!t]
		\subfloat[GCNSI]{\label{fig: powergrid_gcnsi}
			\includegraphics[width=0.3\textwidth]{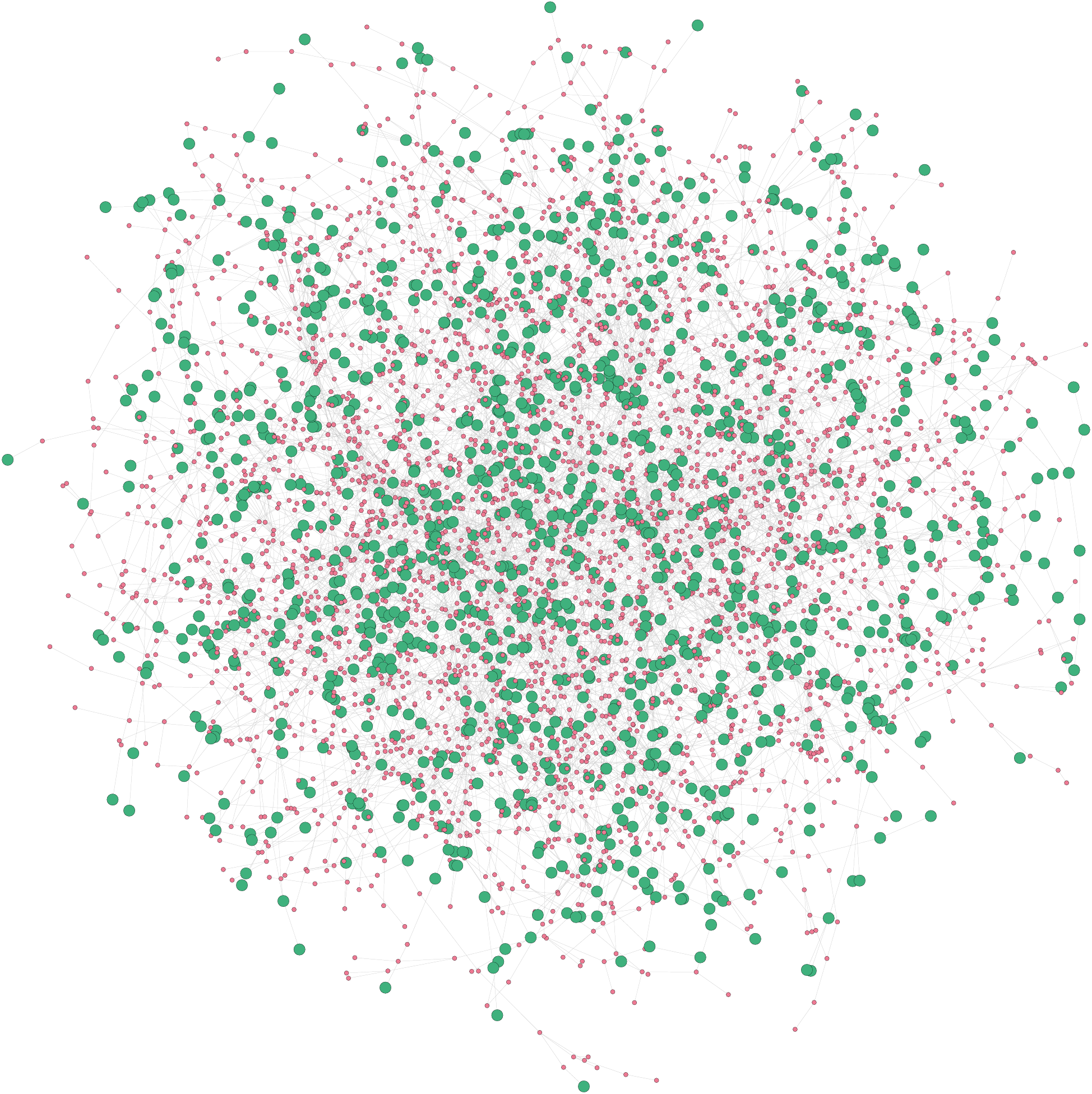}}
		\subfloat[LPSI]{\label{fig: powergrid_lpsi}
			\includegraphics[width=0.3\textwidth]{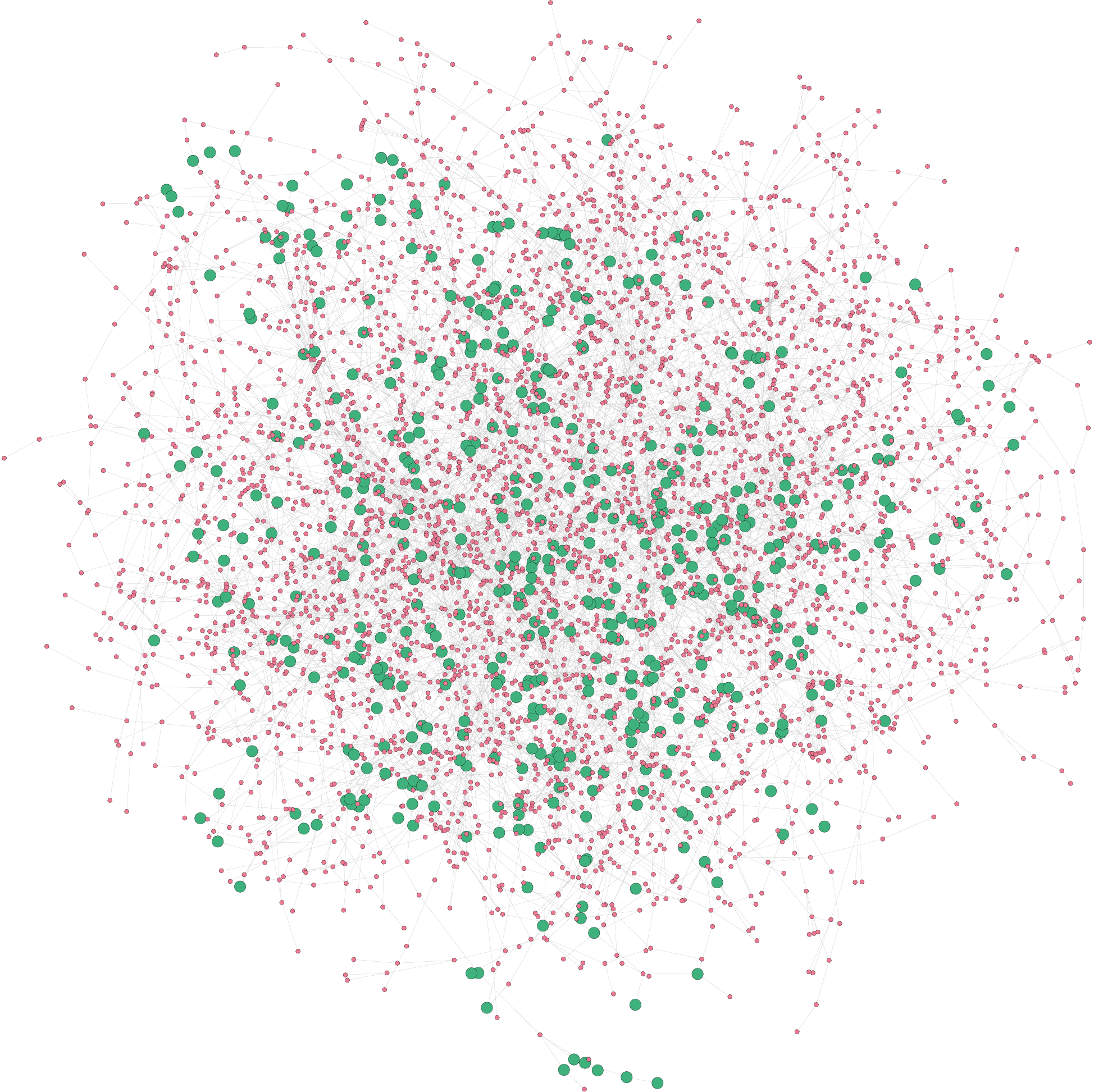}}
		\subfloat[Netsleuth]{\label{fig: powergrid_netslu}
			\includegraphics[width=0.3\textwidth]{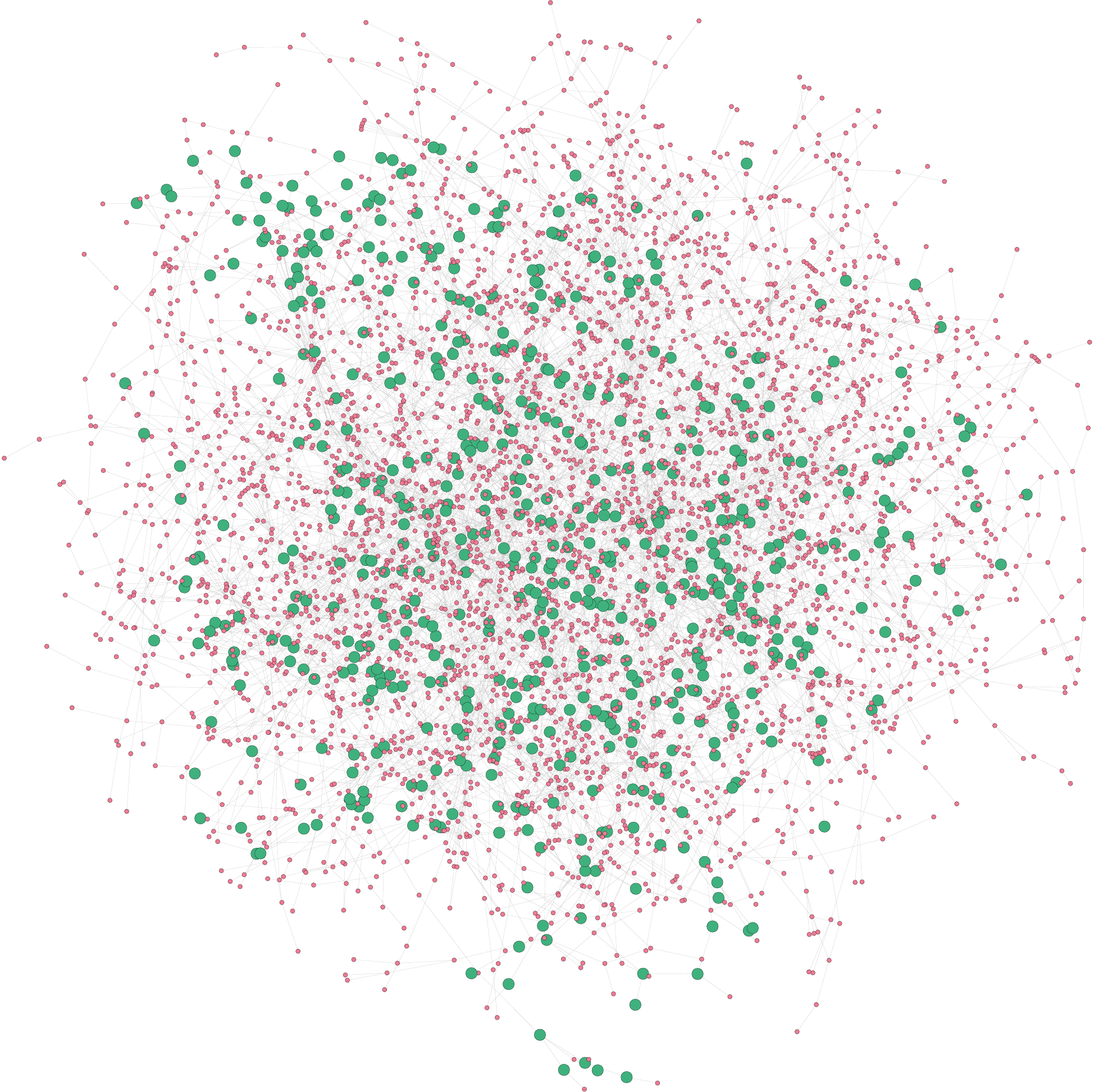}}\\\vspace{-3mm}
		\subfloat[OJC]{\label{fig: powergrid_ojc}
			\includegraphics[width=0.3\textwidth]{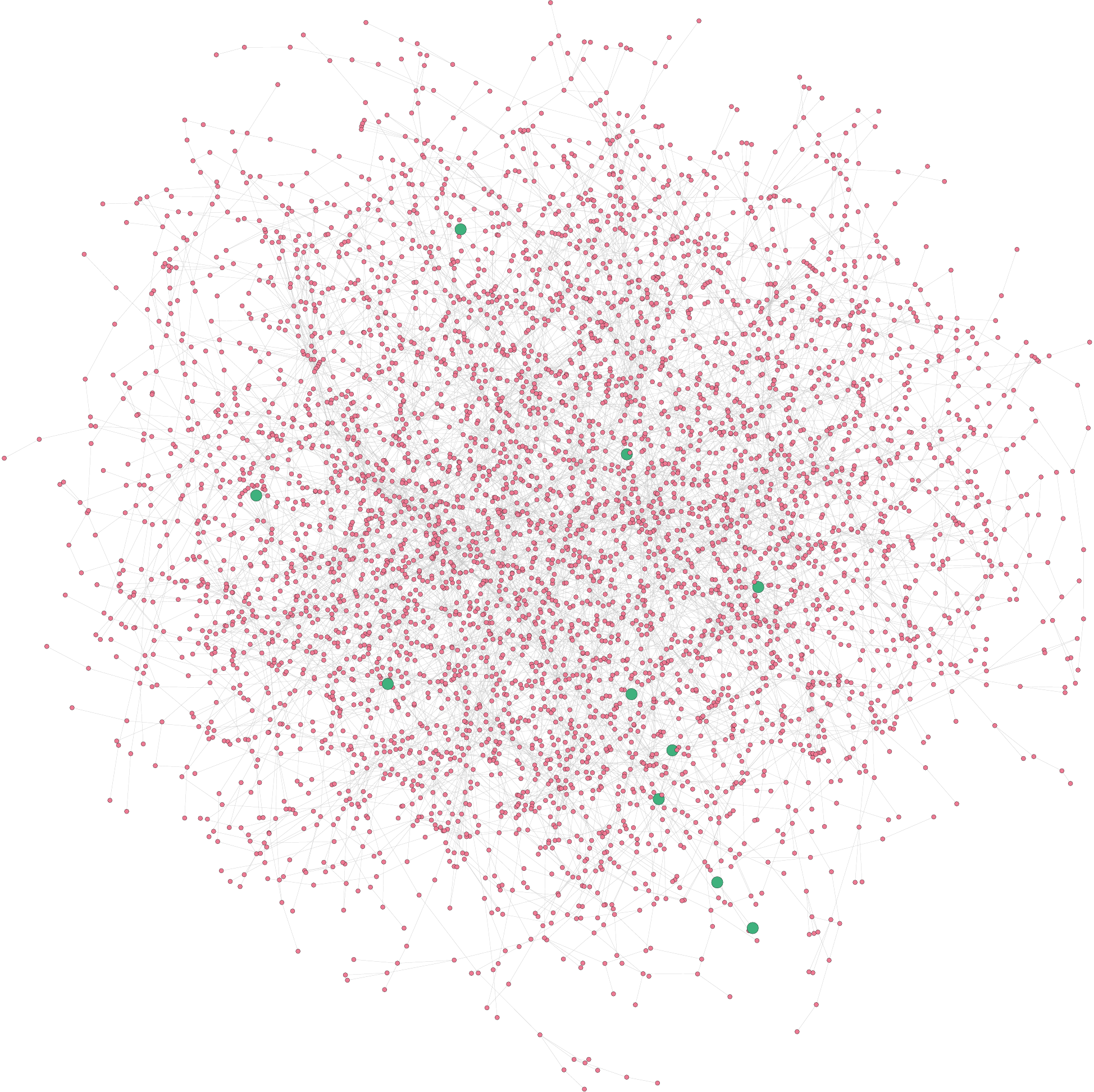}}
		\subfloat[SL-VAE]{\label{fig: powergrid_slvae}
			\includegraphics[width=0.3\textwidth]{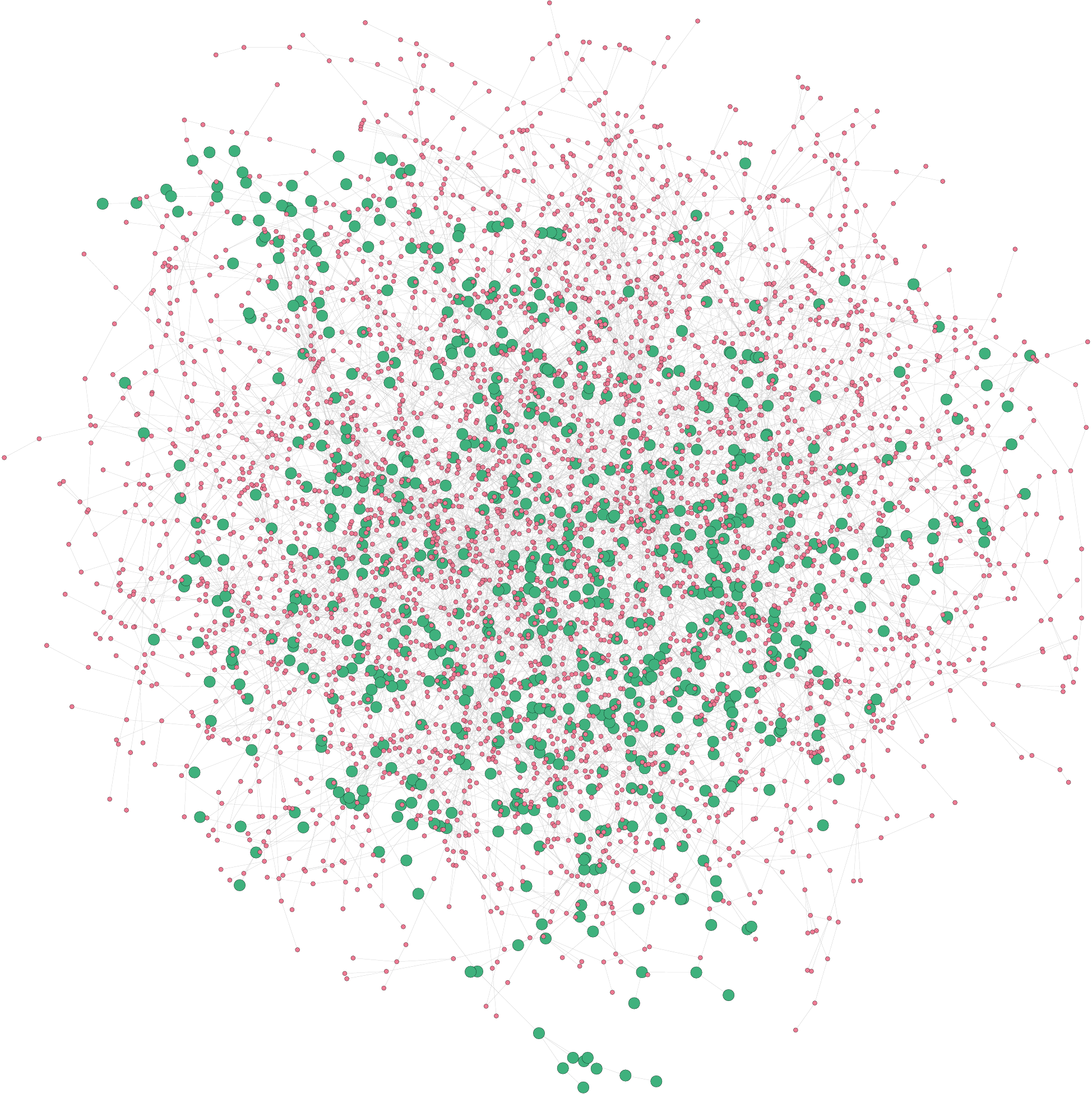}}
		\subfloat[Real]{\label{fig: powergrid_real}
			\includegraphics[width=0.3\textwidth]{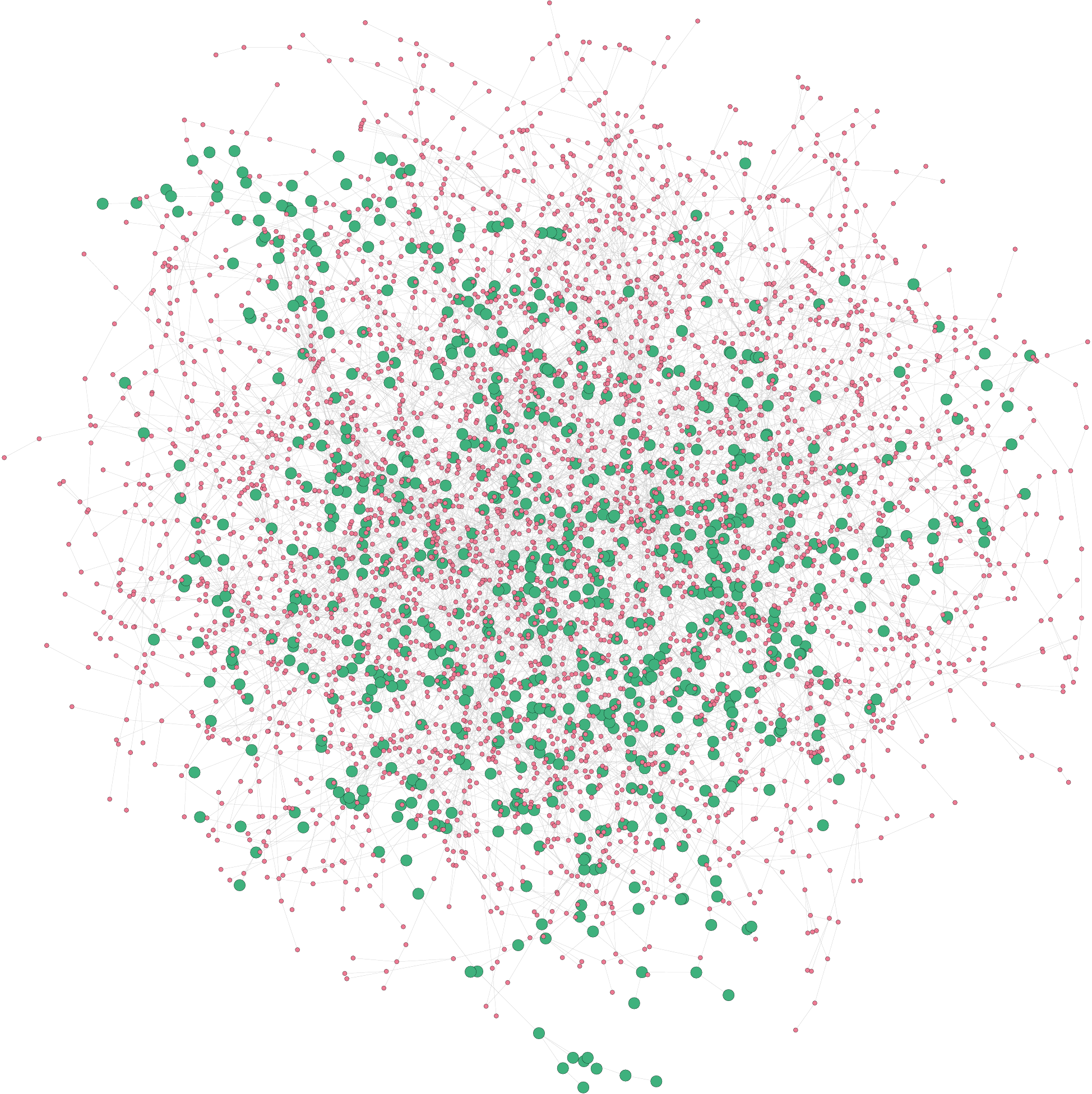}}
		\caption{The visualized comparison between the generated diffusion sources and the ground truth (Power Grid).}
		\label{fig: power}
	\end{figure*}
	
\end{document}